\documentclass[10pt]{article}

\input{macros}

\usepackage{mathrsfs}
\usepackage{todonotes}
\usepackage{booktabs}

\allowdisplaybreaks

\makeatletter
\newenvironment{breakablealgorithm}
{%
    \par
    \refstepcounter{algorithm}%
    \hrule height 0.8pt depth 0pt
    \kern 2pt
    \renewcommand{\caption}[2][\relax]{%
        {\raggedright
        \textbf{\ALG@name~\thealgorithm.} ##2\par}%
        \ifx\relax##1\relax
            \addcontentsline{loa}{algorithm}
            {\protect\numberline{\thealgorithm}##2}%
        \else
            \addcontentsline{loa}{algorithm}
            {\protect\numberline{\thealgorithm}##1}%
        \fi
        \kern 2pt
        \hrule height 0.8pt depth 0pt
        \kern 2pt
    }%
}
{%
    \kern 2pt
    \hrule height 0.8pt depth 0pt
    \par
}
\makeatother

\begin{document}

\title{The Price of Current-Effect Inference in Nonstationary Multi-Armed Bandits}
\author{%
	Sharan Sahu\,\thanks{Department of Statistics and Data Science,  Cornell University; \, Email: \href{mailto:ss4329@cornell.edu}{\texttt{ss4329@cornell.edu}}. } \and
    Martin T. Wells\,\thanks{Department of Statistics and Data Science,  Cornell University; \, Email: \href{mailto:mtw1@cornell.edu}{\texttt{mtw1@cornell.edu}}. }
} 
\date{\today}
\maketitle

\begin{abstract}
Adaptive experiments use bandit algorithms to allocate more observations to interventions with higher expected rewards. However, decision-makers often also require reliable inference about \emph{current treatment effects}: the differences between the expected rewards of interventions and a reference intervention at the present round, rather than averages over past rounds. These objectives create a fundamental tension: reward maximization favors better interventions, whereas estimating their current effects requires recent observations from interventions that may be suboptimal. When outcomes change over time, historical observations become outdated, making accurate inference dependent on continued exploration.

We characterize this tradeoff for $K$-arm bandits whose mean rewards satisfy a known H\"older smoothness condition with exponent $\beta\in(0,1]$ and constant $L$. We establish matching minimax bounds, up to logarithmic factors, on expected dynamic regret subject to simultaneous coverage with high probability for all current effects relative to a reference arm, with confidence-interval width at most $w$. For a horizon $T$, our proposed policy, \textsc{TRACE} (\emph{Tracking Rewards and Current Effects}), achieves expected dynamic regret of order $\widetilde{\mathcal O}_{\beta}\left( \sqrt{KT} +K^{\beta/(2\beta+1)}L^{1/(2\beta+1)}T^{(\beta+1)/(2\beta+1)} +Kw^{-2}\left(1+L^{1/\beta}w^{-1/\beta}\right)
\right)$, where $\widetilde{\mathcal O}_{\beta}$ suppresses logarithmic factors and constants depending only on $\beta$. This \emph{regret--precision tradeoff} separates the standard cost of reward learning from the additional cost of refreshing information about current effects. Notably, the inferential cost remains unavoidable even when the same arm is uniformly optimal. 

We further extend \textsc{TRACE} with an empirical-Bernstein e-process that adapts exploration to realized prediction errors while preserving the worst-case guarantee. When the realized drift within an inference block is sufficiently small, all observations collected by \textsc{TRACE} can be reused for asymptotically normal pointwise inference and, under stronger local stability, locally asymptotically efficient estimation. We also show that regret constraints can limit honest adaptation to unknown smoothness. We support our theoretical findings through extensive numerical experiments.
\end{abstract}

\tableofcontents

\section{Introduction}
\label{sec:introduction}

Bandit algorithms provide a framework for learning which decisions perform well while using that information to improve outcomes during an experiment~\citep{LS20}. In each round, the learner selects an intervention and observes only its outcome, creating a tradeoff between collecting information and choosing interventions that appear more rewarding. This framework has been used in applications that include online recommendation and digital experimentation, adaptive clinical trials and mobile-health interventions, personalized treatment assignment, and sequential policy design~\citep{LCLS10,Scott15,VBW15,TM17,KS21}. However, in many of these settings, cumulative reward is not the only objective. Decision-makers may also want reliable estimates of \emph{current treatment effects}, meaning the differences in expected rewards between interventions and a reference intervention at the present time, rather than effects averaged over the history of the experiment.

Why estimate the current effect rather than an effect averaged over the history of the experiment? The distinction matters whenever treatment effects vary over time, and the inferential question concerns an immediate decision. This variation occurs naturally in biological systems. For example, the efficacy of cancer immunotherapy treatment can change as a function of cancer cells becoming more resistant or due to changes in physiological and immune responses over time~\citep{SharmaEtAl10,ChungEtAl22,WangEtAl24}. A time-averaged effect can therefore place greater weight on observations generated under conditions that are no longer relevant and subsequently describe past performance accurately while providing a poor description of present performance. That is, an intervention can have a positive average effect over the experiment while having little benefit or even a negative effect at the current round. For such decisions, the relevant estimand is naturally the contemporaneous effect rather than its historical average. 

This inferential objective becomes particularly difficult when outcomes change over time. Classical stochastic bandits assume fixed arm means, whereas nonstationary models allow intervention performance to evolve~\citep{lai1985asymptotically, GM11,BGZ14,JXKF23}. Under drift, observations collected earlier may no longer accurately describe current performance. Consequently, reward maximization and current-effect inference place fundamentally different demands on the data. Once an intervention appears inferior, a reward-maximizing learner has little reason to continue selecting it. In contrast, estimating its present effect requires sufficiently recent observations because older observations become progressively less informative about the current state. The following examples illustrate when this distinction matters.

\paragraph{Online experimentation and recommendation.}
Bandit methods can adapt recommendations and online decisions based on observed user feedback~\citep{LCLS10,Scott15}. A platform comparing several recommendation or ranking strategies may also want to know how much each strategy currently improves engagement relative to its current production baseline. When reward distributions change over time, an estimate averaged across the experiment answers a different question from an estimate of the effect now. Temporal nonstationarity can arise in several forms, including seasonality (periodic shifts), changes in customer demographics or user markets (covariate shifts), and product changes (structural shifts), each of which can make historical observations less representative of current performance. 

\paragraph{Adaptive interventions.}
Bandit methods have also been studied for adaptive clinical trials, treatment assignment, and mobile-health interventions~\citep{VBW15,TM17,KS21}. The efficacy of treatment can evolve as the enrolled population changes, behavioral responses adapt, or the underlying biological state varies over time. An estimate averaging over earlier participants can therefore remain statistically precise while no longer accurately describing the treatment effect for participants currently entering the experiment. An experimenter may consequently want to improve participant outcomes while also maintaining an accurate estimate of each intervention's current benefit. However, simply concentrating observations on the intervention with the highest estimated reward may subsequently reduce the recent information available about the alternatives~\citep{VBW15,SLW25}.

\paragraph{Continuous monitoring.}
Experimenters may inspect the results repeatedly and make decisions at data-dependent times. Anytime-valid inference and confidence sequences provide guarantees that remain valid under continuous monitoring~\citep{HRMS21, JKPW22,RGVS23,WR24}. However, when effects change over time, there is an additional challenge: the reported interval must describe the current effect rather than a time-averaged effect. Optional-stopping validity alone does not address the bias created by outdated observations. 

\paragraph{Quantum computing.}
Quantum fault-tolerant computing seeks to perform reliable computations on noisy quantum hardware through quantum error correction and fault-tolerant protocols that suppress logical errors below a threshold \citep{shor1995scheme,aharonov1997fault,fowler2012surface}. Although these methods provide a path to scalable quantum computation, fault tolerance incurs substantial overhead in physical qubits, gates, and calibration effort \citep{fowler2012surface,katabarwa2024early, gong2026selfcalibrating}. In practice, operating and optimizing a fault-tolerant system requires repeatedly selecting among alternative error-correction strategies, calibration settings, decoding procedures, or hardware configurations while observing noisy performance outcomes. This naturally gives rise to a bandit problem, where each action corresponds to a candidate intervention, and rewards depend on an evolving quantum device whose noise characteristics may drift over time \citep{BGZ14,gong2026selfcalibrating}. A central challenge is therefore to balance exploration and exploitation while simultaneously making valid inference about the current effect of an intervention in a nonstationary environment. Understanding this tension between adaptive allocation and current-effect inference is essential for the continual characterization, calibration, and scaling of fault-tolerant quantum computing systems \citep{katabarwa2024early,gong2026selfcalibrating}.\\

\subsection{Inference from adaptive experiments}
\label{sec:intro-adaptive-inference}

Adaptive allocation complicates inference because the observations collected for each intervention depend on previous outcomes. This issue has motivated a broad literature on inference from adaptively collected data, including bias-correction methods for adaptive linear models~\citep{DMST18}, stabilized and adaptively weighted estimators that account for changing assignment probabilities~\citep{HHZWA21}, and weighted $M$-estimators that provide asymptotically valid inference under adaptive sampling~\citep{ZJM21}. Related work develops doubly robust inference for contextual-bandit data~\citep{BDKCV21} and semiparametrically efficient inference for sequential experiments with adaptive treatment assignment~\citep{CMR24}. These approaches explicitly account for the sampling mechanism rather than treating adaptively collected observations as if they were from a fixed design. A related line of work develops guarantees that hold uniformly over time. Nonasymptotic confidence sequences provide time-uniform inference, including empirical-Bernstein constructions~\citep{HRMS21}, while betting-based confidence sequences and e-processes provide general tools for anytime-valid inference~\citep{RGVS23,WR24}. These ideas have also been developed for adaptive experiments~\citep{CMR24} and contextual bandits with adaptive logging and time-varying policy values~\citep{WWRKM24}.

The tension between adaptive allocation and statistical inference has also been studied from an experimental-design perspective, including response-adaptive clinical-trial designs that balance assigning more participants to better-performing treatments against preserving statistical power for treatment comparisons, as well as settings where the experimenter values both outcomes during the experiment and the quality of the policy selected afterward~\citep{VBW15,KS21}. 

Closely related to our work, minimax formulations characterize regret--estimation tradeoffs and Pareto-optimal designs in multi-armed bandit experiments~\citep{SLW25}, and randomized allocation can be combined with bandit policies to obtain asymptotically anytime-valid inference for average treatment effects~\citep{LB24}. In nonstationary settings, \citet{SWZ26} study experimental design under structured temporal trends and characterize the tradeoff between estimating dynamic treatment effects and minimizing regret, while \citet{SWW26} study adaptive A/B testing under state-space dynamics, combining response-adaptive allocation with estimation of time-averaged treatment effects and anytime-valid confidence sequences.

\subsection{Nonstationarity and current-effect inference}
\label{sec:intro-current-effects}

Nonstationary bandit theory studies how temporal changes in reward distributions affect sequential decision-making, beginning with discounted and sliding-window methods for changing environments~\citep{GM11} and later extending to variation-budget formulations that characterize dynamic regret under more general temporal drift~\citep{BGZ14}. These ideas have also been developed for nonstationary contextual and linear bandits~\citep{CLLW19,RVC19,CSZ22}. A complementary line of work imposes local regularity through temporal smoothness, yielding sharper dynamic-regret guarantees for smooth reward trajectories~\citep{JXKF23} and minimax, smoothness-adaptive guarantees over H\"older classes~\citep{Suk25}. Nonstationarity and distribution shift have also been studied more broadly in stochastic optimization and robust  learning~\citep{BGZ15,SW25,SHW26,SSHW26}.

These models raise a distinct statistical challenge when the target is the \emph{current} treatment effect rather than a time-averaged effect over the history of the experiment. Under temporal drift, past observations remain
informative, but become less representative of current conditions as time passes. Using more historical data reduces sampling uncertainty but increases temporal bias, while using only recent data better reflects current conditions at the cost of statistical precision. Adaptive allocation creates an additional difficulty because arms with lower rewards may receive few recent observations
even though they remain necessary for inference. Reliable current-effect inference must therefore account for both the statistical value of recent observations and the regret incurred by collecting them. This leads to our central question:

\begin{quote}
\centering
\emph{What are the fundamental limits of simultaneously learning and performing valid inference in  nonstationary multi-armed bandits, can these limits be attained algorithmically, and to what extent can they be improved by exploiting local stationarity in the environment?}
\end{quote}

\subsection{Our contributions}
\label{sec:intro-contributions}

We formulate current-effect inference in nonstationary bandits as a constrained minimax problem and characterize the resulting regret--precision tradeoff. Our main contributions are:

\begin{enumerate}
\item \textbf{Optimal regret--precision tradeoff.}
We establish matching upper and lower bounds, up to logarithmic factors, for the minimum dynamic regret required for uniformly precise current-effect inference. The rate separates the usual cost of reward learning from an additional inferential cost of repeatedly sampling every arm as past observations become outdated. This cost persists even when the same arm is uniformly optimal, showing that exploration can be unnecessary for reward maximization yet unavoidable for current-effect inference.

\item \textbf{\textsc{TRACE} and adaptive exploration.}
We propose a policy called \textsc{TRACE} (\emph{Tracking Rewards and Current Effects}), which attains the minimax tradeoff up to logarithmic factors. An empirical-Bernstein e-process extension reduces continued exploration when predictable forecasts are accurate while preserving the same worst-case guarantee.

\item \textbf{Local asymptotic inference and efficiency.}
When realized drift is small relative to sampling error, all observations collected by \textsc{TRACE} can be reused for asymptotically normal current-effect inference. Under stronger local stationarity, the resulting estimator is locally asymptotically efficient.

\item \textbf{Limits of smoothness adaptation.}
We show that honest inference cannot generally adapt freely to unknown temporal smoothness under a regret constraint. Even for a constant trajectory, coverage over a rougher H\"older class can force wider intervals because a low-regret policy may not collect enough local information to rule out rougher alternatives, aligning with  classical impossibility results for honest adaptation over smoothness classes~\citep{Low97,Bul12}.

\end{enumerate}

\subsection{Organization}
\label{sec:intro-organization}
The remainder of this paper is organized as follows. In Section~\ref{sec:problem-setup}, we introduce the nonstationary bandit model and formulate current-effect inference under H\"older-smooth drift. In Section~\ref{sec:all-theoretical-results}, we develop our main theoretical results, characterizing the minimax regret--precision tradeoff (Section~\ref{sec:main-results}), introducing \textsc{TRACE} and its adaptive empirical-Bernstein extension (Sections~\ref{sec:algorithm} and \ref{sec:eprocess}), and establishing local asymptotic inference (Section~\ref{sec:local-inference}) and efficiency results together with limits on adaptation to unknown smoothness (Section~\ref{sec:unknown-smoothness}). In Section~\ref{sec:numerical-experiments}, we present numerical experiments illustrating the regret--inference tradeoff and the empirical performance of our proposed methods. Proofs of all main results and supporting lemmas are deferred to the appendices.

\subsection{Notation}
\label{sec:notation}

For $n\in\mathbb N_+$, let $[n]:=\{1,\ldots,n\}$. We write $\mathbf 1\{\cdot\}$ for the indicator function. For $x\in\mathbb R$, $\lfloor x\rfloor$ and $\lceil x\rceil$ denote the floor and ceiling of $x$, respectively. For $a,b\in\mathbb R$, let $a\vee b:=\max\{a,b\}$ and $a\wedge b:=\min\{a,b\}$. For an interval $I\subseteq\mathbb R$, $|I|$ denotes its length. We use $\P$, $\E$, and $\var$ for probability, expectation, and variance, respectively, with subscripts and superscripts used to indicate the underlying distribution or procedure when needed. We write $\Bern(p)$ for the Bernoulli distribution with mean $p$. For probability laws $\mathsf P$ and $\mathsf Q$, their total variation distance is $\mathsf{TV}(\mathsf P,\mathsf Q):=\sup_A|\mathsf P(A)-\mathsf Q(A)|$, and $\mathsf P\triangleleft \mathsf Q$ denotes that $\mathsf P$ is contiguous with respect to $\mathsf Q$. We write $X_n=o_P(r_n)$ and $X_n=\mathcal O_P(r_n)$ for the usual stochastic little-$o$ and big-$O$ relations, with the governing probability law indicated in the subscript when needed. For nonnegative sequences $a_n$ and $b_n$, we write $a_n\lesssim b_n$ if $a_n\le cb_n$ for a universal constant $c>0$, and $a_n\lesssim_\beta b_n$ if the constant may depend only on $\beta$. We write $a_n\asymp b_n$ when both $a_n\lesssim b_n$ and $b_n\lesssim a_n$ hold. Finally, $\widetilde{\mathcal O}(\cdot)$ and $\widetilde{\Theta}(\cdot)$ suppress logarithmic factors, with a subscript such as $\widetilde{\mathcal O}_\beta(\cdot)$ allowing constants depending only on $\beta$. All logarithms are natural unless stated otherwise.

\section{Current-effect bandit model}
\label{sec:problem-setup}

Consider a nonstationary $K$-armed bandit with action set $\cA=\{0,\ldots,K-1\}$ and horizon $T$. We will assign arm $0$ as a reference arm, which need not be optimal, and write $\cA_+:=\cA\setminus\{0\}$ for the remaining interventions. For $\beta\in(0,1]$ and $L\in[0,1]$, define the class of H\"older-smooth bandit environments
\begin{equation}
    \cM_{\beta,L}
    :=
    \left\{
        \mu=(\mu_a)_{a\in\cA}:
        \mu_a:[0,1]\to[0,1],\
        |\mu_a(u)-\mu_a(v)|
        \le
        L|u-v|^\beta
        \text{ for all }u,v\in[0,1],\ a\in\cA
    \right\}.
    \label{eq:holder-class}
\end{equation}
For $\mu\in\cM_{\beta,L}$, let $\mu_{t,a}:=\mu_a(t/T)$ denote the mean reward of arm $a$ at round $t$. The case $L=0$ recovers the stationary setting, while $\beta=1$ corresponds to Lipschitz  drift.

Let $\mathscr{F}_{t-1}=\sigma\left(A_s,Y_s,U_s:s\le t-1\right)$ denote the sigma-algebra generated by all past actions $A_s$, rewards $Y_s$, and revealed algorithmic randomness $U_s$ up to round $t-1$. The learner selects $A_t\in\cA$ using $\mathscr{F}_{t-1}$ and observes $Y_t\mid\mathscr{F}_{t-1},A_t\sim\Bern(\mu_{t,A_t})$. We use Bernoulli rewards for the minimax formulation and lower bounds, although the upper bounds continue to hold for any $Y_t\in[0,1]$ satisfying $\E[Y_t\mid\mathscr{F}_{t-1},A_t]=\mu_{t,A_t}$. We use $\P_\mu^\pi$ and $\E_\mu^\pi$ to denote probability and expectation under environment $\mu$ and procedure $\pi$. The expected dynamic regret of a procedure $\pi$ is then defined as
\begin{equation}
    \mathscr{R}_T(\pi,\mu)
    :=
    \E_\mu^\pi
    \left[
        \sum_{t=1}^T
        \bigl(
            \max_{a \in \mathcal{A}} \mu_{t, a} - \mu_{t, A_{t}}
        \bigr)
    \right].
    \label{eq:dynamic-regret}
\end{equation}
In addition to selecting actions, the learner reports inference for the current performance of each intervention relative to the reference arm. For $a\in\cA_+$, define the \emph{current effect} $\tau_t(a):=\mu_{t,a}-\mu_{t,0}$. Thus, $\tau_t(a)$ is the difference in expected reward between intervention $a$ and the reference intervention at round $t$. By \eqref{eq:holder-class}, $|\tau_t(a)-\tau_s(a)|
\le 2L|(t-s)/T|^\beta$, so older observations can become less representative of the current effect.

A procedure $\pi$ consists of an action-selection rule together with an inferential rule. Before observing $Y_t$, it reports a nonempty closed confidence interval $\mathscr{C}_t(a)\subseteq[-1,1]$ for the current effect $\tau_t(a)$ of each $a\in\cA_+$. We require $\mathscr{C}_t(a)$ to be $\mathscr{F}_{t-1}$-measurable, so the interval is determined before the reward at round $t$ is observed. Fix a predetermined burn-in time $t_0\in[T]$. We now formalize what it means for current-effect inference to be simultaneously valid and precise.

\begin{defn}[Validity and precision]
\label{def:validity}
Fix $w\in(0,2]$ and $\delta\in(0,1)$. A procedure $\pi$ is \emph{$(w,\delta)$-valid} if
\begin{equation}
    \inf_{\mu\in\cM_{\beta,L}}
    \P_\mu^\pi\!\left(
        \bigg\{
            \tau_t(a)\in \mathscr{C}_t(a)
            \ \forall(t,a)\in[T]\times\cA_+
        \bigg\}
        \cap
        \left\{
            \max_{t_0\le t\le T}
            \max_{a\in\cA_+}
            |\mathscr{C}_{t}(a)|
            \le w
        \right\}
    \right)
    \ge 1-\delta .
    \label{eq:validity}
\end{equation}
\end{defn}

\begin{rem}
\label{rmk:burn-in-and-validity}
Coverage is required over the full horizon, while precision is only required after a finite burn-in time $t_0$. Burn-in time is unavoidable as before any rewards are observed, a learner will not have sufficient information to distinguish bandit settings with different current effects.  In our main results, we take $t_0=\lceil T/2\rceil$. Simultaneous coverage also implies validity at any data-dependent round and arm
within the prescribed horizon. In particular, for any such $\nu\in[T]$ and $\widehat a\in\cA_+$, $\P_\mu^\pi\{\tau_\nu(\widehat a)\notin \mathscr{C}_\nu(\widehat a)\}\le\delta$.
\end{rem}

\noindent We measure the cost of this inferential requirement through the constrained minimax regret
\begin{equation}
    \mathscr{R}_T^\star(w,\delta)
    :=
    \inf_{\pi:\,(w,\delta)\text{-valid}}
    \sup_{\mu\in\cM_{\beta,L}}
    \mathscr{R}_T(\pi,\mu).
    \label{eq:minimax-regret}
\end{equation}
We set $\mathscr{R}_T^\star(w,\delta)=+\infty$ if no $(w,\delta)$-valid procedure exists. Dependence on $K$, $\beta$, and $L$ is suppressed in the notation. Throughout
the main results, the learner has full information on $T$, $K$, $\beta$, $L$, $w$, and $\delta$.

\section{Theoretical results}
\label{sec:all-theoretical-results}

We now present our main theoretical results. We first characterize the minimax regret--precision tradeoff for current-effect inference (Section \ref{sec:main-results}). We then introduce \textsc{TRACE}, which attains this tradeoff up to logarithmic factors (Section \ref{sec:algorithm}), followed by an empirical-Bernstein extension that adapts inferential exploration to realized prediction errors while preserving the worst-case guarantee (Section \ref{sec:eprocess}). We next develop a local asymptotic theory showing that, when the realized drift is sufficiently small, observations collected for reward learning can also be used for asymptotically normal and locally efficient inference (Section \ref{sec:local-inference}). Finally, we study the limits of honest adaptation to unknown temporal smoothness under a regret constraint (Section \ref{sec:unknown-smoothness}).

\subsection{The regret--precision tradeoff}
\label{sec:main-results}

We now characterize the minimum regret required to maintain uniformly precise inference for the current effects. Throughout this section, we take $t_0=\lceil T/2\rceil$ and assume $2\le K\le T$. We focus on the feasible regime in which $Kw^{-2}[1+(L/w)^{1/\beta}]
\log \bigl(K[1+(L/w)^{1/\beta}]/\delta\bigr) \le c_\beta T$ for a sufficiently small constant $c_\beta>0$ depending only on $\beta$.

\begin{thm}[Regret--precision tradeoff]
\label{thm:tradeoff}
There exist universal constants $w_0,\delta_0\in(0,1)$ such that, for every
$w\in(0,w_0]$ and $\delta\in(0,\delta_0]$ in the feasible regime,
\begin{equation}
    \mathscr{R}_T^\star(w,\delta)
    =
    \widetilde{\Theta}_{\beta}\left(
        \sqrt{KT}
        +
        K^{\beta/(2\beta+1)}
        L^{1/(2\beta+1)}
        T^{(\beta+1)/(2\beta+1)}
        +
        \frac{K}{w^2}
        \left[
            1+\left(\frac{L}{w}\right)^{1/\beta}
        \right]
    \right).
    \label{eq:tradeoff}
\end{equation}
\end{thm}

\begin{proof}
We prove Theorem~\ref{thm:tradeoff} in Appendix~\ref{appendix:proof-tradeoff}.
\end{proof}

\begin{figure}[t]
    \centering
    \includegraphics[width=0.57\linewidth]{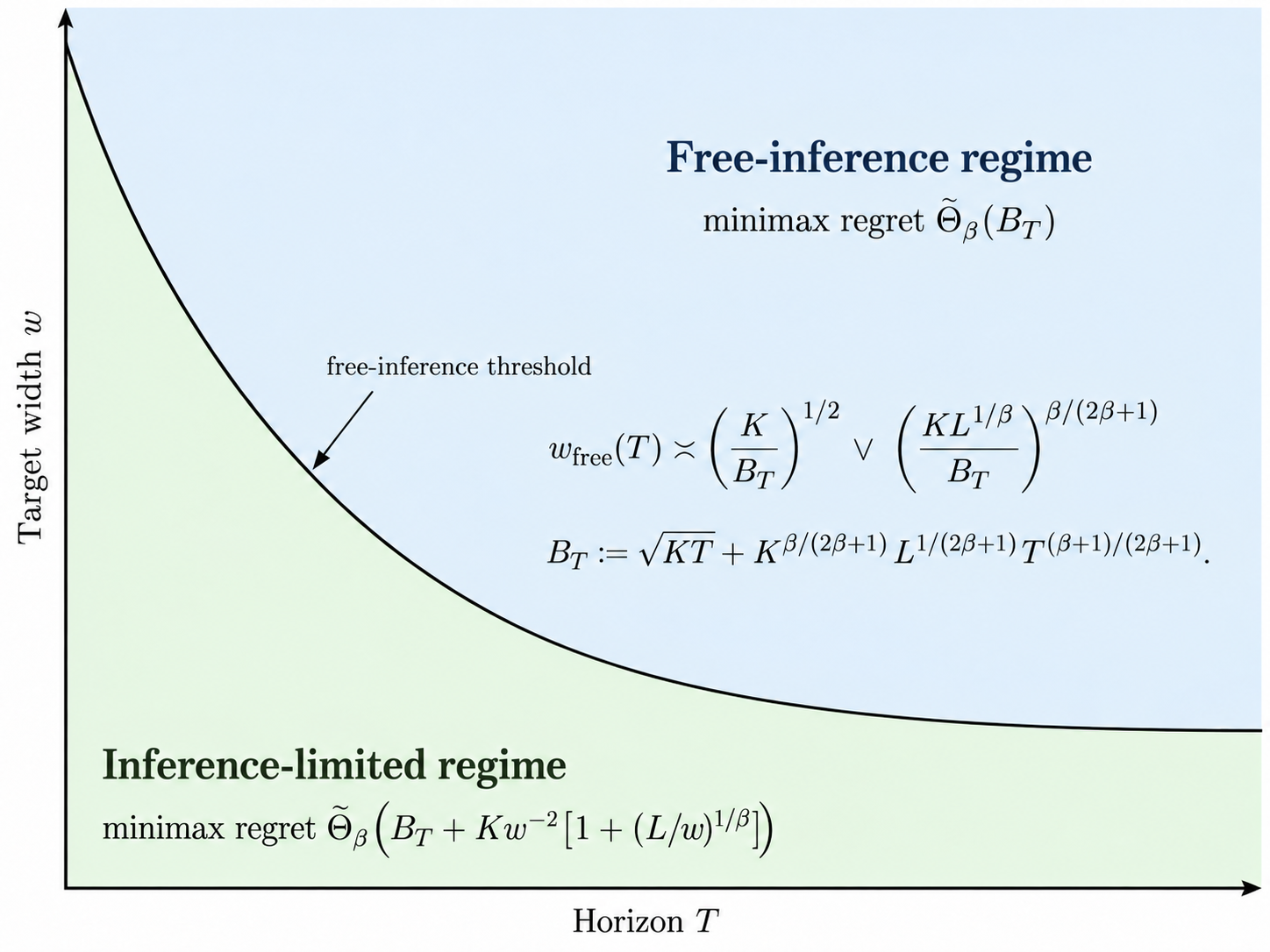}
    \caption{\textbf{Minimax regret--precision tradeoff.} Narrower confidence intervals require more frequent exploration and therefore increase regret. The free-inference threshold $w_{\mathrm{free}}(T)$ separates the inference-limited regime from the regime where current-effect inference is essentially free at the minimax regret scale.}
    \label{fig:minimax-rate}
\end{figure}

The first two terms in \eqref{eq:tradeoff} are the standard cost of reward learning under smooth nonstationarity~\citep{JXKF23,Suk25}. The final term is the additional cost of current-effect inference. More precisely, this term carries the logarithmic factor $\log(K[1+(L/w)^{1/\beta}]/\delta)$, which is suppressed by $\widetilde{\Theta}_{\beta}$ and shown by the lower bound to be necessary. The factor $1+(L/w)^{1/\beta}$ reflects how often information must be refreshed. Observations more than roughly $T(w/L)^{1/\beta}$ rounds old may incur temporal bias larger than the target precision, so the horizon contains roughly $1 +(L/w)^{1/\beta}$ regions in which fresh information is needed. Each region requires order $w^{-2}$ observations from each arm. This cost persists even when the same arm remains optimal throughout the experiment. One can choose $w$ to balance the costs of reward learning and inference, with this balance achievable at increasingly small widths as $T$ grows relative to $K$ and $L$ (see Figure~\ref{fig:minimax-rate}). 

Two special cases of Theorem~\ref{thm:tradeoff} are particularly informative. We first consider Lipschitz drift.

\begin{cor}[Lipschitz drift]
\label{cor:lipschitz}
Suppose $\beta=L=1$ and $K,\delta$ are fixed. In the feasible regime, $\mathscr{R}_T^\star(w,\delta)\asymp T^{2/3}+w^{-3}\log(1/w)$, up to constants depending only on $K$ and $\delta$. In particular, if $w\asymp T^{-2/9}(\log T)^{1/3}$, then $\mathscr{R}_T^\star(w,\delta)\asymp T^{2/3}$. The smallest feasible width satisfies $w\asymp T^{-1/3}(\log T)^{1/3}$, in which case $\mathscr{R}_T^\star(w,\delta)\asymp T$.
\end{cor}

\begin{proof}
Substituting $\beta=L=1$ into Theorem~\ref{thm:tradeoff}, with $K$ and $\delta$ fixed, gives $\mathscr{R}_T^\star(w,\delta)\asymp T^{2/3}+w^{-3}\log(1/w)$. Balancing the inferential term with $T^{2/3}$ yields $w\asymp T^{-2/9}(\log T)^{1/3}$. The feasibility condition requires $w^{-3}\log(1/w)\lesssim T$, whose smallest solution satisfies $w\asymp T^{-1/3}(\log T)^{1/3}$.
\end{proof}

In the stationary case i.e. $L=0$, the refresh cost disappears while the inferential cost remains.

\begin{cor}[Stationary environments]
\label{cor:stationary}
Suppose $L=0$ and $K,\delta$ are fixed. In the feasible regime, $\mathscr{R}_T^\star(w,\delta)\asymp T^{1/2}+w^{-2}$, up to constants depending only on $K$ and $\delta$. Thus, even in a stationary environment,
uniformly precise current-effect inference incurs an additional $w^{-2}$ regret cost beyond standard reward learning.
\end{cor}

\begin{proof}
When $L=0$, the nonstationary reward-learning term in
Theorem~\ref{thm:tradeoff} vanishes and $1+(L/w)^{1/\beta}=1$. Since $K$ and $\delta$ are fixed, the remaining logarithmic factor is constant, giving $\mathscr{R}_T^\star(w,\delta)\asymp T^{1/2}+w^{-2}$.
\end{proof}

\subsection{\textsc{TRACE}: Tracking Rewards and Current Effects}
\label{sec:algorithm}

We next introduce our proposed \textsc{TRACE} policy (implemented in Algorithm~\ref{alg:trace}), which achieves the upper bound in Theorem~\ref{thm:tradeoff}. \textsc{TRACE} uses separate sampling schedules for reward learning and current-effect inference. Within each inference block, it reserves observations from every arm to estimate the current effects. On the remaining rounds, it runs a restarted instance of \textsc{EXP3}. \textsc{EXP3} was introduced by \citet{ACBFS02} for adversarial multi-armed bandits and controls regret relative to the best fixed arm without requiring stationary rewards within a block (cf. Section 11.3 of \cite{LS20}).

Set
\begin{equation}
    J
    :=
    \max\left\{
        4,\,
        \left\lceil
            4\left(\frac{8L}{w}\right)^{1/\beta}
        \right\rceil
    \right\},
    \;\;
    m
    :=
    \left\lceil
        \frac{32}{w^2}
        \log\left(\frac{2KJ}{\delta}\right)
    \right\rceil.
    \label{eq:inference-tuning}
\end{equation}
Here $J$ is the number of inference blocks and $m$ is the number of reserved observations from each arm in each block. To handle nonstationarity in the mean reward $\mu \in \mathcal{M}_{\beta, L}$, we restart \textsc{EXP3} every $h$ rounds by resetting all arm weights to one, which induces a uniform distribution over $\mathcal A$:
\begin{equation}
    h
    :=
    \begin{cases}
        T
        & L=0,\\[1mm]
        \displaystyle
        \displaystyle
        T
        \wedge
        \left\lceil
            \left(
                \frac{T^\beta\sqrt{K\log K}}{L}
            \right)^{2/(2\beta+1)}
        \right\rceil
        & L>0,
    \end{cases}
    \label{eq:reward-restart}
\end{equation}
and let $\eta:=\sqrt{2\log K/(Kh)}$ be the stepsize used by \textsc{EXP3}.

Partition $[T]$ into $J$ consecutive blocks $\mathcal B_1,\ldots,\mathcal B_J$ whose lengths differ by at most one, and let $b_j:=\min\mathcal B_j$. In each block $\mathcal B_j$, reserve the first $Km$ rounds for statistical inference, sampling the $K$ arms in round-robin order so that each arm is sampled exactly $m$ times. Denote these reserved rounds by $\mathcal I_j$. Let $\mathcal G_j:=\mathcal B_j\setminus\mathcal I_j$ denote the remaining reward-learning rounds. On these rounds, the policy learns a randomized action-selection rule aimed at maximizing cumulative reward and minimizing regret.

We further denote the set of all reserved rounds by
$\mathcal I:=\bigcup_{j\leq J}\mathcal I_j$ and the set of all non-reserved rounds by $\mathcal G:=\bigcup_{j\leq J}\mathcal G_j$. We also partition $[T]$ into $R:=\lceil T/h\rceil$ consecutive reward-restart blocks $\mathcal H_1,\ldots,\mathcal H_R$, each of size $h$ except possibly the last.

For $a \in \cA$, let $\mathcal{S}_{j,a} \subseteq \mathcal{I}_{j}$ denote the $m$ reserved rounds
assigned to arm $a$. Define $\widehat\mu_{j,a}:=m^{-1}\sum_{s\in\mathcal S_{j,a}}Y_s$ and $\widehat\tau_j(a):=\widehat\mu_{j,a}-\widehat\mu_{j,0}$ for $a\in\cA_+$. For $t\in\mathcal B_j$ with $j\ge2$, \textsc{TRACE} provides a confidence interval of the form
\begin{equation}
    \mathscr{C}_{t}(a)
    :=
    \left[
        \widehat\tau_{j-1}(a)
        \pm
        \left\{
            \sqrt{
                \frac{2}{m}
                \log\left(\frac{2KJ}{\delta}\right)
            }
            +
            2L\left(\frac{4}{J}\right)^\beta
        \right\}
    \right]
    \cap[-1,1],
    \;\;
    a\in\cA_+.
    \label{eq:confidence-interval}
\end{equation}
During the first inference block, set $\mathscr{C}_{t}(a)=[-1,1]$. Since \eqref{eq:confidence-interval} uses only observations from the preceding completed block, each interval is $\mathscr{F}_{t-1}$-measurable. The choice of $J$ controls the temporal bias between consecutive inference blocks, while $m$ controls the sampling error in the estimated arm means. 

\begin{figure}[t]
    \centering
    \includegraphics[width=0.95\linewidth]{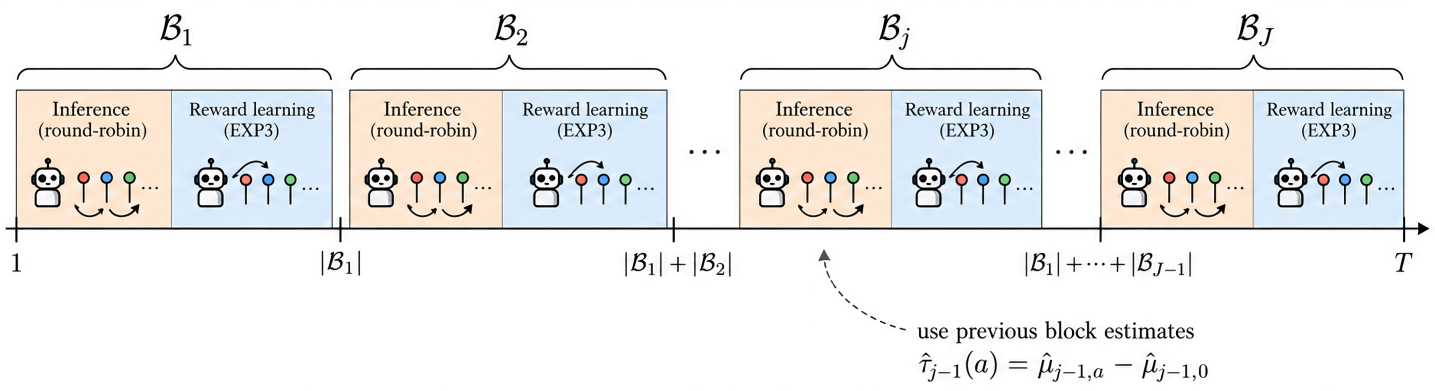}
    \caption{\textbf{Overview of \textsc{TRACE}.} The horizon is partitioned into consecutive blocks $\mathcal{B}_1,\ldots,\mathcal{B}_J$. At the beginning of each block, \textsc{TRACE} samples all arms in round-robin order to collect fresh observations for current-effect inference. On the remaining rounds, it runs EXP3 for reward learning. Inference in block $\mathcal{B}_j$, $j\ge 2$, uses estimates constructed from the preceding block, allowing the procedure to maintain up-to-date confidence intervals while continuing to optimize reward.}
    \label{fig:trace-alg-diagram}
\end{figure}

Within each reward-restart block $\mathcal H_r$, \textsc{EXP3} maintains a nonnegative weight $v_a$ for each arm $a\in\cA$, initialized as $v_a=1$ at the beginning of the block. On each non-reserved round $t\in\mathcal H_r\cap\mathcal G$, the weights induce the sampling probabilities
$p_t(a):=v_a/\sum_{b\in\cA}v_b$.
After observing $Y_t$, \textsc{TRACE} forms the importance-weighted loss estimate
$\widehat\ell_t(a):=\mathbf{1}\left\{A_t=a\right\}(1-Y_t)/p_t(a)$ and updates $v_a\leftarrow v_a\exp{-\eta\widehat\ell_t(a)}$. Thus, $v_a$ is the exponential weight associated with the cumulative estimated loss of arm $a$ since the beginning of the current reward-restart block. On reserved rounds $t\in\mathcal H_r\cap\mathcal I$, the \textsc{EXP3} weights remain unchanged. The restart period $h$ balances the fixed-arm regret of \textsc{EXP3} with the error from comparing a fixed arm against the changing dynamic oracle. We now show TRACE attains the minimax optimal rate as in Theorem~\ref{thm:tradeoff}.

\vspace{1em}

\begin{thm}[Upper bound for \textsc{TRACE}]
\label{thm:upper-bound}
Suppose $KJm\le T/2$. Then \textsc{TRACE} is $(w,\delta)$-valid and, after the first inference block, $|\mathscr{C}_{t}(a)|\le w$ for every $a\in\cA_+$ on every sample
path. Moreover,
\begin{equation}
\begin{aligned}
    \mathscr{R}_T(\pi,\mu)
    \le\;&
    KJm
    +
    2\sqrt{K\log K}
    \left(
        \frac{T}{\sqrt h}+\sqrt h
    \right)
    +
    2LT\left(\frac{h}{T}\right)^\beta
    \\
    \lesssim_\beta\;&
    KJm
    +
    \sqrt{KT\log K}
    +
    (K\log K)^{\beta/(2\beta+1)}
    L^{1/(2\beta+1)}
    T^{(\beta+1)/(2\beta+1)}.
    \label{eq:upper-bound}
\end{aligned}
\end{equation}
\end{thm}

\begin{proof}
We prove Theorem~\ref{thm:upper-bound} in Appendix~\ref{appendix-sec:proof-of-nonadaptive-upper-bound}.
\end{proof}

Since $J\asymp_\beta 1+(L/w)^{1/\beta}$ and
$m\asymp w^{-2}\log(KJ/\delta)$, the regret incurred by the reserved rounds is
$\widetilde{\mathcal O}_\beta
\left(Kw^{-2}[1+(L/w)^{1/\beta}]\right)$.
Together with Theorem~\ref{thm:upper-bound}, this establishes the upper bound in
Theorem~\ref{thm:tradeoff}.

\vspace{1em}

\begin{breakablealgorithm}
\caption{\textsc{TRACE}: Tracking Rewards and Current Effects}
\label{alg:trace}
\begin{algorithmic}[1]
\State \textbf{Input:} $T,K,\beta,L,w,\delta$
\State Set $J$ and $m$ according to \eqref{eq:inference-tuning}
\State Set $h$ according to \eqref{eq:reward-restart} and
$\eta\gets\sqrt{2\log K/(Kh)}$
\State Partition $[T]$ into $\mathcal B_1,\ldots,\mathcal B_J$
\State Set $b_j\gets\min\mathcal B_j$ for every $j\in[J]$
\State Initialize the \textsc{EXP3} weights $v_a\gets1$ for every $a\in\cA$

\For{$t=1,\ldots,T$}
    \State Let $j$ be the unique index such that $t\in\mathcal B_j$

    \If{$j=1$}
        \State Set $\mathscr{C}_{t}(a)\gets[-1,1]$ for every $a\in\cA_+$
    \Else
        \State Set $\mathscr{C}_{t}(a)$ according to \eqref{eq:confidence-interval}
        for every $a\in\cA_+$
    \EndIf

    \If{$(t-1)\bmod h=0$}
        \State Restart \textsc{EXP3} by setting $v_a\gets1$ for every $a\in\cA$
    \EndIf

    \If{$0\le t-b_j<Km$}
        \State Set $A_t\gets(t-b_j)\bmod K$
        \State Observe $Y_t$ and add $t$ to $\mathcal S_{j,A_t}$
        \State Leave the \textsc{EXP3} weights $v_a$ unchanged
    \Else
        \State Set \label{eq:exp3-induced}
        $p_t(a)\gets v_a/\sum_{b\in\cA}v_b$
        for every $a\in\cA$
        \State Draw $A_t\sim p_t$ and observe $Y_t$
        \For{$a\in\cA$}
            \State Set the importance-weighted loss estimate
            $\widehat\ell_t(a)
            \gets
            \mathbf 1\{A_t=a\}(1-Y_t)/p_t(a)$
            \State Update the \textsc{EXP3} weight
            $v_a\gets v_a\exp\{-\eta\widehat\ell_t(a)\}$
        \EndFor
    \EndIf

    \If{$t=b_j+Km-1$}
        \For{$a\in\cA$}
            \State Set
            $\widehat\mu_{j,a}
            \gets
            m^{-1}\sum_{s\in\mathcal S_{j,a}}Y_s$
        \EndFor
        \For{$a\in\cA_+$}
            \State Set
            $\widehat\tau_j(a)
            \gets
            \widehat\mu_{j,a}-\widehat\mu_{j,0}$
        \EndFor
    \EndIf
\EndFor
\end{algorithmic}
\end{breakablealgorithm}

\subsection{Adaptive \textsc{TRACE} with empirical-Bernstein e-processes}
\label{sec:eprocess}

The fixed version of \textsc{TRACE} reserves exactly $m$ observations from every arm in each inference block. This choice is minimax optimal, but it does not adapt to the realized difficulty of the problem and data. We now replace the fixed inferential sampling requirement with a data-dependent stopping rule (Algorithm~\ref{alg:adaptive-trace}). Each arm is sampled until its confidence sequence becomes sufficiently narrow, after which continued exploration of that arm stops.

Our construction relies on an empirical-Bernstein e-process together with a Hoeffding e-process. The empirical-Bernstein bound adapts to the realized
prediction errors and can terminate exploration early, while the Hoeffding bound guarantees that exploration stops after the worst-case number of samples. The resulting procedure retains the worst-case guarantee of \textsc{TRACE} while using fewer inferential samples on favorable sample paths. We use standard empirical-Bernstein and betting constructions from
\citet{HRMS21,WR24}, together with the e-process framework of \citet{RGVS23}.

For this adaptive version, set
\begin{equation}
    J
    :=
    \max\left\{
        4,\,
        \left\lceil
            4\left(\frac{32L}{w}\right)^{1/\beta}
        \right\rceil
    \right\},
    \;\;
    m
    :=
    \left\lceil
        \frac{128}{w^2}
        \log\left(\frac{4KJ}{\delta}\right)
    \right\rceil,
    \label{eq:adaptive-tuning}
\end{equation}
and let $\alpha:=\delta/(4KJ)$. We retain the reward-learning restart period $h$ from \eqref{eq:reward-restart} and the learning rate $\eta=\sqrt{2\log K/(Kh)}$. Note that $m$ only specifies the worst-case number of observations required for inference and plays no algorithmic role in the adaptive setting (see Corollary~\ref{cor:prediction-error}).

\paragraph{Blockwise confidence sequences.}
Fix an inference block $\mathcal B_j$ and let
$b_j:=\min\mathcal B_j$. Rather than estimating the entire time-varying mean trajectory, we construct a confidence sequence for the arm mean $\mu_{b_j,a}$ at a reference round $b_j$. H\"older continuity then bounds how much the mean can change between $b_j$ and the current round, allowing us to enlarge the confidence interval accordingly to obtain a valid confidence set for the current mean.

Let $F_t\in\{0,1\}$ indicate whether round $t$ is used for continued exploration. The decision whether $F_t=1$ is made before observing $Y_t$, so $F_t$ is $\mathscr{F}_{t-1}$-measurable. For $t\in\mathcal B_j$ and $a\in\cA$, define
\begin{equation}
\begin{aligned}
    N_{j,a}(t)
    :=
    \sum_{s=b_j}^t
    F_s\mathbf 1\{A_s=a\}, \;\;
    V_{j,a}(t)
    :=
    \sum_{s=b_j}^t
    F_s\mathbf 1\{A_s=a\}
    \bigl(Y_s-\widehat Y_{s,a}\bigr)^2,
\end{aligned}
\label{eq:adaptive-NV}
\end{equation}
where $\widehat Y_{s,a}\in[0,1]$ is any $\mathscr{F}_{s-1}$-measurable prediction of the reward from arm $a$. When $N_{j,a}(t)>0$, let $\overline Y_{j,a}(t)$ denote the average of the corresponding observations. Both $N_{j,a}(t)$ and $V_{j,a}(t)$ reset at the beginning of each
inference block. Since every block has length at most $2T/J$, H\"older continuity gives $|\mu_{s,a}-\mu_{b_j,a}|\le L(2/J)^\beta$ for every $s\in\mathcal B_j$. For a candidate value $\theta\in[0,1]$ and $\lambda\in[0,1)$, define
\begin{equation}
\begin{aligned}
    E_{j,a,t}^{\pm}(\theta;\lambda)
    :=
    \exp\Bigg\{
        &\lambda
        \sum_{s=b_j}^t
        F_s\mathbf 1\{A_s=a\}
        \left[
            \pm(Y_s-\theta)-d_J
        \right]
        -
        \psi(\lambda)V_{j,a}(t)
    \Bigg\},
    \label{eq:eb-process}
\end{aligned}
\end{equation}
where $\psi(\lambda):=-\log(1-\lambda)-\lambda$ is the sub-exponential CGF function and $d_J:=L(2/J)^\beta$. Under the null $\mu_{b_j,a}=\theta$, each process in \eqref{eq:eb-process} is a nonnegative supermartingale starting from one. Importantly, this remains true under adaptive action selection and data-dependent exploration because the action and exploration decisions are predictable. We use a geometric mixture over $\lambda_k:=2^{-k-1}$, $k\ge0$. We define the sub-exponential uniform boundary $B_{\alpha}(v)$ via its level-$\alpha$ sublevel sets as follows:
\begin{equation}
    B_\alpha(v)
    :=
    \inf\left\{
        x\ge0:
        \sum_{k=0}^{\infty}
        \frac{
            \exp\{
                \lambda_k x-\psi(\lambda_k)v
            \}
        }{
            (k+1)(k+2)
        }
        \ge
        \frac{1}{\alpha}
    \right\}.
    \label{eq:mixture-boundary}
\end{equation}
The weights $[(k+1)(k+2)]^{-1}$ sum to one, so the corresponding mixture is again an e-process. By Theorems~1 and~4 of \citet{HRMS21}, it can be shown that
$
    B_\alpha(v)
    \lesssim
    \sqrt{
        v\left[
            \log(1/\alpha)+\log\log(e+v)
        \right]
    }
    +
    \log(1/\alpha)
    +
    \log\log(e+v) 
$. 
For $N_{j,a}(t)>0$, define the base confidence interval
\begin{equation}
\begin{aligned}
    I_{j,a,t}
    :=
    \Bigg[
        \overline Y_{j,a}(t)
        \pm
        \Bigg\{
            d_J
            +
            \frac{
                B_\alpha(V_{j,a}(t))
            }{
                N_{j,a}(t)
            }
            \wedge
            \left(
                \frac{w}{32}
                +
                \frac{
                    4\log(1/\alpha)
                }{
                    wN_{j,a}(t)
                }
            \right)
        \Bigg\}
    \Bigg]
    \cap[0,1].
    \label{eq:adaptive-reference-ci}
\end{aligned}
\end{equation}
The first statistical term in \eqref{eq:adaptive-reference-ci} is the empirical-Bernstein radius. The second follows from a fixed-parameter Hoeffding e-process with parameter $w/4$. Both bounds hold simultaneously under the allocation $\alpha=\delta/(4KJ)$.

For $t\in\mathcal B_j$, define the running intersection $\widetilde D_{j,a,t} := \bigcap_{s\in\mathcal B_j:\, s\le t,\; N_{j,a}(s)>0} I_{j,a,s}$ with $D_{j,a,t}=[0,1]$ before the first observation from arm $a$.
Set $D_{j,a,t}:=\widetilde D_{j,a,t}$ whenever $\widetilde D_{j,a,t}\neq\varnothing$, and set $D_{j,a,t}:=\{0\}$ otherwise. We next show that these running intersections form simultaneously valid confidence sequences for the block-start means, even under adaptive action selection and arbitrary predictable reward predictions.

\begin{lem}[Blockwise e-process validity]
\label{lem:eprocess-validity}
With probability at least $1-\delta$, simultaneously for every $j\in[J]$, $a\in\cA$, and $t\in\mathcal B_j$,
$\mu_{b_j,a}\in D_{j,a,t}$. The statement holds for arbitrary predictable predictions $\widehat Y_{t,a}\in[0,1]$ and under adaptive action selection.
\end{lem}

\begin{proof}
We prove Lemma~\ref{lem:eprocess-validity} in Appendix~\ref{appendix-sec:proof-of-eprocess-validity}.
\end{proof}

On the event in Lemma~\ref{lem:eprocess-validity}, $\mu_{b_{j}, a} \in I_{j, a, t}$ so $\mu_{b_{j}, a} \in \bigcap_{s\in\mathcal B_j:\, s\le t,\; N_{j,a}(s)>0} I_{j,a,s} := D_{j, a, t}$. Hence $D_{j, a, t} \neq \varnothing$. Lemma~\ref{lem:eprocess-validity} also allows us to stop exploration at a data-dependent time without an additional correction.

\paragraph{Data-dependent exploration.}
At the beginning of block $\mathcal B_j$, every arm is active. While active arms remain, \textsc{TRACE} samples them in a fixed cyclic order. Exploration of arm
$a$ stops as soon as $|D_{j,a,t}|\le w/4$, at which point we save $D_{j,a}^\star:=D_{j,a,t}$ and remove $a$ from the active set. Once all arms have stopped, the restarted \textsc{EXP3} learner selects the remaining actions in the block. For each saved interval, define its endpoints by $\ell_{j,a}^\star:=\inf_{x\in D_{j,a}^\star}x$ and $u_{j,a}^\star:=\sup_{x\in D_{j,a}^\star}x$. For $t\in\mathcal B_j$, $j\ge2$, and $a\in\cA_+$, TRACE provides a confidence interval
\begin{equation}
    \mathscr{C}_{t}(a)
    :=
    \left[
        \ell_{j-1,a}^\star-u_{j-1,0}^\star
        -2L\left(\frac{t-b_{j-1}}{T}\right)^\beta,\;
        u_{j-1,a}^\star-\ell_{j-1,0}^\star
        +2L\left(\frac{t-b_{j-1}}{T}\right)^\beta
    \right]
    \cap[-1,1].
    \label{eq:adaptive-current-ci}
\end{equation}
During the first inference block, set $\mathscr{C}_{t}(a)=[-1,1]$. Since the saved intervals from block $j-1$ are determined before block $j$ begins, $\mathscr{C}_{t}(a)$ is $\mathscr{F}_{t-1}$-measurable. Using Lemma~\ref{lem:eprocess-validity}, one can show that adaptive \textsc{TRACE} attains the minimax optimal regret guarantee of Theorem~\ref{thm:tradeoff}.

\begin{thm}[Adaptive exploration with the \textsc{TRACE} guarantee]
\label{thm:adaptive-trace}
Suppose $KJm\le T/2$. Then Algorithm~\ref{alg:adaptive-trace} is
$(w,\delta)$-valid and, after the first inference block,
$|\mathscr{C}_{t}(a)|\le w$ for every $a\in\cA_+$ on every sample path. Moreover, each arm requires at most $m$ forced observations per block, and
\begin{equation}
    \sup_{\mu\in\cM_{\beta,L}}
    \mathscr{R}_T(\pi,\mu)
    =
    \widetilde{\Theta}_{\beta}\left(
        \sqrt{KT}
        +
        K^{\beta/(2\beta+1)}
        L^{1/(2\beta+1)}
        T^{(\beta+1)/(2\beta+1)}
        +
        \frac{K}{w^2}
        \left[
            1+\left(\frac{L}{w}\right)^{1/\beta}
        \right]
    \right).
    \label{eq:adaptive-worst-case-regret}
\end{equation}
\end{thm}

\begin{proof}
We prove Theorem~\ref{thm:adaptive-trace} in Appendix~\ref{appendix-sec:proof-of-adaptive-trace}.
\end{proof}

The empirical-Bernstein construction becomes beneficial when the prediction error $Y_{s} - \hat{Y}_{s, a}$ is sufficiently small as $V_{j, a}(t)$ depends directly on this quantity and can lead to substantially shorter exploration phases. To quantify this improvement, let $n_{j,a}$ denote the number of continued observations from arm $a$ in block $j$.

\begin{cor}[Prediction-error-dependent exploration]
\label{cor:prediction-error}
Fix a block $j$ and arm $a$. Suppose that, along a sample path, $V_{j,a}(t)\le vN_{j,a}(t)+1$ at every attained sample count before stopping, for some $v\in[0,1]$. Then
\begin{equation}
    n_{j,a}
    \lesssim
    m 
    \wedge
    \left(
        \frac{v}{w^2}+\frac{1}{w}
    \right)
    \left[
        \log\left(\frac{4KJ}{\delta}\right)
        +
        \log\log(e+m)
    \right].
    \label{eq:prediction-dependent-count}
\end{equation}
In particular, exploration stops by any attained sample count satisfying $B_\alpha(V_{j,a}(t))\le wN_{j,a}(t)/16$.
\end{cor}

\begin{proof}
We prove Corollary~\ref{cor:prediction-error} in Appendix~\ref{appendix-sec:proof-of-trace-prediction-error}.
\end{proof}

Up to logarithmic factors, Corollary~\ref{cor:prediction-error} gives $n_{j,a}=\widetilde{\mathcal O}(v/w^2+1/w)$. This interpolates between the worst-case $\widetilde{\mathcal O}(w^{-2})$ sample requirement when
$v=\Theta(1)$ and the faster $\widetilde{\mathcal O}(w^{-1})$ rate when $v\lesssim w$. In particular, uniformly bounded accumulated squared residuals lead to the latter regime.  Since $\widehat Y_{s,a}$ is predictable, $\E[(Y_s-\widehat Y_{s,a})^2\mid\mathscr{F}_{s-1},A_s=a] =\var(Y_s\mid\mathscr{F}_{s-1},A_s=a) +(\mu_{s,a}-\widehat Y_{s,a})^2$. Thus, accurate predictions reduce the contribution from prediction error, while intrinsic reward noise remains even for an ideal predictor. 

Prior methods typically treat the sampling design as given, impose a predetermined amount of inferential
randomization, or use predictive models to improve the precision of the resulting confidence sequence~\citep{LB24,WR24}. In contrast, adaptive \textsc{TRACE} uses the confidence sequence itself to determine how much exploration each arm requires based on predictive uncertainty. 

This feature is particularly useful in online experimentation systems that must simultaneously optimize an objective of interest and maintain up-to-date estimates of treatment effects. Such systems often already maintain forecasting models for various latent factors, and these predictions can be provided directly to adaptive \textsc{TRACE}. When the forecasts are accurate, fewer observations need to be diverted from reward maximization for inference. When prediction quality deteriorates, the procedure automatically reverts toward the worst-case exploration budget while preserving coverage validity.

\vspace{1em}

\begin{breakablealgorithm}
\caption{Adaptive \textsc{TRACE} with empirical-Bernstein e-processes}
\label{alg:adaptive-trace}
\begin{algorithmic}[1]
\State \textbf{Input:} $T,K,\beta,L,w,\delta$ and predictable predictions
$\widehat Y_{t,a}\in[0,1]$
\State Set $J$ and $m$ according to \eqref{eq:adaptive-tuning}
\State Set $\alpha\gets\delta/(4KJ)$
\State Set $h$ according to \eqref{eq:reward-restart} and
$\eta\gets\sqrt{2\log K/(Kh)}$
\State Partition $[T]$ into $\mathcal B_1,\ldots,\mathcal B_J$
\State Initialize the \textsc{EXP3} weights $v_a\gets1$ for every $a\in\cA$

\For{$j=1,\ldots,J$}
    \State Set $b_j\gets\min\mathcal B_j$
    \State Set $\cA_j^{\mathrm{act}}\gets\cA$
    \State Set $N_{j,a}\gets0$, $V_{j,a}\gets0$, and
    $D_{j,a}\gets[0,1]$ for every $a\in\cA$

    \For{$t\in\mathcal B_j$}
        \If{$j=1$}
            \State Set $\mathscr{C}_{t}(a)\gets[-1,1]$ for every $a\in\cA_+$
        \Else
            \State Set $\mathscr{C}_{t}(a)$ according to
            \eqref{eq:adaptive-current-ci} for every $a\in\cA_+$
        \EndIf

        \If{$(t-1)\bmod h=0$}
            \State Restart \textsc{EXP3} by setting $v_a\gets1$ for every $a\in\cA$
        \EndIf

        \If{$\cA_j^{\mathrm{act}}\neq\varnothing$}
            \State Set $F_t\gets1$
            \State Choose $A_t$ from $\cA_j^{\mathrm{act}}$ in a fixed cyclic order
            \State Observe $Y_t$
            \State Update $N_{j,A_t}\gets N_{j,A_t}+1$
            \State Update
            $V_{j,A_t}\gets V_{j,A_t}+(Y_t-\widehat Y_{t,A_t})^2$
            \State Update the sample mean $\overline Y_{j,A_t}$
            \State Form $I_{j,A_t,t}$ according to
            \eqref{eq:adaptive-reference-ci}
            \State Update
            $D_{j,A_t}\gets D_{j,A_t}\cap I_{j,A_t,t}$

            \If{$D_{j,A_t}=\varnothing$}
                \State Set $D_{j,A_t}\gets\{0\}$
            \EndIf

            \If{$|D_{j,A_t}|\le w/4$}
                \State Save $D_{j,A_t}^\star\gets D_{j,A_t}$
                \State Remove $A_t$ from $\cA_j^{\mathrm{act}}$
            \EndIf
            \State Leave the \textsc{EXP3} weights unchanged
        \Else
            \State Set $F_t\gets0$
            \State Set $p_t(a)\gets v_a/\sum_{b\in\cA}v_b$
            for every $a\in\cA$
            \State Draw $A_t\sim p_t$ and observe $Y_t$
            \For{$a\in\cA$}
                \State Set
                $\widehat\ell_t(a)
                \gets\mathbf 1\{A_t=a\}(1-Y_t)/p_t(a)$
                \State Update
                $v_a\gets v_a\exp\{-\eta\widehat\ell_t(a)\}$
            \EndFor
        \EndIf
    \EndFor
\EndFor
\end{algorithmic}
\end{breakablealgorithm}

\subsection{Local asymptotic inference}
\label{sec:local-inference}

Our finite-sample guarantees are designed to remain valid under the worst-case drift allowed by $\cM_{\beta,L}$. However in practice, it is indeed possible for the realized drift over a given inference block to be much smaller. Such is the case when each block is sufficiently locally stationary. In this regime, it is intuitive that reward-learning and inferential observations need not be treated separately, since both can provide useful information about the current effects. We therefore study an estimator that uses all observations from the preceding block, rather than only those reserved for inference. We show that, when the realized drift over that block is negligible relative to the sampling error, this estimator is asymptotically normal and yields standard pointwise confidence intervals for the current effects. Under a stronger local stationarity condition, it is also locally asymptotically efficient. Importantly, this
requires no change to the action policy: the gain comes entirely from reusing data that \textsc{TRACE} already collects for reward learning.

Throughout this subsection, $K$ is fixed and $T\to\infty$, while the mean trajectories and algorithmic parameters may depend on $T$. Let
$t_T=b_{j_T}$ be a deterministic inference-block boundary with $j_T\ge2$, and let $\mathcal W_T:=\mathcal B_{j_T-1}$ denote the preceding block. Every
observation in $\mathcal W_T$ is therefore available before inference is reported at $t_T$. Let $n_{T,a}>0$, $a\in\cA$, be deterministic reference sample sizes. These sequences will approximate the number of observations collected from each arm in $\mathcal W_T$. We first make explicit the joint probability laws under which the subsequent local analysis is carried out.

\paragraph{Probability laws.}
For each $T$, let $\mu^{(0)} = \bigl(\mu_{s,a}^{(0)}\bigr)_{s\le T,a\in\cA}$ denote the baseline mean trajectories, and let $\mathsf P_{T,0}$ denote the joint law of the actions and rewards induced by these trajectories and the policy $\pi$. Thus under $\mathsf P_{T,0}$, $A_s\mid\mathscr{F}_{s-1} \sim
\pi_s(\,\cdot\mid\mathscr{F}_{s-1})$ and $Y_{s} \mid \mathscr{F}_{s-1}, A_{s} = a \sim \operatorname{Bernoulli}\bigl(\mu_{s,a}^{(0)}\bigr)$.
If the mean trajectories are random and exogenous to the learner, $\mathsf P_{T,0}$ also includes the randomness of the environment process.

Unless stated otherwise, all probabilities, convergence in probability, and stochastic-order statements in the asymptotic-normality result below are taken under the baseline law $\mathsf P_{T,0}$. 

\paragraph{Using all observations in a block.}
The estimator $\widehat\tau_{j_T-1}(a)$ used by \textsc{TRACE} relies only on the observations explicitly reserved for inference. For local inference, we instead consider a post-processing estimator that uses every observation in $\mathcal W_T$, including those generated by \textsc{EXP3}. For $a\in\cA$, define $N_{T,a}:=\sum_{s\in\mathcal W_T}\mathbf 1\{A_s=a\}$ and
\begin{equation}
    \widetilde\mu_{T,a}
    :=
    \frac{
        \sum_{s\in\mathcal W_T}
        \mathbf 1\{A_s=a\}Y_s
    }{
        N_{T,a}
    },
    \;\;
    \widetilde\tau_T(a)
    :=
    \widetilde\mu_{T,a}-\widetilde\mu_{T,0},
    \; a\in\cA_+.
    \label{eq:all-observation-estimator}
\end{equation}

To characterize the asymptotic behavior of
$\widetilde\tau_T(a)$, we require two local conditions. First, the number of observations from each arm $N_{T, a}$ must stabilize around a some deterministic reference sample size $n_{T,a}$. Second, the drift within each block $\mathcal W_T$ must be negligible relative to the corresponding sampling error. These conditions are summarized below.

\begin{ass}[Local sampling and drift]
\label{ass:local-inference}
Let $n_{T,a}>0$, $a\in\cA$, be deterministic reference sample sizes. The following conditions hold under the baseline law $\mathsf P_{T,0}$:

\begin{enumerate}[label=(\roman*)]
    \item \emph{Count stability.} For some $p_a\in(0,1)$,
    \begin{equation}
        \min_{a\in\cA}n_{T,a}\to\infty,
        \;\;
        \max_{a\in\cA}
        \left|
            \frac{N_{T,a}}{n_{T,a}}-1
        \right|
        \xrightarrow{\mathsf P_{T,0}}0,
        \;\;
        \mu_{t_T,a}\xrightarrow{\mathsf P_{T,0}}p_a
        \;\; \text{for every }a\in\cA.
        \label{eq:local-count-stability}
    \end{equation}

    \item \emph{Locally negligible drift.} There exist deterministic
    $b_{T,a}\ge0$ and $\delta_T\to0$ such that
    \begin{equation}
        \mathsf P_{T,0}\left(
            \max_{s\in\mathcal W_T}
            |\mu_{s,a}-\mu_{t_T,a}|
            \le b_{T,a}
            \text{ for every }a\in\cA
        \right)
        \ge1-\delta_T,
        \;\;
        \max_{a\in\cA}
        \sqrt{n_{T,a}}\,b_{T,a}
        \to0.
        \label{eq:local-small-drift}
    \end{equation}
\end{enumerate}
\end{ass}

Assumption~\ref{ass:local-inference}(i) is standard in asymptotic inference for adaptive bandits ~\citep{KZ24,AWZ25,YZ26}. The rates $n_{T,a}$ may differ across arms and need not be proportional to $T$. Assumption~\ref{ass:local-inference}(ii) requires the bias from temporal variation within $\mathcal W_T$ to be negligible relative to the sampling error. This is analogous to the bias-negligibility conditions used for centered asymptotic inference in locally stationary models~\citep{Vogt12,Zhao15}. It should be noted that for deterministic trajectories, \eqref{eq:local-small-drift} is deterministic.

\begin{rem}
\label{rmk:small-realized-drift}
Assumption~\ref{ass:local-inference}(ii) holds naturally when the mean rewards evolve gradually due to exogenous changes. For example, suppose $\mu_{r,a}-\mu_{r-1,a}=d_{r,a}+\zeta_{r,a}$, where $d_{r,a}$ is predictable
and $\zeta_{r,a}$ is a martingale difference with sub-Weibull tails. Let $D_T:=\max_{a\in\cA}\max_{s\in\mathcal W_T}
|\sum_{r=s+1}^{t_T}d_{r,a}|$. Standard maximal martingale concentration (cf. Section 2.2 of  \cite{HRMS21}), using truncation when necessary, gives
$\max_{a\in\cA}\max_{s\in\mathcal W_T} |\mu_{s,a}-\mu_{t_T,a}| =\widetilde{\mathcal O}_{P}(D_T+v_T\sqrt{H_T})$ under law $\mathsf P_{T, 0}$,
where $H_T:=|\mathcal W_T|$ and $v_T$ controls the stochastic increments. Hence Assumption~\ref{ass:local-inference}(ii) holds whenever $\max_{a\in\cA}\sqrt{n_{T,a}}(D_T+v_T\sqrt{H_T})\to0$ up to logarithmic factors.
\end{rem}

For $a,b\in\cA_+$, define
\begin{equation}
\begin{aligned}
    \Sigma_T(a,b)
    &:=
    \frac{p_0(1-p_0)}{n_{T,0}}
    +
    \mathbf 1\{a=b\}
    \frac{p_a(1-p_a)}{n_{T,a}},\\
    \widehat\Sigma_T(a,b)
    &:=
    \frac{
        \widetilde\mu_{T,0}(1-\widetilde\mu_{T,0})
    }{
        N_{T,0}
    }
    +
    \mathbf 1\{a=b\}
    \frac{
        \widetilde\mu_{T,a}(1-\widetilde\mu_{T,a})
    }{
        N_{T,a}
    }.
    \label{eq:local-covariance}
\end{aligned}
\end{equation}
The common off-diagonal term accounts for the reference-arm estimate shared by all current-effect estimators. We define $\widehat\Sigma_T$ arbitrarily on the event that some $N_{T,a}=0$. Under this construction of $\widetilde\mu_{T,a}$ and $\widehat\Sigma_T(a,b)$, one can show the \textsc{TRACE} estimator produces asymptotically normal current-effect inference.

\begin{thm}[Local asymptotic normality]
\label{thm:local-normality}
Suppose Assumption~\ref{ass:local-inference}(i) and Assumption~\ref{ass:local-inference}(ii)  hold. Then, for every $a\in\cA_+$,
\begin{equation}
\begin{aligned}
    \widetilde\tau_T(a)-\tau_{t_T}(a)
    ={}&
    \sum_{s\in\mathcal W_T}
    \left(
        \frac{\mathbf 1\{A_s=a\}}{n_{T,a}}
        -
        \frac{\mathbf 1\{A_s=0\}}{n_{T,0}}
    \right)
    (Y_s-\mu_{s,A_s}) +
    o_{P}\left(
        \sqrt{
            \frac{1}{n_{T,a}}
            +
            \frac{1}{n_{T,0}}
        }
    \right),
    \label{eq:local-linearization}
\end{aligned}
\end{equation}
where $o_{P}$ is taken under baseline law $\mathsf P_{T,0}$. Let $\widetilde{\boldsymbol\tau}_T :=(\widetilde\tau_T(a))_{a\in\cA_+}$ and
$\boldsymbol\tau_{t_T} :=(\tau_{t_T}(a))_{a\in\cA_+}$. Then, for every fixed nonzero $c\in\mathbb R^{K-1}$,
\begin{equation}
    \frac{c^\top\widehat\Sigma_Tc}{c^\top\Sigma_Tc}
    \xrightarrow{\mathsf P_{T,0}}1,
    \;\;
    \frac{
        c^\top(
            \widetilde{\boldsymbol\tau}_T
            -
            \boldsymbol\tau_{t_T}
        )
    }{
        \sqrt{c^\top\widehat\Sigma_Tc}
    }
    \Rightarrow
    \mathcal{N}(0,1).
    \label{eq:local-studentized-clt}
\end{equation}
\end{thm}

\begin{proof}
We prove Theorem~\ref{thm:local-normality} in Appendix~\ref{appendix-sec:proof-of-local-normality}.
\end{proof}

Theorem~\ref{thm:local-normality} yields conventional pointwise inference for
the current effect at a deterministic block boundary. In particular, for every
fixed $\alpha\in(0,1)$ and $a\in\cA_+$,
\begin{equation}
    \mathsf P_{T, 0}\left(
        \tau_{t_T}(a)
        \in
        \left[
            \widetilde\tau_T(a)
            \pm
            z_{1-\alpha/2}
            \sqrt{\widehat\Sigma_T(a,a)}
        \right]
    \right)
    \longrightarrow
    1-\alpha,
    \label{eq:local-wald-ci}
\end{equation}
where $z_q$ denotes the $q$-quantile of the standard normal distribution. A similar argument can be applied to to the original \textsc{TRACE} reserved-sample estimator by restricting the sums in \eqref{eq:local-linearization} to $\mathcal S_{j_T-1,a}$ and taking $n_{T,a}=m$. Crucially, one does not need to modify the \textsc{TRACE} estimator to obtain asymptotically normal current-effect inference as all we require is that $\widetilde\tau_T(a)$ reuses the rewards already
collected during reward-learning rounds. As a result, one can reduce the asymptotic variance relative to inference based only on the reserved samples.

\paragraph{Local efficiency.}
Theorem~\ref{thm:local-normality} shows that the all-observation estimator is asymptotically normal when the preceding block is locally stable. We now show that, under a slightly stronger local condition, its asymptotic variance is also optimal. Thus, when the local approximation is accurate, reusing the rewards
collected by \textsc{EXP3} is not only valid but asymptotically efficient. To study local efficiency, we consider a sequence of local alternatives around the baseline
experiment $\mathsf P_{T,0}$.

Fix the history before $\mathcal W_T$ and consider the experiment generated by
the actions and rewards within that block, with the same adaptive action rule
under every parameter value. Let $p=(p_a)_{a\in\cA}\in(0,1)^K$, and suppose
the baseline mean trajectories admit the local representation
$\mu_{s,a}^{(0)}=p_a+g_{T,s,a}$, where $g_{T,t_T,a}=0$. For
$u\in\mathbb R^K$, let $\mathsf P_{T,u}$ denote the locally varying experiment
under which
\begin{equation}
    Y_s
    \mid
    \mathscr{F}_{s-1},A_s=a
    \sim
    \operatorname{Bernoulli}\left(
        p_a+\frac{u_a}{\sqrt{n_{T,a}}}+g_{T,s,a}
    \right),
    \;
    s\in\mathcal W_T.
    \label{eq:local-varying-experiment}
\end{equation}
The perturbation $g_{T,s,a}$ may be deterministic or exogenous to the learner, and all reward probabilities are assumed to remain in $[0,1]$. For comparison, let $\mathsf Q_{T,u}$ denote the locally constant experiment
with the same adaptive action rule and
\begin{equation}
    Y_s
    \mid
    \mathscr{F}_{s-1},A_s=a
    \sim
    \operatorname{Bernoulli}\left(
        p_a+\frac{u_a}{\sqrt{n_{T,a}}}
    \right),
    \;
    s\in\mathcal W_T.
    \label{eq:local-constant-experiment}
\end{equation}
Thus, $\mathsf P_{T,u}$ retains the within-block temporal variation $g_{T,s,a}$, whereas $\mathsf Q_{T,u}$ freezes the trajectory at its value at
$t_T$. For $a\in\cA_+$, the current effect at $t_T$ is
$\theta_{T,a}(u):= p_a-p_0+u_a/\sqrt{n_{T,a}}-u_0/\sqrt{n_{T,0}}$. An estimator is \emph{regular} if its centered and normalized limiting distribution is unchanged under every fixed local perturbation $u$. We now state the result for local asymptotic efficiency of \textsc{TRACE}.

\begin{thm}[Local asymptotic efficiency]
\label{thm:local-efficiency}
Suppose Assumption~\ref{ass:local-inference}(i) holds under $\mathsf{P}_{T,0}$ and there
exist deterministic $b_{T,a}\ge0$ and $\delta_T\to0$ such that
\begin{equation}
    \P\left(
        \max_{s\in\mathcal W_T}|g_{T,s,a}|
        \le b_{T,a}
        \text{ for every }a\in\cA
    \right)
    \ge1-\delta_T,
    \;\;
    \sum_{a\in\cA}n_{T,a}b_{T,a}^2\to0.
    \label{eq:local-equivalence-condition}
\end{equation}
Then the following hold.

\begin{enumerate}[label=(\roman*)]
    \item \emph{(Local equivalence).} For every fixed $u\in\mathbb R^K$,
    $\mathsf{TV}(\mathsf{P}_{T,u},\mathsf{Q}_{T,u})\to0$. Hence any statistic with a
    limiting distribution under $\mathsf{P}_{T,u}$ has the same limiting distribution
    under $\mathsf{Q}_{T,u}$.

    \item \emph{(Convolution lower bound).} Fix $a\in\cA_+$ such that
    $n_{T,a}/n_{T,0}$ converges in $[0,\infty]$. If
    $\widehat\theta_T(a)$ is regular and
    $(\widehat\theta_T(a)-\theta_{T,a}(u))/\sqrt{\Sigma_T(a,a)}
    \Rightarrow L$ under $\mathsf{Q}_{T,u}$ for every fixed $u$, where $L$ does not
    depend on $u$, then $L=N(0,1)\ast M$ for some probability law $M$.

    \item \emph{(Efficiency).} For every fixed $u$,
    $(\widetilde\tau_T(a)-\theta_{T,a}(u))/\sqrt{\Sigma_T(a,a)}
    \Rightarrow \mathcal{N}(0,1)$ under $\mathsf{Q}_{T,u}$. Hence
    $\widetilde\tau_T(a)$ attains the convolution lower bound and is locally
    asymptotically efficient.
\end{enumerate}
\end{thm}

\begin{proof}
We prove Theorem~\ref{thm:local-efficiency} in Appendix~\ref{appendix-sec:proof-of-local-efficiency}.
\end{proof}

Theorem~\ref{thm:local-efficiency} shows that sufficiently small drift implies the varying block is asymptotically equivalent to a constant-mean adaptive-bandit experiment. Consequently, no regular estimator can asymptotically improve on the variance of
$\widetilde\tau_T(a)$. Practically, this means that when a block is locally stable, the rewards collected for reward maximization can be reused without losing first-order efficiency, rather than discarding them and relying only on the reserved inference samples.

\subsection{Limits of adaptation to unknown smoothness}
\label{sec:unknown-smoothness}

Our results so far assume that $\beta$ and $L$ are known. It is natural to ask whether a procedure can instead adapt to the realized smoothness of the reward trajectories, producing sharper confidence intervals when the environment is substantially smoother than the worst case allowed by $\cM_{\beta,L}$. The difficulty is that honest inference must remain valid over the entire class, so even when the observed trajectory appears nearly stationary, the learner must still rule out nearby alternatives with localized variation that are consistent with the available data. Theorem~\ref{thm:smoothness-adaptation} shows that a regret constraint can limit the information needed to distinguish such alternatives, thereby preventing full adaptation even when the true environment is constant.

\begin{thm}[Regret-limited smoothness adaptation]
\label{thm:smoothness-adaptation}
Consider the two-arm problem, fix $\delta\in(0,1/8]$, and suppose $\pi$ has honest simultaneous coverage over $\cM_{\beta,L}$, where $\beta\in(0,1]$ and $L>0$. Let $\mu^\circ_0\equiv1/4$ and $\mu^\circ_1\equiv3/4$, and suppose $\mathscr R_T(\pi,\mu^\circ)\le r$. For any $0<\Delta\le1/16$ satisfying
\begin{equation}
    T^{-1}
    <
    \left(\frac{2\Delta}{L}\right)^{1/\beta}
    \le
    \frac{1}{32},
    \;\;
    r\Delta^{2+1/\beta}
    \le
    c_\beta L^{1/\beta},
    \label{eq:adaptation-range}
\end{equation}
for a sufficiently small constant $c_\beta>0$,
\begin{equation}
    \max_{\lceil T/2\rceil\le t\le T}
    \E_{\mu^\circ}^\pi |\mathscr{C}_{t}|
    \ge
    \frac{\Delta}{2}.
    \label{eq:adaptation-lower-bound}
\end{equation}
\end{thm}

\begin{proof}
We prove Theorem~\ref{thm:smoothness-adaptation} in Appendix~\ref{appendix-sec:proof-of-smoothness-adaptation}.
\end{proof}

Taking $\Delta \asymp c_{0} L^{1/(2\beta+1)}r^{-\beta/(2\beta+1)}$ for sufficiently small $c_{0}$ therefore gives, whenever \eqref{eq:adaptation-range} is satisfied, $\max_{\lceil T/2\rceil\le t\le T}\E_{\mu^\circ}^\pi|\mathscr{C}_{t}| \gtrsim_\beta L^{1/(2\beta+1)}r^{-\beta/(2\beta+1)}$.  This phenomenon is related to classical impossibility results for honest adaptation over smoothness   
classes~\citep{Low97,Bul12}. In particular, a data-dependent bandwidth is not sufficient if the learner has not observed the relevant arm often enough to verify that the wider bandwidth is valid. Thus, adaptation to realized inferential difficulty is possible without sacrificing validity, while adaptation to the regularity of the underlying environment can be fundamentally limited by the regret constraint.

\section{Numerical experiments}
\label{sec:numerical-experiments}
In this section, we present synthetic and real-data experiments to validate our theoretical findings. We first use synthetic nonstationary bandits to study the regret--precision tradeoff, prediction-adaptive exploration, and the role of local stationarity in current-effect inference. We then consider three real-world datasets. On Yahoo! R6B \citep{10.1007/s10791-019-09351-2}, we study the tradeoff between dynamic regret and current-effect precision. On the ZOZOTOWN Open Bandit Dataset \citep{saito2021open}, we evaluate the prediction-adaptive stopping rule in Adaptive \textsc{TRACE}. Finally, on HeartSteps \citep{klasnja2018efficacy, liao2020heartsteps}, we study time-varying treatment effects and the efficiency gain from local observation reuse. Additional numerical results are provided in Appendix~\ref{app:additional-experimental-results}.

\subsection{Synthetic data experiment}
\label{sec:exp-synthetic}

\paragraph{Setting and algorithms.}
We consider a $K=3$ arm nonstationary bandit over $T=300{,}000$ rounds with $\beta=1$, $L=0.05$, and $\delta=0.05$. For each arm $a\in\{0,1,2\}$, we set
\[
    \mu_{t,a}
    =
    \frac12+c_a
    +
    \frac{L}{2\pi}
    \sin\left(\frac{2\pi t}{T}+\frac{2\pi a}{3}\right),
    \;\;
    (c_0,c_1,c_2)=(0.02,0,-0.02),
\]
and generate bounded rewards with mean $\mu_{t,a}$. Arm~$0$ remains uniformly optimal, allowing us to isolate the inferential cost even when the reward-maximizing arm does not change. One can show that indeed $\mu_{t, a}$ is $L$-Lipschitz in $t$. We compare \textsc{TRACE} and Adaptive \textsc{TRACE} over target widths $w\in\{0.20,0.25,0.30,0.35,0.40\}$, together with restarted \textsc{EXP3} and \textsc{SW-UCB} \citep{garivier2008upper} as reward-only baselines. For Adaptive \textsc{TRACE}, we use oracle forecasts $\widehat Y_{t,a}=\Pi_{[0,1]}(\mu_{t,a}+\xi_{t,a})$, where $\xi_{t,a}\sim\mathcal N(0,\rho^2)$ are generated independently and revealed before the experiment to ensure predictability. We set $\rho=0.01$ in the main experiment and separately vary $\rho$ to illustrate adaptation to prediction quality.

\begin{figure}[H]
    \centering
    \includegraphics[width=0.32\textwidth]{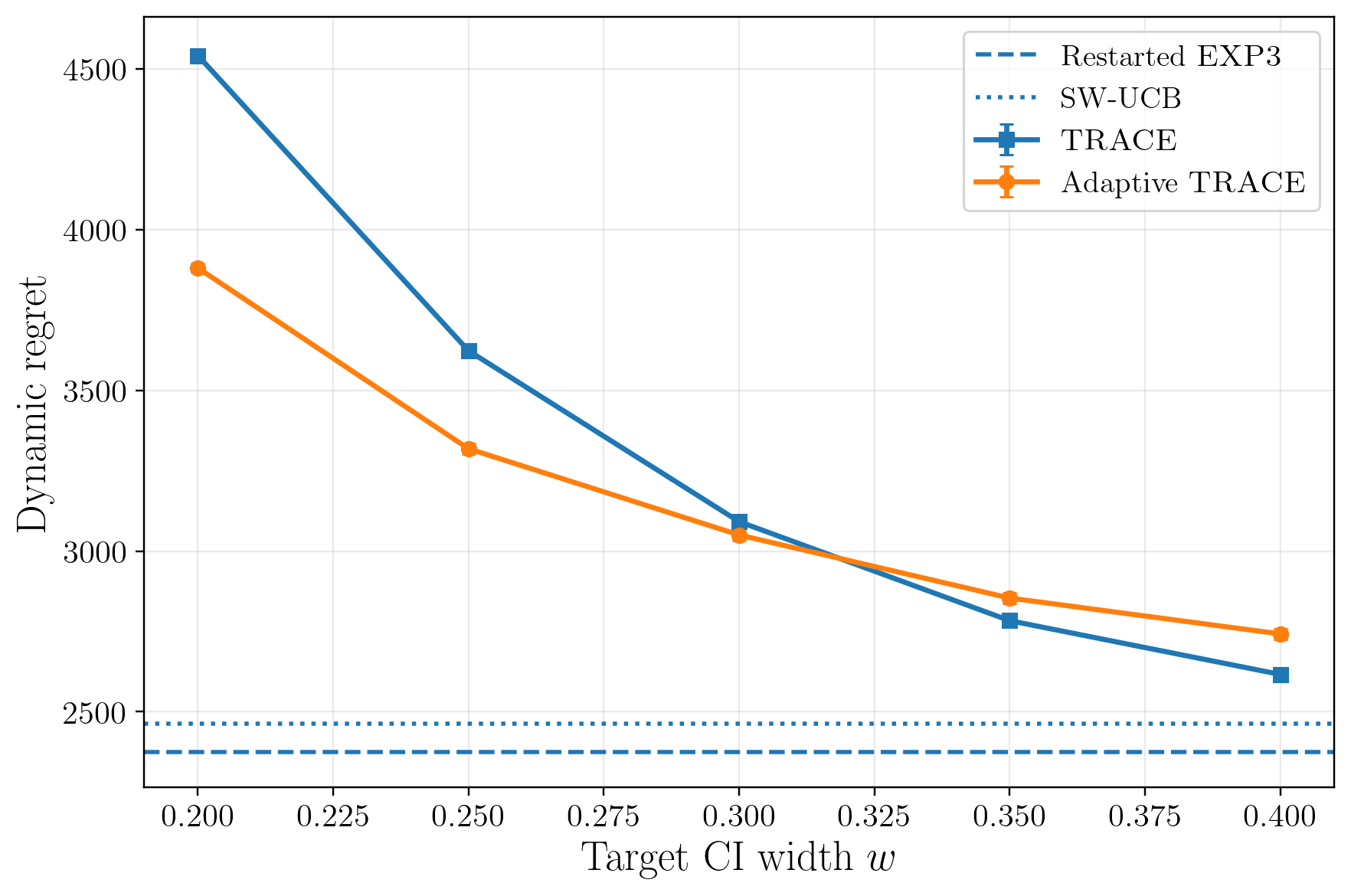}
    \hfill
    \includegraphics[width=0.32\textwidth]{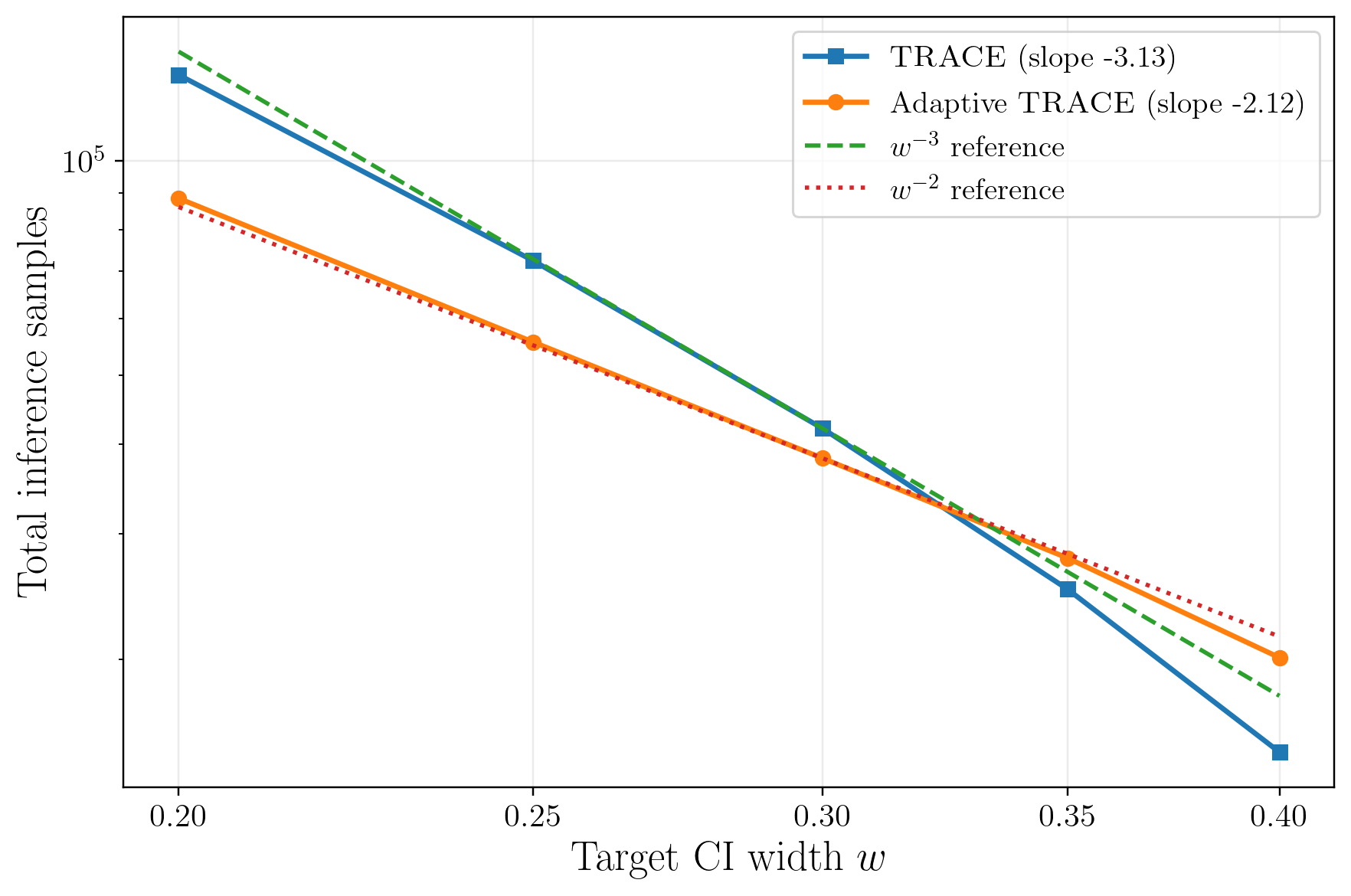}
    \hfill
    \includegraphics[width=0.32\textwidth]{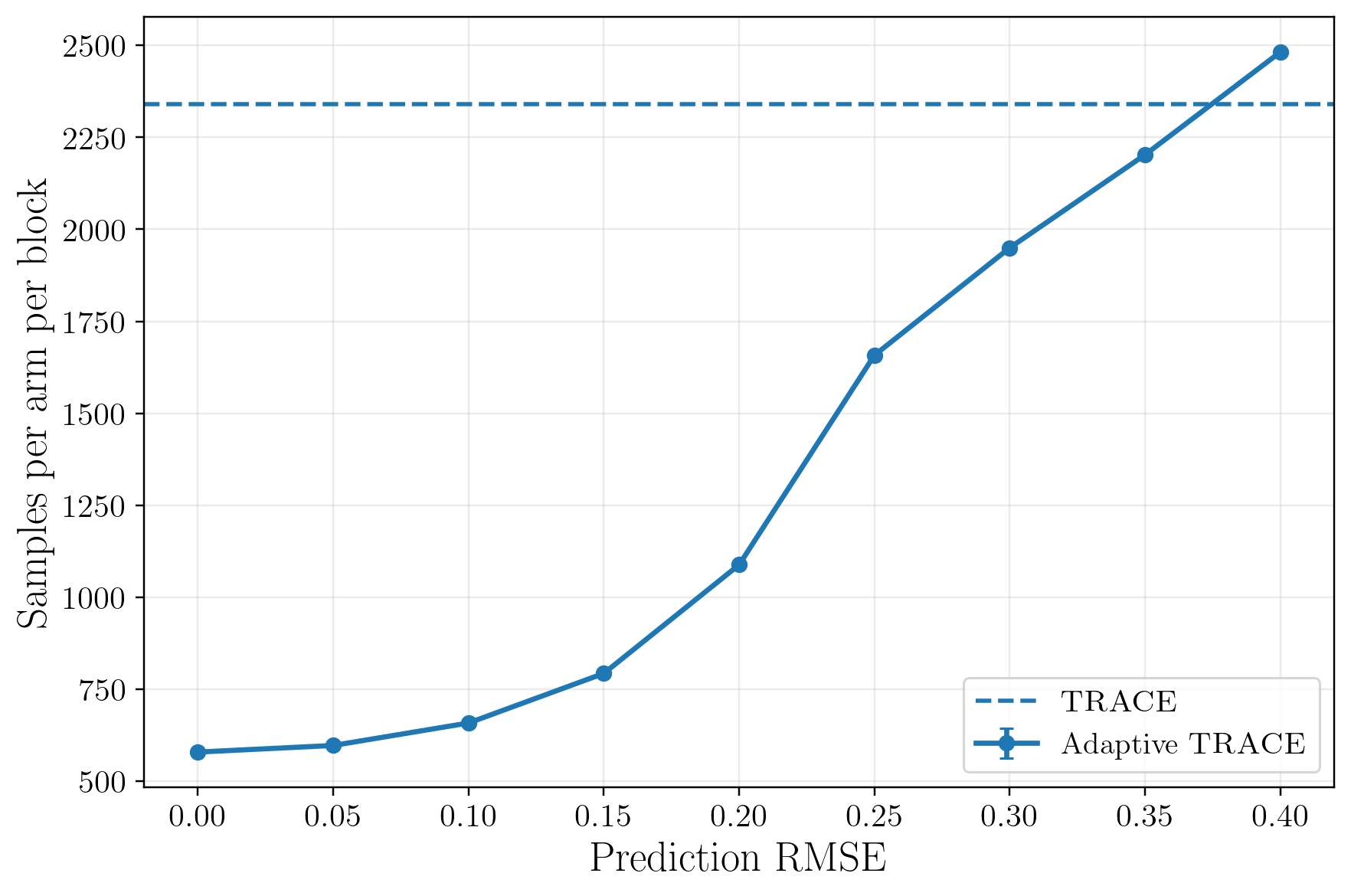}
    \caption{\textbf{Synthetic experiments.}
    \textit{Left:} Dynamic regret as the target width $w$ varies.
    \textit{Middle:} Total inference samples, with $w^{-3}$ and $w^{-2}$ reference rates.
    \textit{Right:} Samples required per arm and block as prediction RMSE increases.}
    \label{fig:synthetic-main}
\end{figure}

To evaluate our local-stationarity result, we additionally consider a two-arm block of length $H\in\{500,1000,2000,4000,8000\}$ with $\mu_{t,0}=0.5$ and $\mu_{t,1}=0.55-b_H(1-t/(H-1))$. We compare confidence intervals constructed using all observations collected by \textsc{TRACE} with those using only reserved inference observations. We consider $b_H=0$, locally negligible drift $b_H=2H^{-3/4}$, and non-negligible drift $b_H=0.2H^{-1/4}$.

\begin{figure}[H]
    \centering
    \includegraphics[width=0.49\textwidth]{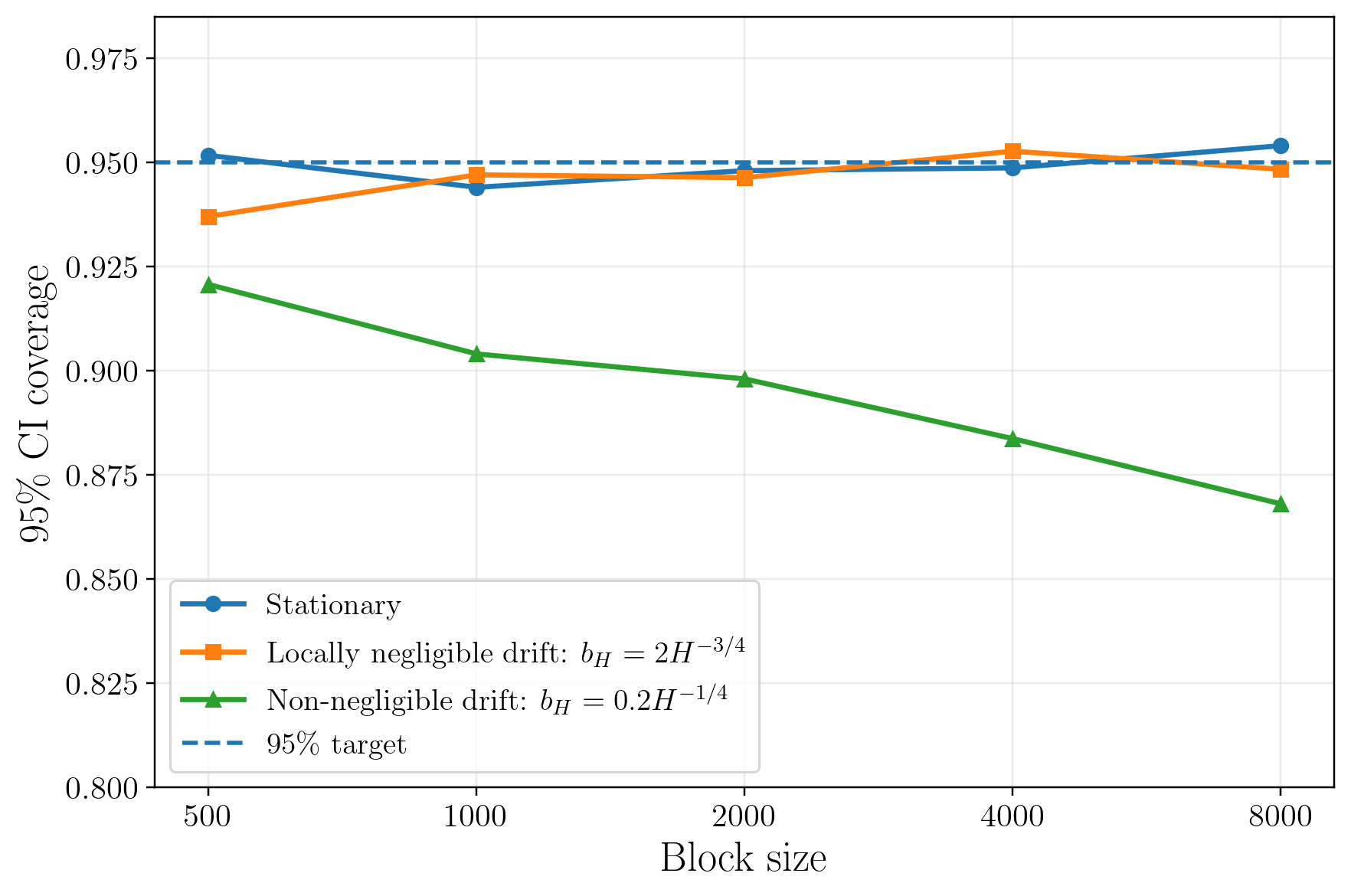}
    \hfill
    \includegraphics[width=0.49\textwidth]{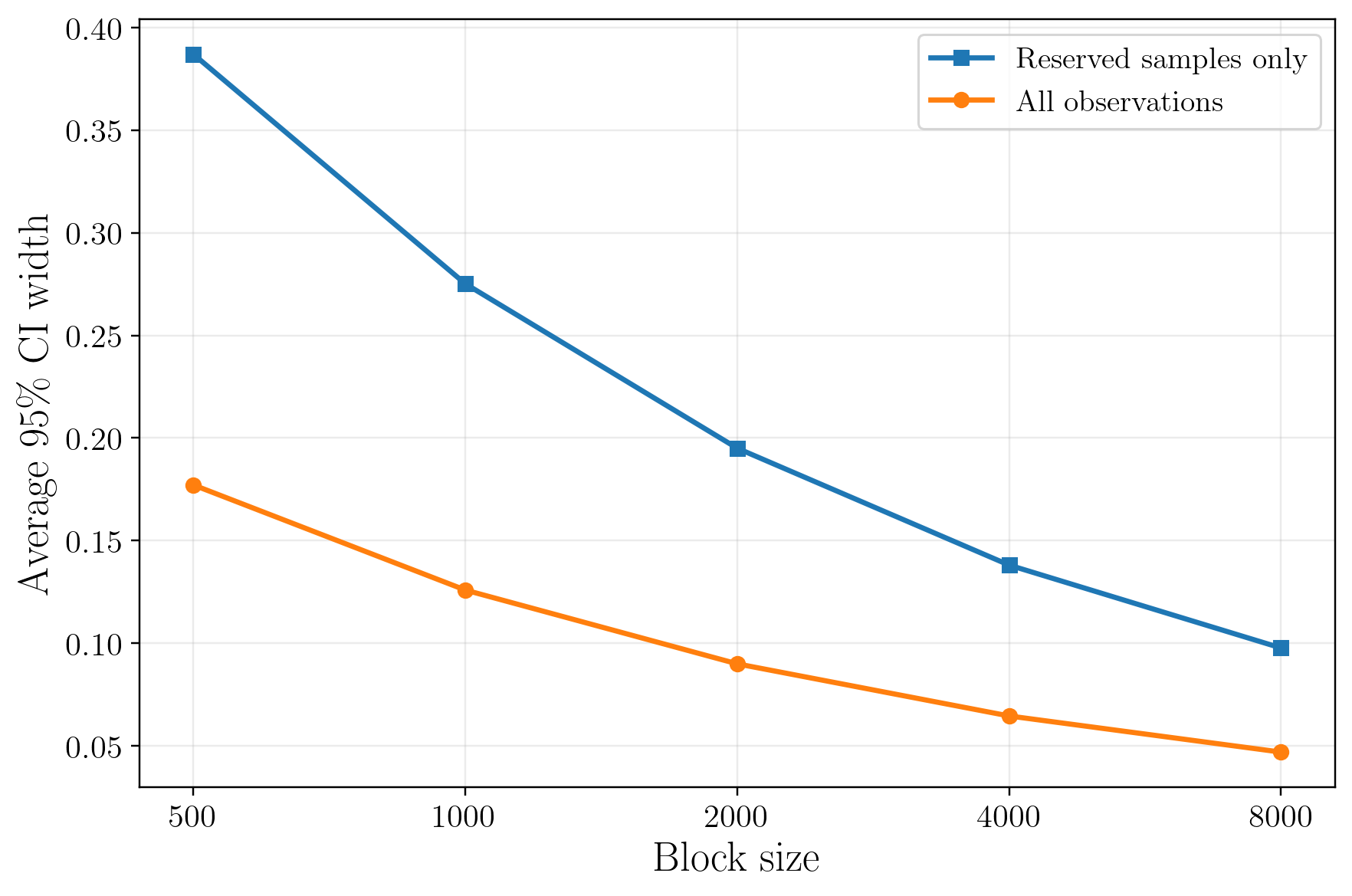}
    \caption{\textbf{Synthetic local-stationarity experiment.}
    \textit{Left:} Coverage of the nominal $95\%$ confidence interval under the three drift regimes.
    \textit{Right:} Average confidence-interval width using all observations versus reserved inference observations only.}
    \label{fig:synthetic-local}
\end{figure}

\paragraph{Results.}
Figure~\ref{fig:synthetic-main} confirms the main predictions of our finite-sample theory. Fixed \textsc{TRACE} exhibits approximately $w^{-3}$ inferential scaling, while Adaptive \textsc{TRACE} is close to $w^{-2}$ when prediction errors are small (see Corollary~\ref{cor:lipschitz}). The right panel further shows that Adaptive \textsc{TRACE} stops earlier when forecasts are accurate and requires more samples as prediction quality deteriorates. Figure~\ref{fig:synthetic-local} supports our local asymptotic result: reusing reward-learning observations maintains approximately nominal coverage under locally negligible drift and yields substantially narrower intervals, while coverage deteriorates when the local-stationarity condition is violated.

\subsection{Yahoo! R6B: the empirical price of current-effect inference}
\label{sec:exp-yahoo}

\begin{figure}[H]
    \centering
    \includegraphics[width=0.32\textwidth]{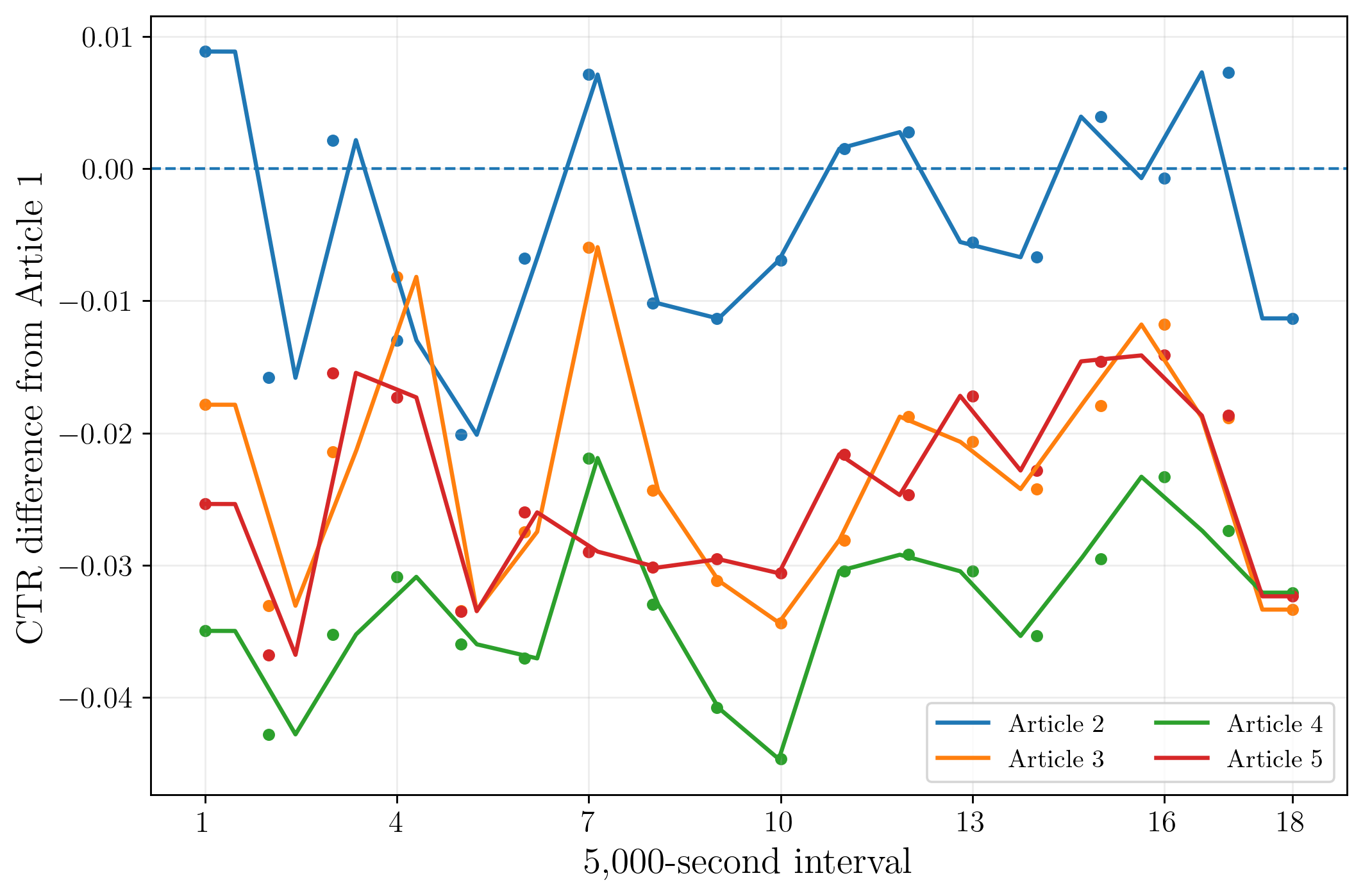}
    \hfill
    \includegraphics[width=0.32\textwidth]{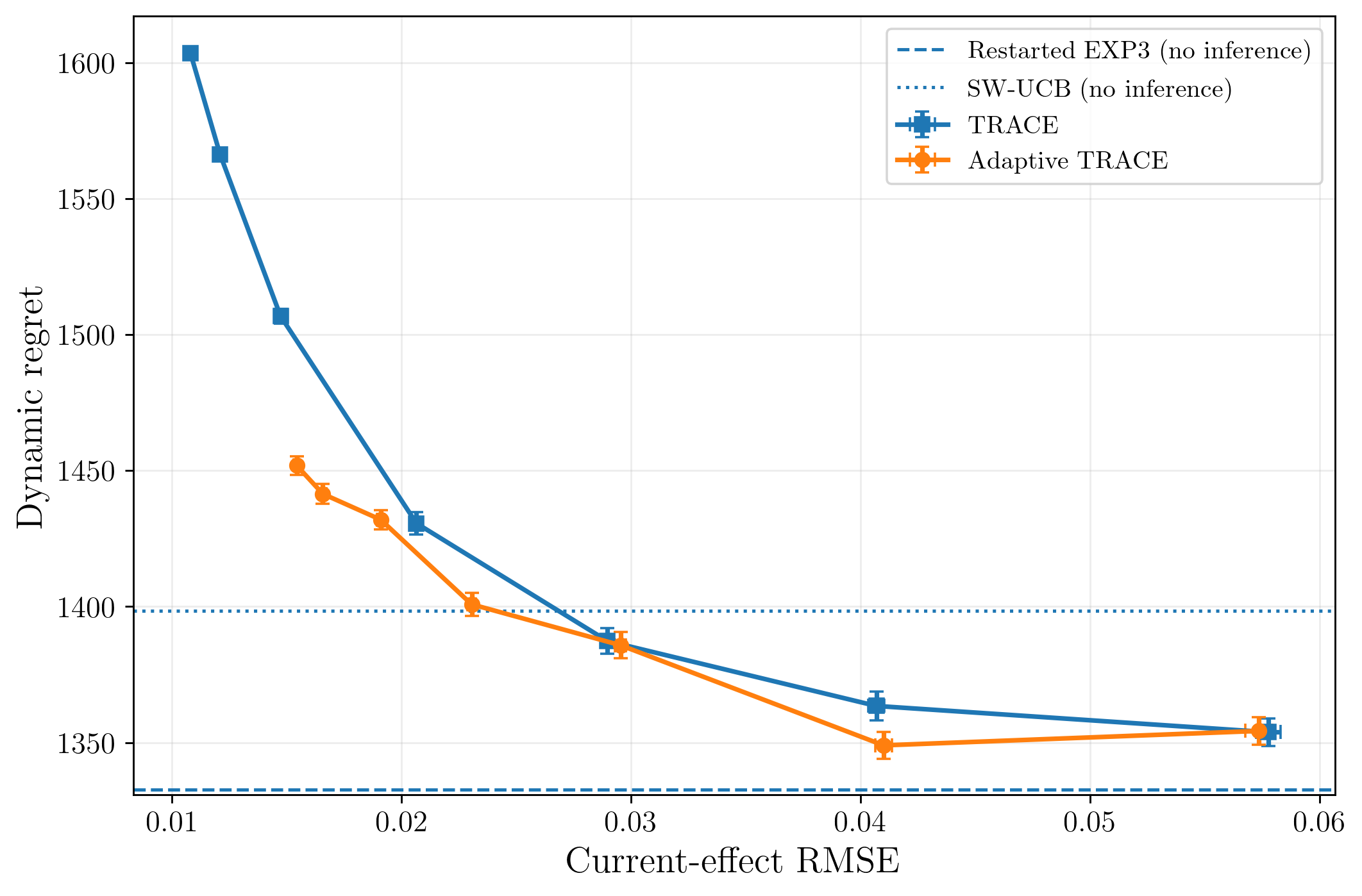}
    \hfill
    \includegraphics[width=0.32\textwidth]{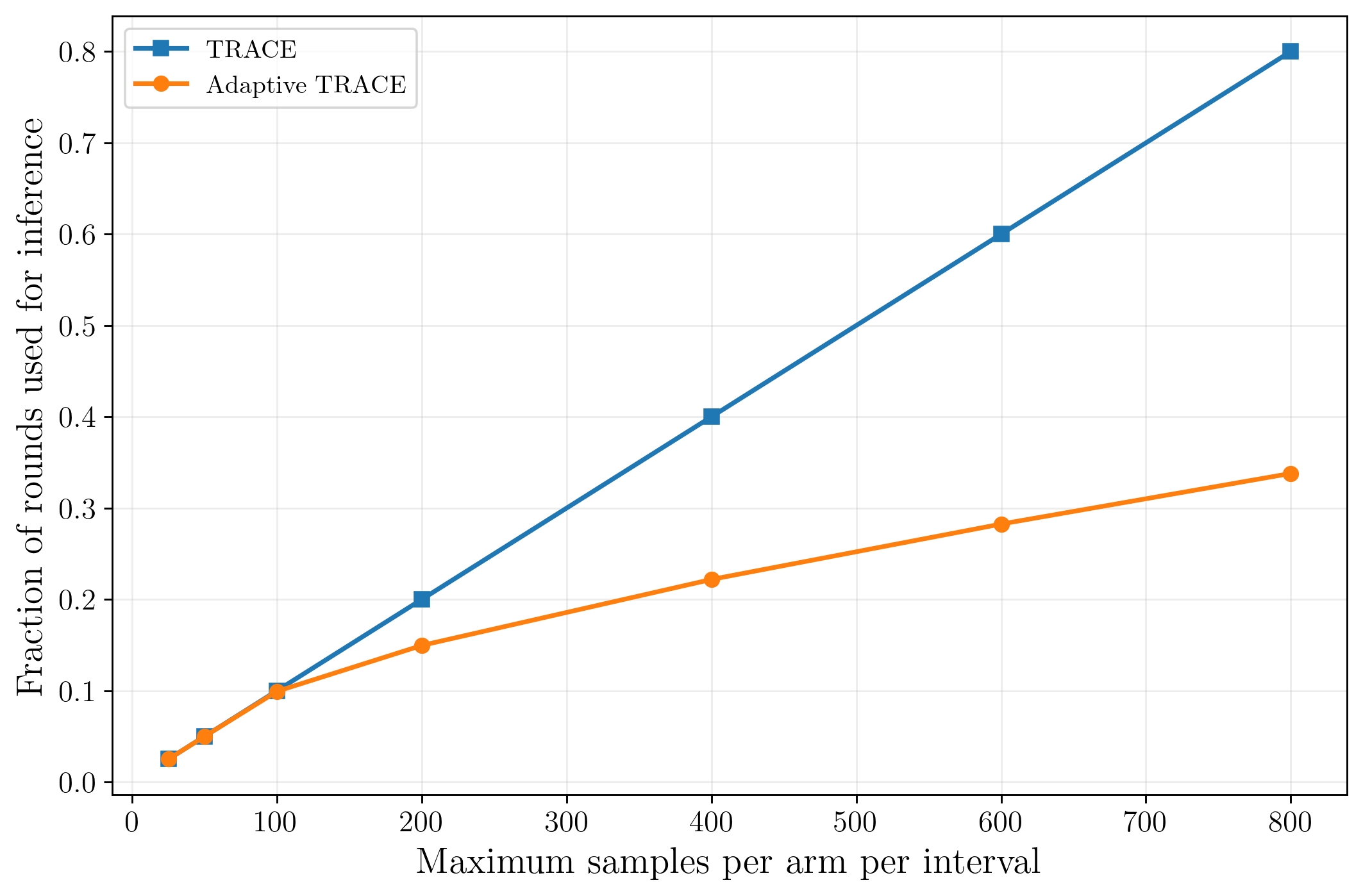}
    \caption{\textbf{Yahoo! R6B experiment.}
    \textit{Left:} Current click-through-rate effects relative to Article~1.
    \textit{Middle:} Dynamic regret versus current-effect RMSE for \textsc{TRACE} and Adaptive \textsc{TRACE}, with restarted \textsc{EXP3} and \textsc{SW-UCB} as reward-only baselines.
    \textit{Right:} Fraction of rounds used for inference as the maximum budget $m$ increases.}
    \label{fig:yahoo-main}
\end{figure}

\paragraph{Setting and algorithms.}
The Yahoo! R6B dataset \citep{10.1007/s10791-019-09351-2} contains randomized click logs from Yahoo!'s front page, where news articles were displayed to users and clicks were recorded as binary rewards. Using empirical click-through rates computed from the randomized Yahoo! R6B logs, we construct a nonstationary news recommendation problem with $K=5$ articles over $T=90{,}000$ rounds. We partition the horizon into consecutive $5{,}000$-round intervals and let $\overline{\mu}_{j,a}$ denote the empirical click-through rate of article $a$ within interval $j$, with $s_j$ denoting the corresponding interval endpoint. To obtain a reward trajectory satisfying our H\"older smoothness assumption, for $t\in[s_j,s_{j+1}]$, we linearly interpolate the empirical rates by setting $\mu_{t,a}=(1-\lambda_t)\overline{\mu}_{j,a}+\lambda_t\overline{\mu}_{j+1,a}$, where $\lambda_t=(t-s_j)/(s_{j+1}-s_j)$. We then generate Bernoulli rewards with mean $\mu_{t,a}$. The best article changes $11$ times across the $18$ intervals, yielding a strongly nonstationary environment. The left panel of Figure~\ref{fig:yahoo-main} shows the corresponding current effects relative to Article~1.

We compare \textsc{TRACE} (Algorithm~\ref{alg:trace}) and Adaptive \textsc{TRACE} (Algorithm~\ref{alg:adaptive-trace}) over varying inferential budgets $m\in\{25,50,100,200,400,600,800\}$. Fixed \textsc{TRACE} uses $m$ inference observations per arm in each interval, while Adaptive \textsc{TRACE} treats $m$ as the maximum inferential budget. We also report restarted \textsc{EXP3} and \textsc{SW-UCB} as reward-only baselines. For Adaptive \textsc{TRACE}, we use a Beta$(1,1)$-smoothed historical click-through rate as the predictable reward forecast. Specifically, before observing the reward at round $t$, we set
\[
    \widehat Y_{t,a}
    :=
    \frac{
        1+\sum_{s<t}\mathbf 1\{A_s=a\}Y_s
    }{
        2+\sum_{s<t}\mathbf 1\{A_s=a\}
    },
\]
which is the posterior mean under a Beta$(1,1)$ prior for the Bernoulli click probability of arm $a$. Since $\widehat Y_{t,a}$ depends only on observations collected before round $t$, it is $\mathscr F_{t-1}$-measurable and can be used directly in the empirical-Bernstein e-process. After observing $Y_t$, the corresponding success and observation counts are updated for future predictions.

\paragraph{Results.}
The middle panel of Figure~\ref{fig:yahoo-main} illustrates that increasing the inferential budget improves current-effect estimation but increases dynamic regret, which is consistent with our theoretical prediction. Adaptive \textsc{TRACE} improves this tradeoff by stopping inferential exploration early. The right panel shows that this gain becomes more pronounced as the inferential budget increases. Overall, the results support both the regret--inference tradeoff and the benefit of adaptive stopping predicted by our theory.

\subsection{ZOZOTOWN: prediction-adaptive inferential exploration}
\label{sec:exp-zozo}

\paragraph{Setting and algorithms.}
The ZOZOTOWN Open Bandit Dataset \citep{saito2021open} contains logged fashion recommendations from a large-scale online experiment, where each observation records the item shown to a user and whether the item was clicked. The experiment includes both uniformly randomized and bandit-generated recommendations across several product campaigns. We use the uniformly randomized Women's campaign, which contains recommendations among women's fashion items, and focus on five item IDs with sufficiently many randomized observations for our analysis.

We evaluate the empirical-Bernstein stopping rule using two predictable reward forecasts. The naive predictor sets $\widehat{Y}_{t,a}=0.5$. Our primary predictor uses a Beta$(1,1)$-smoothed historical click-through-rate predictor as in \cref{sec:exp-yahoo}. For each item, the predictor is initialized using randomized impressions observed before the evaluation period and is then updated sequentially as new rewards are observed. We compare the resulting prediction error and inferential sample requirement across the five selected items.

\begin{figure}[H]
    \centering
    \includegraphics[width=0.49\textwidth]{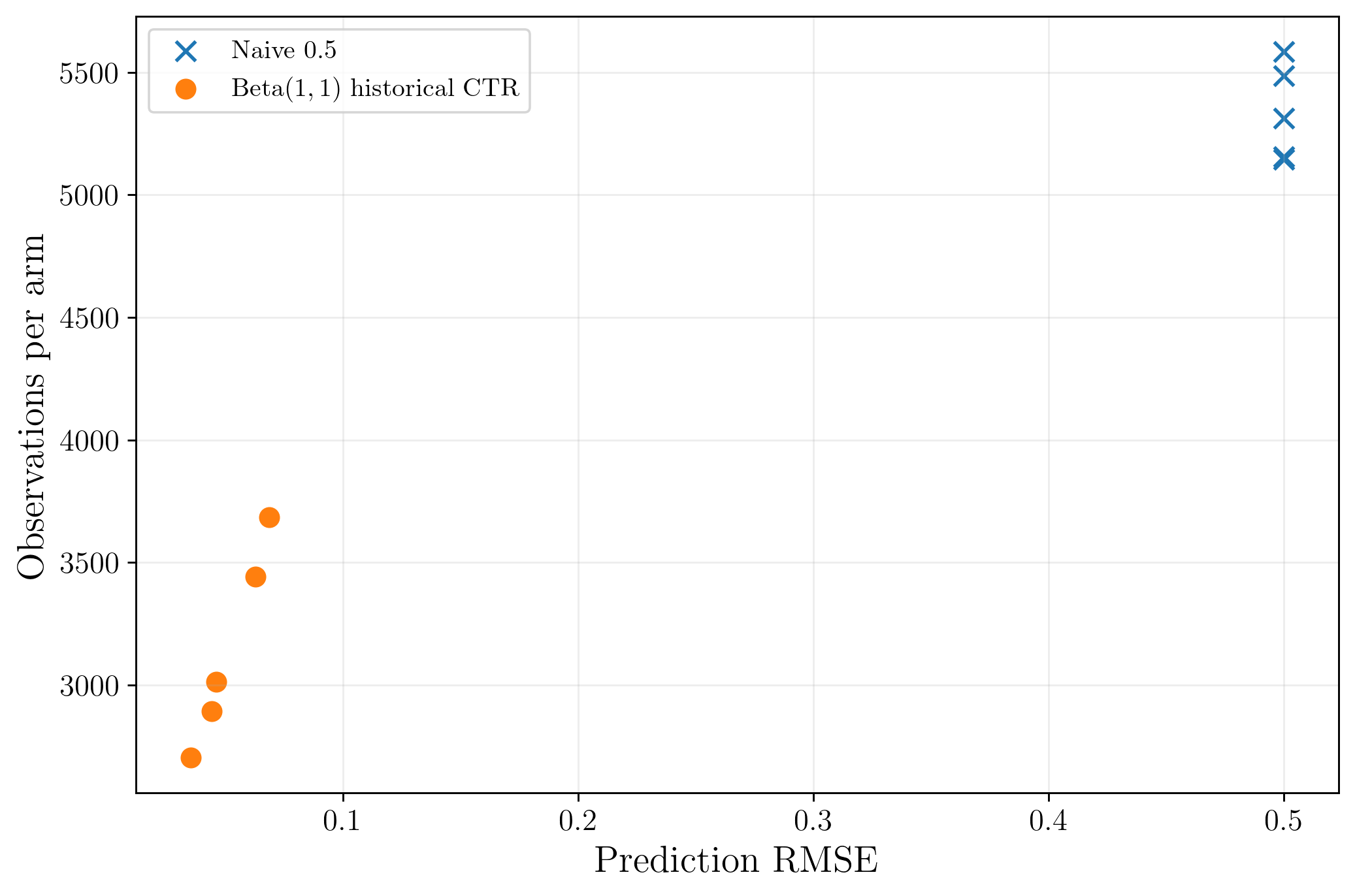}
    \hfill
    \includegraphics[width=0.49\textwidth]{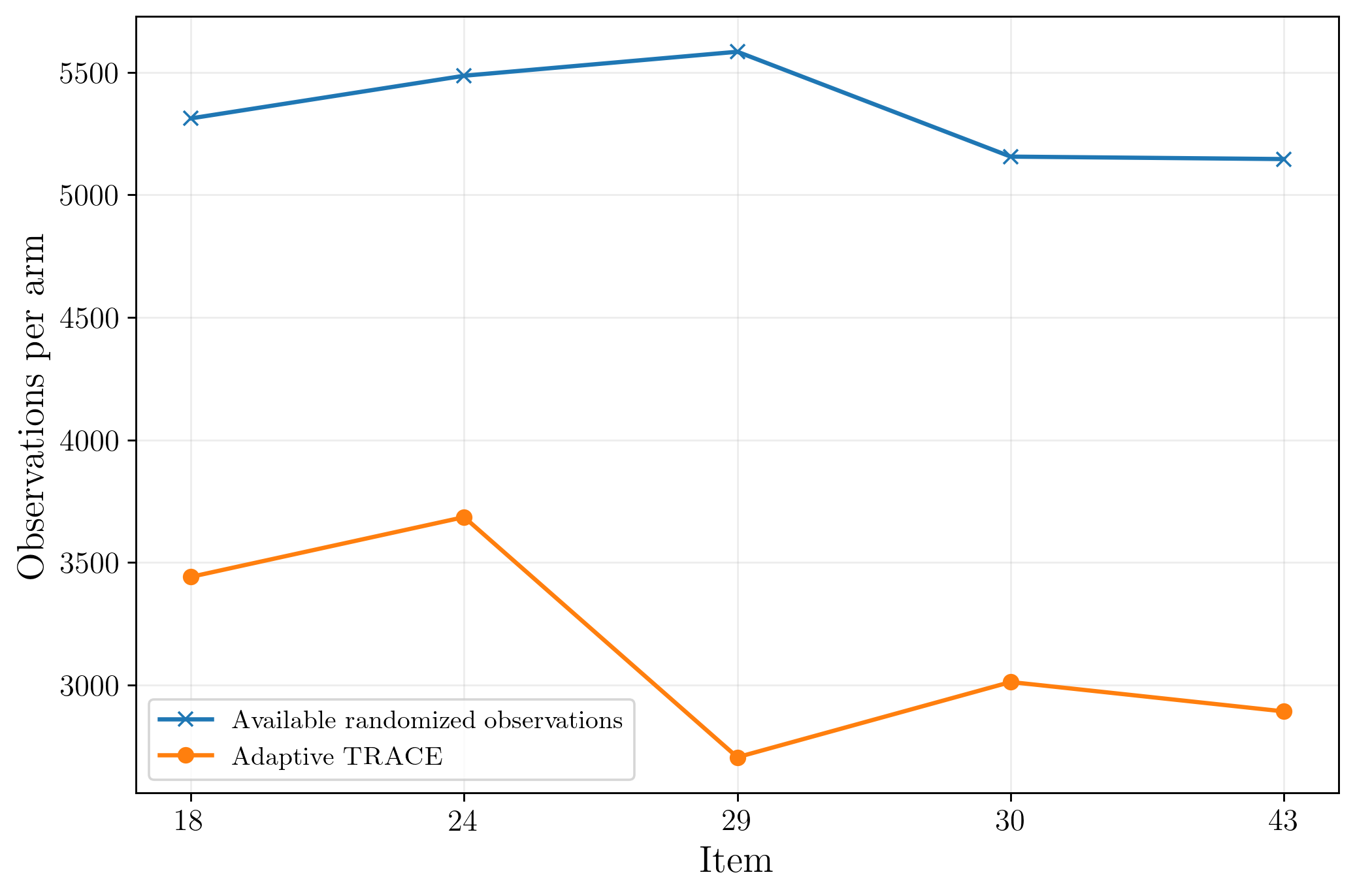}
    \caption{\textbf{ZOZOTOWN experiment.}
    \textit{Left:} Number of observations required for inference as a function of prediction RMSE.
    \textit{Right:} Available observations and observations used by Adaptive \textsc{TRACE} with the Beta$(1, 1)$ predictor across the five selected items.}
    \label{fig:zozo}
\end{figure}

\paragraph{Results.}
The left panel of Figure~\ref{fig:zozo} illustrates that more accurate predictable forecasts require fewer observations for inference, which is consistent with our theoretical prediction. The naive predictor exhausts the available randomized observations, while the calibrated predictors stop substantially earlier. As shown in the right panel, Adaptive \textsc{TRACE} reduces the number of inference observations on average across the five items. Overall, the results support the prediction-adaptive stopping behavior of Adaptive \textsc{TRACE}.

\subsection{HeartSteps: time-varying current treatment effects}
\label{sec:exp-heartsteps}

The HeartSteps study \citep{klasnja2018efficacy,liao2020heartsteps} is a $42$-day micro-randomized trial designed to evaluate mobile interventions for increasing physical activity. At five decision points per day, available participants were randomized to receive or not receive an activity suggestion. The outcome of interest is the log-step count during the $30$ minutes following each decision point. For participant $i$ and decision time $t$, we define $Y_{i,t}=\log(1+\texttt{jbsteps30}_{i,t})$, where $\texttt{jbsteps30}_{i,t}$ is the number of steps recorded during the subsequent $30$ minutes. We analyze $6{,}061$ randomized decision points over the six study weeks.

We estimate the proximal causal excursion effect using weighted and centered least squares (WCLS) \citep{boruvka2018assessing,shi2023assessing}. At available decision points, activity suggestions were randomized with probability $0.6$. Since the reference randomization probability equals the trial randomization probability, the WCLS probability weights equal one and treatment is centered as $A_{i,t}-0.6$. We therefore estimate
\[
    (\widehat{\alpha},\widehat{\beta})
    =
    \arg\min_{\alpha,\beta}
    \sum_{i,t}
    I_{i,t}
    \left[
        Y_{i,t}
        -
        g_t(H_{i,t})^\top\alpha
        -
        (A_{i,t}-0.6)f_t(S_{i,t})^\top\beta
    \right]^2,
\]
where $I_{i,t}$ denotes availability and $g_t(H_{i,t})$ includes pre-decision $30$-minute activity as a prognostic control. For the time-moderated analysis, we take $f_t(S_{i,t})=(1,d_{i,t})^\top$, where $d_{i,t}\in\{0,\ldots,41\}$ denotes study day, so that the proximal effect is $\tau(d)=\beta_0+\beta_1 d$. We use participant-clustered standard errors with the small-sample correction used in the original HeartSteps analysis and additionally report weekly randomized contrasts without imposing a linear time trend.

Using randomized HeartSteps observations, we construct a nonstationary treatment-effect problem by estimating smooth daily treatment and control mean trajectories for the Bernoulli proximal outcome
$Y_{i,t}^{\mathrm{bin}}:=\mathbf 1\{\texttt{jbsteps30}_{i,t}>0\}$.
The resulting HeartSteps-derived environment contains $6{,}179$ randomized decision points from $37$ participants and has known current effect
$\tau_t:=\mu_{t,1}-\mu_{t,0}$ at each study day. We estimate the mean trajectories using Gaussian kernel smoothing, with bandwidth selected by leave-one-day-out binomial cross-validation. For target days $t\in\{15,21,28,35,42\}$, we use preceding windows of length
$h\in\{1,2,3,5,7,10,14\}$ days and construct the all-observation estimator and plug-in Wald interval from \eqref{eq:all-observation-estimator} and \eqref{eq:local-wald-ci}. We measure local nonstationarity by
$\Gamma_h:=\max_{a\in\{0,1\}}\sqrt{N_{t,a}}\max_{s\in\mathcal W_{t,h}}|\mu_{s,a}-\mu_{t,a}|$,
the finite-sample analogue of Assumption~\ref{ass:local-inference}. For each target day and window length, we evaluate coverage and efficiency over $3{,}500$ Monte Carlo replications and compare the all-observation estimator with inference based on an outcome-independent $25\%$ reserved subset.

\begin{figure}[H]
    \centering
    \includegraphics[width=0.32\textwidth]{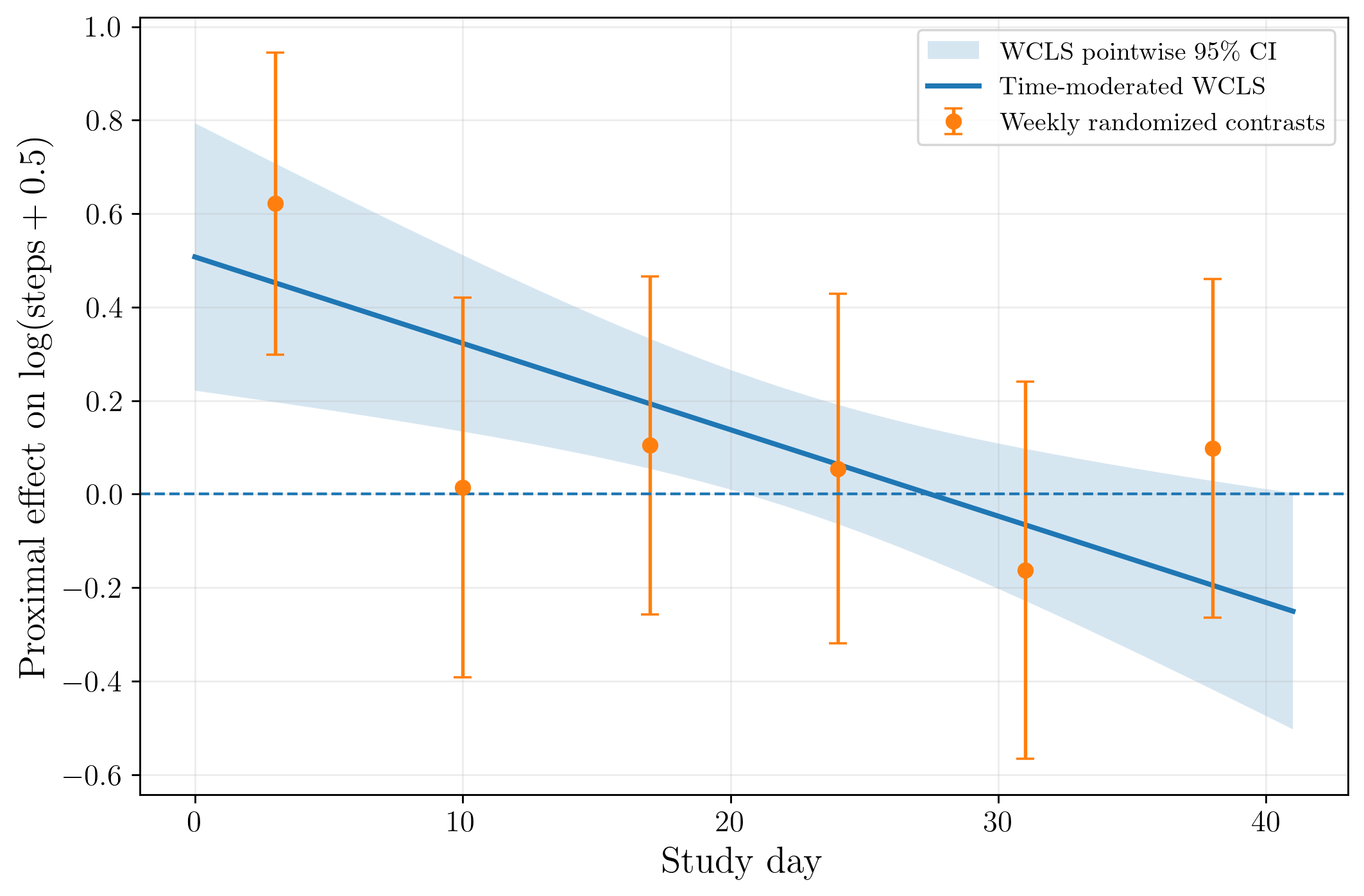}
    \hfill
    \includegraphics[width=0.32\textwidth]{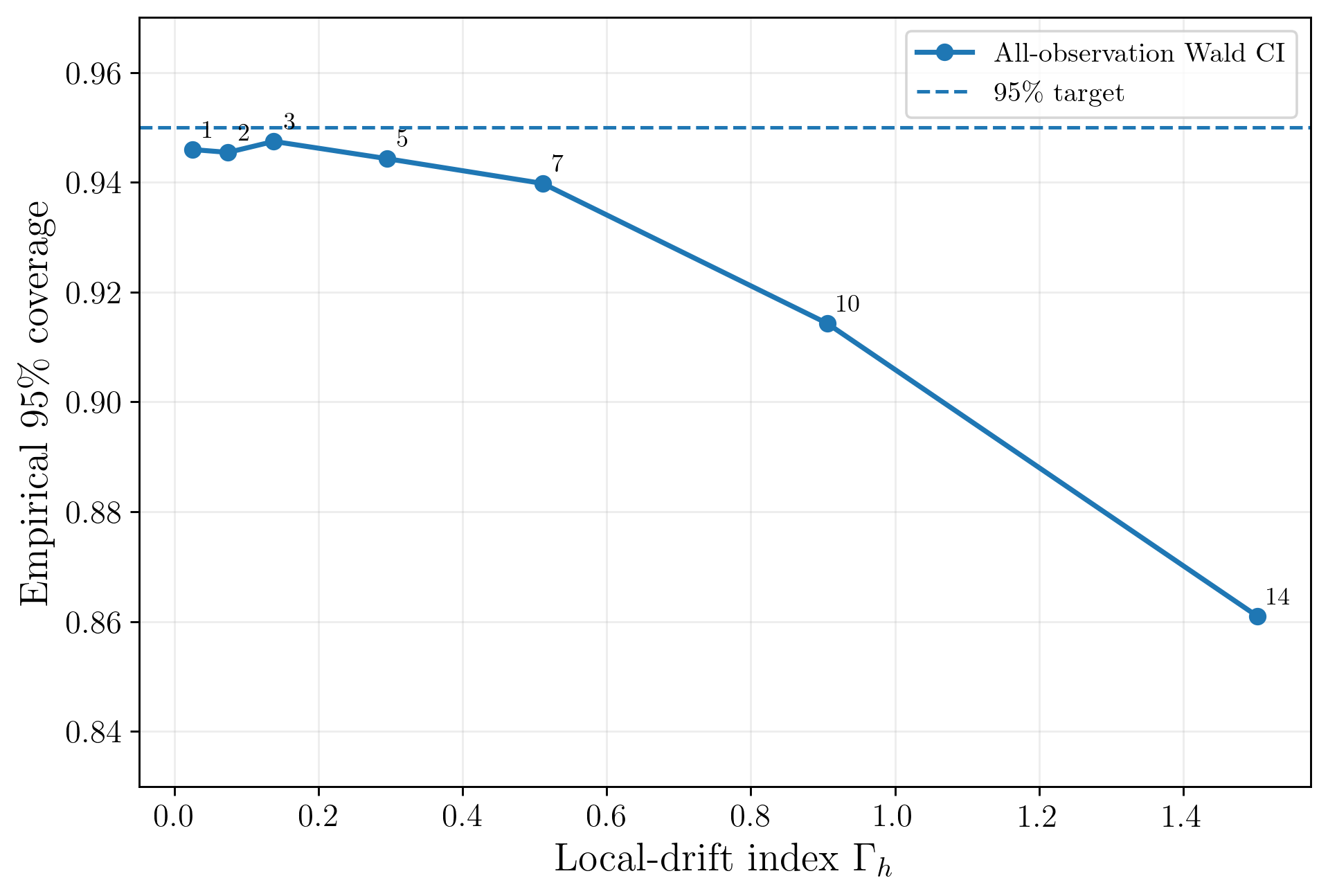}
    \hfill
    \includegraphics[width=0.32\textwidth]{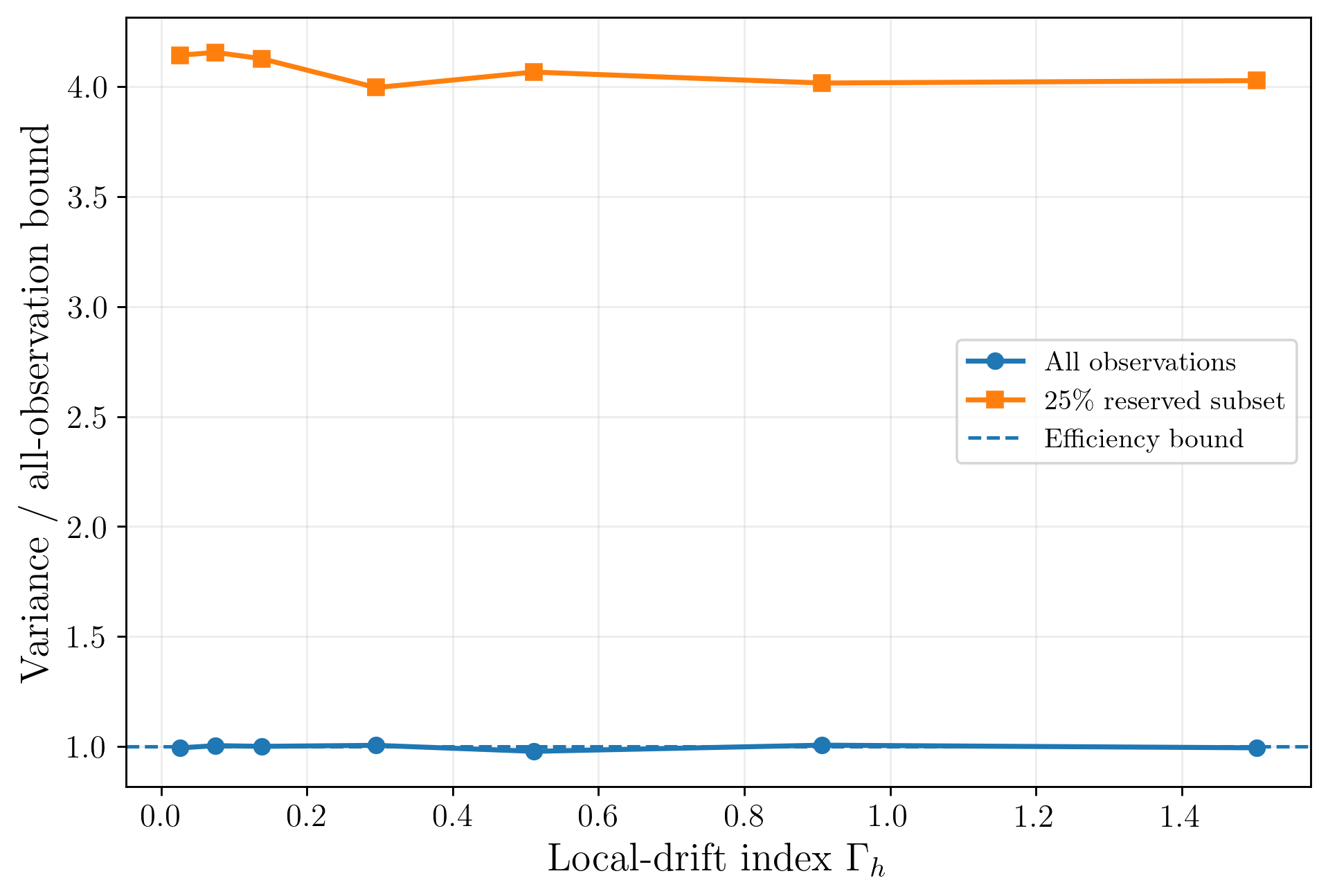}
    \caption{\textbf{HeartSteps experiment.}
    \textit{Left:} Time-varying proximal treatment effect estimated using WCLS, together with weekly randomized contrasts.
    \textit{Middle:} Empirical coverage of the all-observation Wald interval as a function of the local-drift index $\Gamma_h$. Numbers indicate the local window length in days.
    \textit{Right:} Empirical variance relative to the all-observation efficiency bound, comparing the all-observation estimator with inference based on a $25\%$ reserved subset.}
    \label{fig:heartsteps}
\end{figure}

\begin{table}[H]
    \centering
    \caption{\textbf{WCLS estimates for HeartSteps.}}
    \label{tab:heartsteps-wcls}
    \begin{tabular}{lccc}
        \toprule
        Estimand & Estimate & $95\%$ CI & $p$-value \\
        \midrule
        Marginal proximal effect
        & $0.1312$
        & $[-0.0058,\,0.2682]$
        & $0.0599$ \\
        Effect at study day $0$
        & $0.5074$
        & $[0.2009,\,0.8140]$
        & $0.0020$ \\
        Change in effect per day
        & $-0.01848$
        & $[-0.03090,\,-0.00607]$
        & $0.00479$ \\
        \bottomrule
    \end{tabular}
\end{table}

\begin{table}[H]
    \centering
    \small
    \setlength{\tabcolsep}{6pt}
    \renewcommand{\arraystretch}{1.1}
    \caption{\textbf{HeartSteps local-inference results.}
    Empirical $95\%$ coverage and variance relative to the all-observation efficiency bound, averaged over the five target days.}
    \label{tab:heartsteps-local}
    \begin{tabular}{ccccc}
        \toprule
        Window
        & $\Gamma_h$
        & Coverage
        & Variance ratio
        & Reserved variance ratio \\
        \midrule
        $1$ day   & $0.026$ & $94.6\%$ & $0.994$ & $4.14$ \\
        $2$ days  & $0.075$ & $94.5\%$ & $1.004$ & $4.16$ \\
        $3$ days  & $0.138$ & $94.7\%$ & $1.001$ & $4.13$ \\
        $5$ days  & $0.295$ & $94.4\%$ & $1.006$ & $4.00$ \\
        $7$ days  & $0.511$ & $94.0\%$ & $0.978$ & $4.07$ \\
        $10$ days & $0.906$ & $91.4\%$ & $1.006$ & $4.02$ \\
        $14$ days & $1.503$ & $86.1\%$ & $0.994$ & $4.03$ \\
        \bottomrule
    \end{tabular}
\end{table}

\paragraph{Results.}
The left panel of Figure~\ref{fig:heartsteps} shows that the proximal treatment effect declines over study time, with the weekly randomized contrasts exhibiting the same qualitative pattern. This highlights the distinction between inference on a current treatment effect and inference on a single study-wide average effect.

The local-inference results support the asymptotic theory. When the preceding window is locally stable, the all-observation Wald interval has approximately nominal coverage, the studentized estimator is close to standard normal, and its empirical variance is near the local asymptotic variance lower bound. As the window expands and temporal drift becomes non-negligible relative to sampling error, coverage deteriorates due to increasing bias. Using all observations also yields substantially lower variance than inference based only on the reserved subset.

\section{Conclusion}
\label{sec:conclusion}

We studied the tradeoff between reward maximization and inference for current treatment effects in nonstationary bandits. Our main result characterizes, up to logarithmic factors, the minimum dynamic regret required to maintain simultaneously valid and uniformly precise confidence intervals under H\"older drift. The resulting regret--precision frontier separates the usual cost of reward learning from the additional cost of maintaining sufficiently recent information about each arm. This inferential cost remains unavoidable even when the same arm is uniformly optimal.

We introduced \textsc{TRACE}, which attains this tradeoff by separating reward-learning and inferential sampling, together with an empirical-Bernstein extension that adapts exploration to realized prediction errors while preserving the same worst-case guarantee. We further showed that locally stable periods permit reward-learning observations to be reused for sharper pointwise inference, yielding asymptotically normal and, under stronger conditions, locally asymptotically efficient estimators. Our numerical experiments support these theoretical conclusions. There are many natural and interesting directions for future work such as extending these ideas to contextual bandits, structured interventions, and richer nonstationary decision problems.

\newpage

\bibliographystyle{plainnat}
\bibliography{refs}

\newpage 

\appendix

\section{Proof of Theorem~\ref{thm:tradeoff}}
\label{appendix:proof-tradeoff}

We prove the upper and lower bounds separately.  
Theorem~\ref{thm:upper-bound} proves the upper bound (see Appendix~\ref{appendix-sec:proof-of-nonadaptive-upper-bound} for the proof). We now prove the lower bound. We first prove the following result which we will subsequently use in our analysis. We take $w_0\le1/16$ and $\delta_0\le1/8$ throughout.

\begin{lem}[Simultaneous sign recovery]
\label{lem:simultaneous-sign-recovery}
Let $\mathcal{J}$ be a finite index set with $|\mathcal{J}|=d$, and consider Bernoulli environments indexed by $\theta\in\{-1,1\}^{\mathcal{J}}$. For each $j\in\mathcal{J}$, let $a_j\in\cA$ and let $\mathcal{S}_j\subseteq[T]$ be
deterministic such that the arm--round sets $\{(t,a_j):t\in\mathcal{S}_j\}$ are pairwise disjoint. Coordinate $\theta_j$ affects only the reward distribution of arm $a_j$ on $\mathcal{S}_j$. Suppose
the two possible Bernoulli means lie in $[3/16,5/16]$ and differ by at most $2\Delta$. Define
$N_j:=\sum_{t\in\mathcal{S}_j}\mathbf 1\{A_t=a_j\}$. If an estimator satisfies $\P_\theta^\pi(\widehat\theta=\theta)\ge1-\delta$ for every $\theta$, then
\begin{equation}
    2^{-d}\sum_{\theta\in\{-1,1\}^{\mathcal J}}
    \E_\theta^\pi\left[\sum_{j\in\mathcal J}N_j\right]
    \gtrsim
    \frac{d}{\Delta^2}\log\left(\frac{d}{\delta}\right).
    \label{eq:simultaneous-sign-recovery}
\end{equation}
\end{lem}

\begin{proof}
Put the uniform prior on $\Theta\in\{-1,1\}^{\mathcal J}$. For any fixed realization of $(A_t,Y_t,U_t)_{t\le T}$, the policy kernels do not depend on $\theta$, while every reward likelihood depends on at most one coordinate. Hence the likelihood, and therefore the posterior, factors over
$j\in\mathcal J$.

Let $\Lambda_j$ be the log-likelihood ratio between $\theta_j=1$ and $\theta_j=-1$, and let $e_j$ denote the smaller posterior probability for coordinate $j$. Then $e_j=(1+\exp|\Lambda_j|)^{-1}$. Since the posterior factors across coordinates, the largest posterior probability assigned to any $\theta\in\{-1,1\}^{\mathcal J}$ is $\prod_{j\in\mathcal J}(1-e_j)$. Since the estimator $\widehat{\theta}$ satisfies $\P_{\theta}^{\pi}(\widehat{\theta}=\theta)\geq 1-\delta$ for every $\theta$, averaging over the uniform prior on $\Theta$ gives
\[
    1-\delta
    \leq
    \P(\widehat{\theta}=\Theta)
    =
    \E\!\left[\P(\Theta=\widehat{\theta}\mid\mathscr{F}_{T})\right]
    \leq
    \E\!\left[\prod_{j\in\mathcal{J}}(1-e_j)\right].
\]
Thus, $\E[1-\prod_{j\in\mathcal J}(1-e_j)]\leq\delta$. By Markov's inequality, $\P(\prod_{j\in\mathcal J}(1-e_j)\geq1/2)\geq1-2\delta\geq3/4$. On this event, $1-x\leq\exp(-x)$ implies $\sum_{j\in\mathcal J}e_j\leq\log 2$. Jensen's inequality then gives $\sum_{j\in\mathcal J}\log(1/e_j)\geq d\log(d/\log 2)$. Therefore,
\begin{equation}
    \E\left[\sum_{j\in\mathcal J}\log(1/e_j)\right]
    \geq \frac{3d}{4}\log d.
    \label{eq:posterior-log-error}
\end{equation}

For $j\in\mathcal J$, let $\theta^{(j)}$ denote $\theta$ with coordinate $j$ flipped, and define
\[
    \mathcal I
    :=
    2^{-d}
    \sum_{\theta}
    \sum_{j\in\mathcal J}
    D_{\mathrm{KL}}
    \left(
        \mathsf P_\theta^\pi
        \,\|\, 
        \mathsf P_{\theta^{(j)}}^\pi
    \right).
\]
To see why averaging over the uniform prior gives $\mathcal I=\E[\sum_j\Theta_j\Lambda_j]$, fix $\theta$ and $j$. Since $\theta$ and $\theta^{(j)}$ differ only in coordinate $j$, all likelihood factors cancel except the $j$th coordinate. Thus,
\[
    \log
    \frac{d\mathsf P_\theta^\pi}
         {d\mathsf P_{\theta^{(j)}}^\pi}
    =
    \begin{cases}
        \Lambda_j, & \theta_j=1,\\
        -\Lambda_j, & \theta_j=-1,
    \end{cases}
    =
    \theta_j\Lambda_j.
\]
Therefore, $D_{\mathrm{KL}}
    \left(
        \mathsf P_\theta^\pi
        \,\|\,
        \mathsf P_{\theta^{(j)}}^\pi
    \right)
    =
    \E_\theta^\pi[\theta_j\Lambda_j]$, 
and hence
\[
    \mathcal I
    =
    2^{-d}\sum_{\theta}\sum_{j\in\mathcal J}
    \E_\theta^\pi[\theta_j\Lambda_j]
    =
    \E\left[\sum_{j\in\mathcal J}\Theta_j\Lambda_j\right].
\]

Conditional on the observed data, $\E[\Theta_j\Lambda_j\mid\mathscr{F}_T]=(1-2e_j)|\Lambda_j|$. Indeed, if $\Lambda_j\geq0$, then the posterior probabilities of $\Theta_j=1$ and $\Theta_j=-1$ are $1-e_j$ and $e_j$, respectively, so $\E[\Theta_j\mid\mathscr{F}_T]=1-2e_j$. If $\Lambda_j<0$, then $\E[\Theta_j\mid\mathscr{F}_T]=-(1-2e_j)$ while $\Lambda_j=-|\Lambda_j|$, giving the same product. Moreover, since $e_j=(1+\exp|\Lambda_j|)^{-1}$, we have $|\Lambda_j|=\log((1-e_j)/e_j)$. Thus,
\[
    \mathcal I
    =
    \E\left[
        \sum_{j\in\mathcal J}
        (1-2e_j)
        \log\left(\frac{1-e_j}{e_j}\right)
    \right].
\]
For $e\in(0,1/2]$,
\[
    \log(1/e)
    -
    (1-2e)\log((1-e)/e)
    \leq
    \log2+2e\log(1/e)
    <2.
\]
Using \cref{eq:posterior-log-error},
$\mathcal I\ge(3d/4)\log d-2d$.

We also need the dependence on $\delta$. For fixed $\theta$ and $j$, let $E_{\theta,j}:=\{\widehat\theta_j=\theta_j\}$. Since exact recovery implies correct recovery of coordinate $j$, $\mathsf P_\theta^\pi(E_{\theta,j})\geq1-\delta$. Under $\theta^{(j)}$, the true $j$th coordinate is $-\theta_j$, so the same event $E_{\theta,j}$ is an error and therefore $\mathsf P_{\theta^{(j)}}^\pi(E_{\theta,j})\leq\delta$. Applying data processing to the map from the observed history to $\mathbf{1}\{E_{\theta,j}\}$ gives
\[
    D_{\mathrm{KL}}
    \left(
        \mathsf P_\theta^\pi
        \,\|\,
        \mathsf P_{\theta^{(j)}}^\pi
    \right)
    \geq
    \operatorname{kl}\left(
        \mathsf P_\theta^\pi(E_{\theta,j}),
        \mathsf P_{\theta^{(j)}}^\pi(E_{\theta,j})
    \right).
\]
Since binary KL is increasing in its first argument and decreasing in its second whenever the first exceeds the second,
\begin{align*}
    \operatorname{kl}\left(
        \mathsf P_\theta^\pi(E_{\theta,j}),
        \mathsf P_{\theta^{(j)}}^\pi(E_{\theta,j})
    \right)
    &\geq
    \operatorname{kl}(1-\delta,\delta) \\
    &=
    (1-\delta)\log\frac{1-\delta}{\delta}
    +
    \delta\log\frac{\delta}{1-\delta} \\
    &=
    (1-2\delta)\log\frac{1-\delta}{\delta}\\
    &\geq
    \frac12\log(1/\delta),
\end{align*}
where the last inequality holds for $\delta\leq1/8$. Therefore,
\[
    D_{\mathrm{KL}}
    \left(
        \mathsf P_\theta^\pi
        \,\|\,
        \mathsf P_{\theta^{(j)}}^\pi
    \right)
    \geq
    \frac12\log(1/\delta).
\]
Summing over $j$ and averaging over $\theta$ gives $\mathcal I\ge(d/2)\log(1/\delta)$. Combining the two bounds gives $\mathcal I\gtrsim d\log(d/\delta)$. Finally, the standard divergence decomposition for adaptive bandits \citep[Lemma~15.1]{LS20} gives
\[
    D_{\mathrm{KL}}
    \left(
        \mathsf P_\theta^\pi
        \,\|\,
        \mathsf P_{\theta^{(j)}}^\pi
    \right)
    =
    \E_\theta^\pi
    \left[
        \sum_{t\in\mathcal S_j}
        \mathbf 1\{A_t=a_j\}
        \operatorname{kl}(p_t,q_t)
    \right],
\]
where $p_t$ and $q_t$ denote the Bernoulli means under $\theta$ and $\theta^{(j)}$, respectively. Indeed, the two environments agree at every arm--round pair except $(t,a_j)$ for $t\in\mathcal S_j$, so all other per-round KL terms vanish. For Bernoulli distributions, using $\log x\leq x-1$,
\begin{align*}
    \operatorname{kl}(p,q)
    =
    p\log\frac pq
    +(1-p)\log\frac{1-p}{1-q} \leq
    \frac{(p-q)^2}{q(1-q)}.
\end{align*}
Since $p,q\in[3/16,5/16]$, we have $q(1-q)\geq39/256>1/8$, while $|p-q|\leq2\Delta$. Hence $\operatorname{kl}(p,q)\leq32\Delta^2$, and therefore
\[
    D_{\mathrm{KL}}
    \left(
        \mathsf P_\theta^\pi
        \,\|\,
        \mathsf P_{\theta^{(j)}}^\pi
    \right)
    \leq
    32\Delta^2
    \E_\theta^\pi[N_j].
\]
Averaging over $\theta$ and summing over $j$ gives
\[
    \mathcal I
    \leq
    32\Delta^2
    2^{-d}
    \sum_{\theta}
    \E_\theta^\pi
    \left[
        \sum_{j\in\mathcal J}N_j
    \right].
\]
Combining this with $\mathcal I\gtrsim d\log(d/\delta)$ proves
\[
    2^{-d}
    \sum_{\theta}
    \E_\theta^\pi
    \left[
        \sum_{j\in\mathcal J}N_j
    \right]
    \gtrsim
    \frac{d}{\Delta^2}
    \log\left(\frac{d}{\delta}\right),
\]
which proves \cref{eq:simultaneous-sign-recovery}.
\end{proof}

We also use the following H\"older bump calculation. Let $\psi_i(u):=(1-|Tu-s_i|/\ell)_+^\beta$ with pairwise disjoint support interiors. Since $|x^\beta-y^\beta|\le|x-y|^\beta$ for $\beta\in(0,1]$, each $\psi_i$ is $\beta$-H\"older with constant $(T/\ell)^\beta$. If $u$ and $v$ lie in different supports, choose a point $z$ between them where all bumps vanish. Then
\begin{align}
    \left|
        \Delta\sum_i\theta_i\psi_i(u)
        -
        \Delta\sum_i\theta_i\psi_i(v)
    \right|
    &\le
    \Delta\left(\frac{T}{\ell}\right)^\beta
    \left(|u-z|^\beta+|v-z|^\beta\right) \notag \\
    &\le
    2\Delta\left(\frac{T}{\ell}\right)^\beta
    |u-v|^\beta.
    \label{eq:holder-bump-bound}
\end{align}
Thus the sum of disjoint bumps has H\"older constant at most $2\Delta(T/\ell)^\beta$.

\paragraph{Inference lower bound.}
We first consider the regime $L=0$ or
$(2w/L)^{1/\beta}>1/32$. For $\theta\in\{-1,1\}^{\cA_+}$, define $\mu_0^\theta\equiv3/4$ and
$\mu_a^\theta\equiv1/4+w\theta_a$ for $a\in\cA_+$. Then $\tau_t^\theta(a)=-1/2+w\theta_a$. For each $a\in\cA_+$, the two possible current effects are $-1/2-w$ and $-1/2+w$, which are separated by $2w$. On the validity event, $\mathscr{C}_{t}(a)$ contains the true effect and has width at most $w$, so it cannot contain both possible values and therefore uniquely determines $\theta_a$. Since this holds simultaneously for every $a\in\cA_+$, the intervals $\{\mathscr{C}_{t}(a):a\in\cA_+\}$ recover the full sign vector $\theta$. Applying Lemma~\ref{lem:simultaneous-sign-recovery} with $d=K-1$ and $\Delta=w$ gives
\[
    2^{-(K-1)}
    \sum_\theta
    \E_\theta^\pi
    \left[
        \sum_{a\in\cA_+}N_a(T)
    \right]
    \gtrsim
    \frac{K}{w^2}\log\left(\frac{K}{\delta}\right).
\]
Arm $0$ is optimal and every pull of $a\in\cA_+$ incurs regret at least
$1/4$. Hence
\begin{equation}
    \sup_{\mu\in\cM_{\beta,L}}
    \mathscr R_T(\pi,\mu)
    \gtrsim 2^{-(K-1)}
    \sum_\theta
    \E_\theta^\pi
    \left[
        \sum_{a\in\cA_+}N_a(T)
    \right] \gtrsim
    \frac{K}{w^2}\log\left(\frac{K}{\delta}\right).
    \label{eq:stationary-inference-lower}
\end{equation}
When $(2w/L)^{1/\beta}>1/32$,
$1+(L/w)^{1/\beta}\lesssim_\beta1$, so
\cref{eq:stationary-inference-lower} already has the required rate.

Now suppose $(2w/L)^{1/\beta}\le1/32$. If
$T(2w/L)^{1/\beta}\le1$, no $(w,\delta)$-valid procedure exists. Indeed, let $t_0=\lceil T/2\rceil$, set $\psi(u):=(1-T|u-t_0/T|)_+^\beta$, and define
$\mu_0^\theta\equiv3/4$ and $\mu_1^\theta(u)=1/4+w\theta\psi(u)$ for $\theta\in\{-1,1\}$. By \cref{eq:holder-bump-bound}, the H\"older constant is at most $2wT^\beta\le L$. The two environments agree on $\mathscr{F}_{t_0-1}$, while their current effects at $t_0$ differ by $2w$. Since $\mathscr{C}_{t_0}(1)$ is $\mathscr{F}_{t_0-1}$-measurable, an interval of width at most $w$ cannot cover both values with probability at least $1-\delta$, contradicting validity. We therefore consider $1<T(2w/L)^{1/\beta}$ and $(2w/L)^{1/\beta}\le1/32$. Set $\ell:=\lceil T(2w/L)^{1/\beta}\rceil$, $q:=\lfloor T/(4\ell)\rfloor$, and $s_i:=\lfloor T/2\rfloor+(2i-1)\ell$ for $i\in[q]$. Then $\ell\le T/16$ and $q \asymp_\beta 1+(L / w )^{1/\beta}$.

For $\theta\in\{-1,1\}^{[q]\times\cA_+}$, define
\begin{equation}
    \mu_0^\theta(u)=\frac34,
    \;\;
    \mu_a^\theta(u)
    =
    \frac14
    +
    w\sum_{i=1}^q
    \theta_{i,a}\psi_i(u),
    \;\;
    a\in\cA_+.
    \label{eq:inference-local-alternatives}
\end{equation}
The support interiors are disjoint and $2w(T/\ell)^\beta\le L$, so $\mu^\theta\in\cM_{\beta,L}$. Also, $\mu_a^\theta(u)\in[3/16,5/16]$, while arm $0$ is always optimal by at least $1/4$. At $s_i$,
$\tau_{s_i}^\theta(a)=-1/2+w\theta_{i,a}$. Hence the intervals $\{\mathscr{C}_{s_i}(a):i\in[q],a\in\cA_+\}$ must recover all $d=q(K-1)$ signs of $\theta$. 
Lemma~\ref{lem:simultaneous-sign-recovery} with $\Delta=w$ gives
\begin{align}
    \sup_{\mu\in\cM_{\beta,L}}
    \mathscr R_T(\pi,\mu)
    &\gtrsim
    \frac{q(K-1)}{w^2}
    \log\left(\frac{q(K-1)}{\delta}\right) \notag \\
    &\gtrsim_\beta
    \frac{K}{w^2}
    \left[
        1+\left(\frac{L}{w}\right)^{1/\beta}
    \right]
    \log\left(
        \frac{K[1+(L/w)^{1/\beta}]}{\delta}
    \right).
    \label{eq:inference-lower-final}
\end{align}

\paragraph{Reward-learning lower bound.}
We next prove the standard reward-learning terms. The construction is the localized testing argument used for smooth nonstationary bandits
\citep{JXKF23,Suk25}.

For the stationary term, let
$\varepsilon:=\frac1{16}\sqrt{K/T}$. Define the null environment by $\mu_b^0\equiv1/2$ for every $b\in\cA$. For each $a\in\cA$, define $\mu_a^a\equiv1/2+\varepsilon$ and $\mu_b^a\equiv1/2$ for $b\neq a$. Let $N_a(T):=\sum_{t=1}^T\mathbf 1\{A_t=a\}$. By the adaptive divergence decomposition \citep[Lemma~15.1]{LS20}, $D_{\mathrm{KL}}(\mathsf P_0^\pi\|\mathsf P_a^\pi) \le8\varepsilon^2\E_0^\pi[N_a(T)]$. Since $\sum_{a\in\cA}\E_0^\pi[N_a(T)]=T$, Pinsker's inequality and Jensen's inequality give
\[
    \frac1K
    \sum_{a\in\cA}
    \mathsf{TV}(\mathsf P_0^\pi,\mathsf P_a^\pi)
    \le
    2\varepsilon\sqrt{\frac{T}{K}}
    =
    \frac18.
\]
Since $0\le N_a(T)\le T$, it follows that $\frac1K\sum_{a\in\cA}\E_a^\pi[N_a(T)] \leq 5T/8$. Thus, $\sup_{\mu\in\cM_{\beta,L}}\mathscr R_T(\pi,\mu)\gtrsim\sqrt{KT}$.

For the nonstationary term, suppose $L\ge\sqrt{K/T}$ and set
\begin{equation}
    \ell
    :=
    \left\lceil
        \frac14
        \left(
            \frac{KT^{2\beta}}{L^2}
        \right)^{1/(2\beta+1)}
    \right\rceil,
    \;\;
    \Delta
    :=
    \frac1{128}\sqrt{\frac{K}{\ell}},
    \;\;
    q
    :=
    \left\lfloor\frac{T}{2\ell}\right\rfloor.
    \label{eq:reward-lower-tuning}
\end{equation}
Place bumps at $s_i=(2i-1)\ell$ and let $\psi_i(u):=(1-|Tu-s_i|/\ell)_+^\beta$. For $\theta=(\theta_1,\ldots,\theta_q)\in\cA^q$, define $\mu_a^\theta(u):=\frac12+\Delta\sum_{i=1}^q\mathbf 1\{\theta_i=a\}\psi_i(u)$ for $a\in\cA$. Since $\ell\ge\frac14(KT^{2\beta}/L^2)^{1/(2\beta+1)}$, we have $2\Delta(T/\ell)^\beta\le L$. Thus $\mu^\theta\in\cM_{\beta,L}$.

Fix $i\in[q]$ and the coordinates $\theta_{-i}$. Let
$\mathsf P_{-i}^\pi$ be the law with the $i$-th bump removed, and let
$\mathsf P_{i,a}^\pi$ be the law with $\theta_i=a$. Define
$\mathcal S_i:=\{t:|t-s_i|<\ell\}$ and
$N_{i,a}:=\sum_{t\in\mathcal S_i}\mathbf 1\{A_t=a\}$. The adaptive divergence
decomposition gives
\[
    D_{\mathrm{KL}}
    \left(
        \mathsf P_{-i}^\pi
        \,\|\,
        \mathsf P_{i,a}^\pi
    \right)
    \le
    8\Delta^2\E_{-i}^\pi[N_{i,a}].
\]
Since $\sum_a\E_{-i}^\pi[N_{i,a}]\le2\ell$, Pinsker's inequality and Jensen's
inequality give
\begin{equation}
    \frac1K
    \sum_{a\in\cA}
    \mathsf{TV}
    \left(
        \mathsf P_{-i}^\pi,
        \mathsf P_{i,a}^\pi
    \right)
    \le
    \sqrt{\frac{8\Delta^2\ell}{K}}
    \le
    \frac18.
    \label{eq:local-tv}
\end{equation}

Let $\mathcal C_i:=\{t:|t-s_i|\le\ell/2\}$. Then
$|\mathcal C_i|\ge\ell$ and $\psi_i(t/T)\ge1/2$ on $\mathcal C_i$. Under
$\mathsf P_{i,a}^\pi$, arm $a$ is therefore optimal on $\mathcal C_i$ with
gap at least $\Delta/2$. Let
$M_{i,a}:=\sum_{t\in\mathcal C_i}\mathbf 1\{A_t=a\}$. Using
\cref{eq:local-tv},
\[
    \frac1K
    \sum_{a\in\cA}
    \E_{i,a}^\pi[M_{i,a}]
    \le
    \frac{|\mathcal C_i|}{K}
    +
    \frac{|\mathcal C_i|}{8}
    \le
    \frac{5}{8}|\mathcal C_i|.
\]
Hence the average regret on the $i$-th bump is at least a constant multiple of
$\ell\Delta$. Averaging over $\theta\in\cA^q$ and summing over $i$ gives $\sup_{\mu\in\cM_{\beta,L}}
\mathscr R_T(\pi,\mu) \gtrsim q\ell\Delta$.
When $L\ge\sqrt{K/T}$, the choice in
\cref{eq:reward-lower-tuning} satisfies $\ell\lesssim T$, so
$q\ell\gtrsim T$. Therefore,
\begin{align}
\label{eq:nonstationary-reward-lower}
    \sup_{\mu\in\cM_{\beta,L}}
    \mathscr R_T(\pi,\mu)
    &\gtrsim
    T\sqrt{\frac{K}{\ell}} \gtrsim
    K^{\beta/(2\beta+1)}
    L^{1/(2\beta+1)}
    T^{(\beta+1)/(2\beta+1)}.
\end{align}

When $L<\sqrt{K/T}$, the nonstationary term in \cref{eq:nonstationary-reward-lower} is at most $\sqrt{KT}$, so the stationary lower bound already gives the required reward-learning rate.

Combining \cref{eq:inference-lower-final} and \cref{eq:nonstationary-reward-lower} with the stationary lower bound above, and using that the maximum of three nonnegative quantities is at least one third of their sum, gives
\begin{align*}
    \mathscr R_T^\star(w,\delta)
    \gtrsim_\beta
    \sqrt{KT}
    +
    K^{\beta/(2\beta+1)}
    L^{1/(2\beta+1)}
    T^{(\beta+1)/(2\beta+1)}
    +
    \frac{K}{w^2}
    \left[
        1+\left(\frac{L}{w}\right)^{1/\beta}
    \right]
    \log\left(
        \frac{K[1+(L/w)^{1/\beta}]}{\delta}
    \right).
\end{align*}
Together with the upper bound, this proves Theorem~\ref{thm:tradeoff}.

\section{Proof of Theorem~\ref{thm:upper-bound}}
\label{appendix-sec:proof-of-nonadaptive-upper-bound}
We first establish that \textsc{TRACE} is a $(w, \delta)$-valid procedure. Recall the following notation: $\mathcal{S}_{j, a} \subseteq \mathcal{I}_{j}$ denotes the rounds reserved for inference for arm $a$ in block $\mathcal{B}_{j}$. For $t \in \mathcal{B}_{j}$, we compute the estimated current-effect $\widehat{\tau}_{j-1}(a) = \widehat{\mu}_{j-1, a} - \widehat{\mu}_{j-1, 0}$ for $a \in \mathcal{A}_{+}$ where $\widehat{\mu}_{j, a} = m^{-1} \sum_{s \in \mathcal{S}_{j, a}} Y_{s}$. 

Throughout the proof, we will $\overline{\mu}_{j, a} = m^{-1} \sum_{s \in \mathcal{S}_{j, a}} \mu_{s, a}$ to denote the mean reward for arm $a$ in block $\mathcal{B}_{j}$. First note that we can decompose the difference in the estimated and true current-effects as follows:
\begin{align}
    \widehat{\tau}_{j-1}(a) - \tau_t(a)
    &= \widehat{\mu}_{j-1,a} - \widehat{\mu}_{j-1,0}
    - \mu_{t,a} + \mu_{t,0} \notag \\
    &= \underbrace{
        \left(\widehat{\mu}_{j-1,a} - \overline{\mu}_{j-1,a}\right)
        -
        \left(\widehat{\mu}_{j-1,0} - \overline{\mu}_{j-1,0}\right)
    }_{(i)}
    +
    \underbrace{
        \left(\overline{\mu}_{j-1,a} - \mu_{t,a}\right)
        -
        \left(\overline{\mu}_{j-1,0} - \mu_{t,0}\right)
    }_{(ii)}.
    \label{eq:estimated-true-effect-difference}
\end{align}
We first bound (i). Consider the quantity $Z_{\ell} = Y_{\ell} - \mu_{s_{\ell}, a}$ for $s_{\ell} \in \mathcal{S}_{j, a}$ and $a \in \mathcal{A}$. Conditionally on $A_{s_{\ell}} = a$, we have $Y_{\ell} \mid \mathscr{F}_{s_{\ell} - 1}, A_{s_{\ell}} = a \sim \mathrm{Bern}(\mu_{s_{\ell}, a})$. Thus $\mathbb{E} \left[ Y_{\ell} \mid \mathscr{F}_{s_{\ell} - 1}, A_{s_{\ell}} = a \right] = \mu_{s_{\ell}, a}$ so $\mathbb{E} \left[ Z_{\ell} \mid \mathscr{F}_{s_{\ell} - 1} \right] = 0$. Furthermore since $\lvert Z_{\ell} \rvert \leq 1$, we note that $(Z_{\ell})_{\ell=1}^{m}$ is a martingale difference sequence (MDS) with bounded increments being atmost $1$. Therefore by the Azuma-Hoeffding inequality \citep[Corollary 2.20]{wainwright2019high} and a union bound over all $j \in [J], a \in \mathcal{A}$, we establish:
\begin{equation}
\label{eq:azuma-hoeffding-mu}
\mathbb{P} \left( \forall (j, a) \in [J] \times \ \mathcal{A},  \lvert \widehat{\mu}_{j, a} - \overline{\mu}_{j, a} \rvert \leq \sqrt{\frac{1}{2m} \log \left( \frac{2KJ}{\delta} \right)} \right) \geq 1 -\delta.
\end{equation}

We now bound (ii). Since $\mu_{a} \in \mathcal{M}_{\beta, L}$, the following holds:
\begin{align}
    \label{eq:holder-drift-mu}
    \overline{\mu}_{j, a} - \mu_{t, a} = \frac{1}{m} \sum_{s \in \mathcal{S}_{j, a}} ( \mu_{s, a} - \mu_{t, a} ) \leq \frac{1}{m} \sum_{s \in \mathcal{S}_{j, a}} L \left\lvert \frac{s - t}{T}\right\rvert^{\beta}.
\end{align}
Since $t \in \mathcal{B}_{j}$ and $s \in \mathcal{S}_{j, a} \subseteq \mathcal{B}_{j-1}$, it follows that $\lvert s - t \rvert \leq \lvert \mathcal{B}_{j-1} \rvert + \lvert \mathcal{B}_{j} \rvert \leq 2 \lceil T/J \rceil$. Given that $\lvert \mathcal{S}_{j, a} \rvert = m$, it follows that $\overline{\mu}_{j, a} - \mu_{t, a} \leq L (4 / J)^{\beta}$. Using \cref{eq:azuma-hoeffding-mu} and \cref{eq:holder-drift-mu} with \eqref{eq:estimated-true-effect-difference}, we find with probability atleast $1-\delta$:
\begin{equation}
    \label{eq:current-estimated-effect-bound-final}
    \lvert \widehat{\tau}_{j-1}(a) - \tau_{t}(a) \rvert \leq \sqrt{\frac{2}{m} \log \left( \frac{2KJ}{\delta} \right)} + 2L \left( \frac{4}{J} \right)^{\beta}.
\end{equation}
Thus it suffices to take a confidence interval of the form:
\begin{equation}
    \mathscr{C}_{t}(a)
    :=
    \left[
        \widehat\tau_{j-1}(a)
        \pm
        \left\{
            \sqrt{
                \frac{2}{m}
                \log\left(\frac{2KJ}{\delta}\right)
            }
            +
            2L\left(\frac{4}{J}\right)^\beta
        \right\}
    \right]
    \cap[-1,1],
    \;\;
    a\in\cA_+.
\end{equation}
This is exactly the confidence interval proposed by \textsc{TRACE} (see \cref{eq:confidence-interval}). We now require $\lvert \mathscr{C}_{t}(a) \rvert \leq w$. Thus it suffices to find $J$ and $m$ such that $\widehat{\tau}_{j-1}(a) - \tau_{t}(a) \leq w$. One finds $J^{\star} \asymp (L / w)^{1/\beta}$ and $m^{\star} \asymp \log (KJ^{\star} / \delta)/w^{2}$ suffices up to constants (see \cref{eq:inference-tuning}). Under such a choice of $J^{\star}$ and $m^{\star}$ with \cref{eq:current-estimated-effect-bound-final}, we can conclude that \textsc{TRACE} is a $(w, \delta)$-valid procedure.

We now show that \textsc{TRACE} achieves the minimax optimal regret rate under class of all $(w, \delta)$-valid policies with H\"older-drift class $\mathcal{M}_{\beta, L}$. First we recall the following notation: Within block $\mathcal{B}_{j}$, we reserve the first $Km$ rounds for statistical inference and denote these reserved rounds $\mathcal{I}_{j}$. The other $\lceil T/J \rceil - Km$ rounds are used for reward-learning, denoted $\mathcal{G}_{j} = \mathcal{B}_{j} \setminus \mathcal{I}_{j}$. We denote $\mathcal{I} = \cup_{j \in [J]} \mathcal{I}_{j}$ the set of all reserved rounds in horizon $T$ and $\mathcal{G} = \cup_{j \in [J]} \mathcal{G}_{j}$ the set of all non-reserved rounds. By definition, we have $[T] = \mathcal{I} \sqcup \mathcal{G}$. 

Additionally, we partition $[T]$ into consecutive reward-restart blocks $\{\mathcal{H}_{r}\}_{r=1}^{R}$ of size $h$ (with the final block possibly shorter), where $R := \lceil T/h \rceil$. Thus, $[T] = \bigcup_{r=1}^{R} \mathcal{H}_{r}$. By construction of \textsc{TRACE}, on each round $t \in \mathcal{H}_{r} \cap \mathcal{G}$, the algorithm follows the \textsc{EXP3} update in \cref{eq:exp3-induced} (Line~22). We define $\mathcal{N}_{r} := \mathcal{H}_{r} \cap \mathcal{G}$. Since the blocks $\{\mathcal{H}_{r}\}_{r=1}^{R}$ are pairwise disjoint, the sets $\{\mathcal{N}_{r}\}_{r=1}^{R}$ are also pairwise disjoint. Moreover, $\bigcup_{r=1}^{R} \mathcal{N}_{r} = \mathcal{G}$. Thus we find that $[T] = \mathcal{I} \sqcup \left( \bigcup_{r=1}^{R} \mathcal{N}_{r} \right) = \left( \cup_{j \in [J]} \cup_{a \in \mathcal{A}} \mathcal{S}_{j, a}\right) \sqcup \left( \bigcup_{r=1}^{R} \mathcal{N}_{r} \right)$.

Let $a_{t}^{\star} = \arg \max_{a \in \mathcal{A}} \mu_{t, a}$ denote the best arm at round $t$, and let $s_{r} := \min \mathcal{H}_{r}$ denote the first round of reward-restart block $\mathcal{H}_{r}$. Then we bound the dynamic regret as follows:
\begin{align}
    \E_{\mu}^{\pi} \left[ \sum_{t=1}^{T} \max_{a \in \mathcal{A}} \mu_{t, a} - \mu_{t, A_{t}} \right] &= \underbrace{\E_{\mu}^{\pi} \left[ \sum_{j \in [J]} \sum_{a \in \mathcal{A}} \sum_{s \in \mathcal{S}_{j, a}} \max_{a \in \mathcal{A}} \mu_{s, a} - \mu_{s, A_{s}} \right]}_{(i)} + \underbrace{\sum_{r \in [R]} \E_{\mu}^{\pi} \left[ \sum_{t \in \mathcal{N}_{r}} \max_{a \in \mathcal{A}} \mu_{t, a} - \mu_{t, a_{s_{r}}^{\star}} \right]}_{(ii)} \notag \\
    &+ \underbrace{\sum_{r \in [R]} \E_{\mu}^{\pi} \left[ \sum_{t \in \mathcal{N}_{r}} \mu_{t, a_{s_{r}}^{\star}} - \mu_{t, A_{t}} \right]}_{(iii)},
\end{align}
where the decomposition follows from the partition of $[T]$ into inference and reward-learning rounds and, for each reward-restart block $\mathcal{H}_{r}$, adding and subtracting $\mu_{t, a_{s_{r}}^{\star}}$. Term (i) corresponds to the regret we pay for pulling all arms while we are in an inference block, term (ii) corresponds to the regret incurred from using the best arm at the start of each reward-restart block as a fixed comparator as the environment drifts within the block, and term (iii) corresponds to the \textsc{EXP3} regret relative to this fixed comparator. 

We now bound (iii). We recall the relevant quantities required for the analysis. We denote the importance-weighted estimator $\hat{\ell}_{t}(a) = \mathbf 1\left\{ A_{t} = a \right\} (1-Y_{t})/ p_{t}(a)$ where $p_{t}(a)$ is the induced sampling distribution using exponential weighting. For completeness, we briefly recall the \textsc{EXP3} argument used on each restart block. See \citet[Theorem~11.1]{LS20} for a standard treatment. Within $\mathcal{H}_r$, the algorithm initializes $w_t(a)=1$ for every $a\in\mathcal A$, forms $p_t(a)=w_t(a)/\sum_{b\in\mathcal A}w_t(b)$, and, on each reward-learning round $t\in\mathcal N_r$, updates
$w_{t+1}(a)=w_t(a)\exp\{-\eta\hat{\ell}_t(a)\}$.

Let $W_t:=\sum_{a\in\mathcal A}w_t(a)$. For each $t\in\mathcal N_r$, using $e^{-x}\leq1-x+x^2/2$ for $x\geq0$ and $\log(1+x)\leq x$, we obtain
\begin{align*}
    \log\frac{W_{t+1}}{W_t}
    &= \log\left(
        \sum_{a\in\mathcal A}
        p_t(a)e^{-\eta\hat{\ell}_t(a)}
    \right) \\
    &\leq
    -\eta\sum_{a\in\mathcal A}p_t(a)\hat{\ell}_t(a)
    +
    \frac{\eta^2}{2}
    \sum_{a\in\mathcal A}p_t(a)\hat{\ell}_t(a)^2.
\end{align*}
Summing over $t\in\mathcal N_r$ telescopes the left-hand side. Also, for any fixed arm $a$, its final weight satisfies
$W_{\mathrm{end}}\geq w_{\mathrm{end}}(a)
=\exp\{-\eta\sum_{t\in\mathcal N_r}\hat{\ell}_t(a)\}$, while $W_{\mathrm{start}}=K$. Taking $a=a_{s_r}^{\star}$ and combining the lower and upper bounds on the potential gives
\begin{equation}
    \sum_{t \in \mathcal{N}_{r}}
    \left\{
        \sum_{a \in \mathcal{A}}
        p_{t}(a) \hat{\ell}_{t}(a)
        -
        \hat{\ell}_{t}(a_{s_{r}}^{\star})
    \right\}
    \leq
    \frac{\log K}{\eta}
    +
    \frac{\eta}{2}
    \sum_{t \in \mathcal{N}_{r}}
    \sum_{a \in \mathcal{A}}
    p_{t}(a) \hat{\ell}_{t}(a)^2.
\end{equation}

We now translate this estimated-loss bound into expected regret. Recall that
$\hat{\ell}_t(a)=\mathbf{1}\{A_t=a\}(1-Y_t)/p_t(a)$. Since $A_t\sim p_t$ conditional on $\mathscr{F}_{t-1}$ and $\E_\mu^\pi[Y_t\mid\mathscr{F}_{t-1},A_t=a]=\mu_{t,a}$, for every $a\in\mathcal A$ we have $\E_\mu^\pi[\hat{\ell}_t(a)\mid\mathscr{F}_{t-1}]=1-\mu_{t,a}$. Consequently,
$\E_\mu^\pi[\sum_a p_t(a)\hat{\ell}_t(a)\mid\mathscr{F}_{t-1}]
=1-\E_\mu^\pi[\mu_{t,A_t}\mid\mathscr{F}_{t-1}]$ and
$\E_\mu^\pi[\hat{\ell}_t(a_{s_r}^{\star})\mid\mathscr{F}_{t-1}]
=1-\mu_{t,a_{s_r}}^{\star}$. Thus, by the tower property,
\begin{align*}
    \E_{\mu}^{\pi}
    \left[
        \sum_{t \in \mathcal{N}_{r}}
        \mu_{t, a_{s_{r}}^{\star}}
        -
        \mu_{t, A_{t}}
    \right]
    &=
    \E_{\mu}^{\pi}
    \left[
        \sum_{t \in \mathcal{N}_{r}}
        \left\{
            \sum_{a \in \mathcal{A}}
            p_{t}(a)\hat{\ell}_{t}(a)
            -
            \hat{\ell}_{t}(a_{s_{r}}^{\star})
        \right\}
    \right] \\
    &\leq
    \frac{\log K}{\eta}
    +
    \frac{\eta}{2}
    \E_{\mu}^{\pi}
    \left[
        \sum_{t \in \mathcal{N}_{r}}
        \sum_{a \in \mathcal{A}}
        p_{t}(a)\hat{\ell}_{t}(a)^2
    \right].
\end{align*}

It remains to control the second-order term. For each $a\in\mathcal A$,
\[
    \E_\mu^\pi
    \left[
        p_t(a)\hat{\ell}_t(a)^2
        \mid
        \mathscr{F}_{t-1}
    \right]
    =
    \E_\mu^\pi
    \left[
        \frac{\mathbf{1}\{A_t=a\}(1-Y_t)^2}{p_t(a)}
        \,\middle|\,
        \mathscr{F}_{t-1}
    \right]
    \leq 1,
\]
where we used $(1-Y_t)^2\leq1$. Summing over the $K$ arms gives
$\E_\mu^\pi[\sum_a p_t(a)\hat{\ell}_t(a)^2\mid\mathscr{F}_{t-1}]\leq K$. Therefore,
$\E_\mu^\pi[\sum_{t\in\mathcal N_r}\sum_a p_t(a)\hat{\ell}_t(a)^2]\leq K|\mathcal N_r|$. Since $|\mathcal N_r|\leq|\mathcal H_r|\leq h$, we obtain
\[
    \E_{\mu}^{\pi}
    \left[
        \sum_{t \in \mathcal{N}_{r}}
        \mu_{t, a_{s_{r}}^{\star}}
        -
        \mu_{t, A_{t}}
    \right]
    \leq
    \frac{\log K}{\eta}
    +
    \frac{\eta Kh}{2}.
\]
Taking $\eta=\sqrt{2\log K/(Kh)}$ yields
\[
    \E_{\mu}^{\pi}
    \left[
        \sum_{t \in \mathcal{N}_{r}}
        \mu_{t, a_{s_{r}}^{\star}}
        -
        \mu_{t, A_{t}}
    \right]
    \leq
    \sqrt{2Kh\log K}.
\]

We now bound (ii). Consider $a \in \mathcal{A}$ and $t \in \mathcal{N}_{r}$. Then the following holds:
\begin{align}
    \mu_{t, a} - \mu_{t, a_{s_{r}}^{\star}} &= \mu_{t, a} - \mu_{s_{r}, a} + \mu_{s_{r}, a} - \mu_{s_{r}, a_{s_{r}}^{\star}} + \mu_{s_{r}, a_{s_{r}}^{\star}} - \mu_{t, a_{s_{r}}^{\star}} \notag \\ 
    &\leq \mu_{t, a} - \mu_{s_{r}, a} + \mu_{s_{r}, a_{s_{r}}^{\star}} - \mu_{t, a_{s_{r}}^{\star}} \notag \\
    &\leq 2L \left( \frac{h}{T} \right)^{\beta}.
\end{align}
where the first inequality holds from the optimality of $a_{s_{r}}^{\star}$ at round $s_{r}$ and the last inequality holds since $\mu_{a} \in \mathcal{M}_{\beta, L}$ and $\lvert t-s_{r}\rvert \leq h$. Finally, we bound (i) by simply noting that for any fixed $s \in [T]$, $\max_{a \in \mathcal{A}} \mu_{s, a} - \mu_{s, A_{s}} \leq 1$. Combining yields:
\begin{align}
    \E_{\mu}^{\pi} \left[ \sum_{t=1}^{T} \max_{a \in \mathcal{A}} \mu_{t, a} - \mu_{t, A_{t}} \right] &\leq KJ \lvert \mathcal{S}_{j, a} \rvert + R\sqrt{2Kh\log K} + 2L \left( \frac{h}{T} \right)^{\beta} \sum_{r \in [R]} \lvert \mathcal{N}_{r} \rvert  \\
    &\leq KJm + \sqrt{2K\log K} \left( \frac{T}{\sqrt{h}} + \sqrt{h} \right) + 2LT \left( \frac{h}{T} \right)^{\beta},
\end{align}
where the last inequality follows from $\lvert \mathcal{S}_{j,a} \rvert = m$, $\sum_{r \in [R]} \lvert \mathcal{N}_{r} \rvert = \lvert \mathcal{G} \rvert \leq T$, and $R = \lceil T/h \rceil \leq T/h + 1$. Balancing the two leading terms by taking $T\sqrt{K\log K/h} \asymp LT(h/T)^{\beta}$, equivalently $h \asymp_{\beta} \left(T^{2\beta}K\log K/L^{2}\right)^{1/(2\beta+1)}$, and imposing $h \leq T$, we take $h \asymp_{\beta} T \wedge \left(T^{2\beta}K\log K/L^{2}\right)^{1/(2\beta+1)}$. Noting that $h \leq T$ implies $\sqrt{Kh\log K} \leq T\sqrt{K\log K/h}$, we conclude that
\begin{align}
    \E_{\mu}^{\pi} \left[ \sum_{t=1}^{T} \max_{a \in \mathcal{A}} \mu_{t, a} - \mu_{t, A_{t}} \right]
    \lesssim_{\beta} KJm + \sqrt{KT\log K} + (K\log K)^{\frac{\beta}{2\beta+1}} L^{\frac{1}{2\beta+1}} T^{\frac{\beta+1}{2\beta+1}}.
\end{align}
This concludes the proof.

\section{Proof of Lemma~\ref{lem:eprocess-validity}}
\label{appendix-sec:proof-of-eprocess-validity}
Before proving Lemma~\ref{lem:eprocess-validity}, we first establish a preliminary result showing that the stochastic processes used in our analysis are non-negative supermartingales. Throughout this section, we use the notation
$I_{s,a}:=F_{s}\mathbf 1\{A_s=a\}$,
$N_{j,a}(t):=\sum_{s=b_j}^{t} I_{s,a}$,
$V_{j,a}(t):=\sum_{s=b_j}^{t} I_{s,a}(Y_s-\widehat Y_{s,a})^2$,
$\overline{Y}_{j,a}(t):=N_{j,a}(t)^{-1}\sum_{s=b_j}^{t} I_{s,a}Y_s$ whenever $N_{j,a}(t)>0$,
and $d_J:=L(2/J)^\beta$.

\begin{lem}
    \label{appendix-lem:supermartingale-result}
    Fix a block $j \in [J]$, an arm $a \in \mathcal A$, and $\lambda \in [0,1)$. Consider the stochastic processes
    \begin{align}
        E_{j,a,t}^{\pm}(\theta;\lambda)
        &=
        \exp\left\{
        \lambda \sum_{s=b_j}^{t}
        I_{s,a}\bigl[\pm(Y_s-\theta)-d_J\bigr]
        -
        \psi(\lambda)V_{j,a}(t)
        \right\}, \\
        H_{j,a,t}^{\pm}(\theta)
        &=
        \exp\left\{
        \frac{w}{4}
        \sum_{s=b_j}^{t}
        I_{s,a}\bigl[\pm(Y_s-\theta)-d_J\bigr]
        -
        \frac{w^2}{128}N_{j,a}(t)
        \right\}.
    \end{align}
    Under the null hypothesis $H_0(\theta): \theta=\mu_{b_j,a}$, and assuming $\widehat Y_{s,a}\in[0,1]$ is $\mathscr{F}_{s-1}$-measurable, both
    $\{E_{j,a,t}^{\pm}(\theta;\lambda)\}_{t\ge b_j-1}$ and
    $\{H_{j,a,t}^{\pm}(\theta)\}_{t\ge b_j-1}$ are nonnegative
    $\{\mathscr{F}_t\}$-supermartingales, each initialized at one, i.e.,
    $E_{j,a,b_j-1}^{\pm}(\theta;\lambda)=H_{j,a,b_j-1}^{\pm}(\theta)=1$.
\end{lem}

\begin{proof}[Proof of Lemma~\ref{appendix-lem:supermartingale-result}]
We define the quantity
$R_{j,a,t}^{\pm}(\theta;\lambda) :=
E_{j,a,t}^{\pm}(\theta;\lambda)/
E_{j,a,t-1}^{\pm}(\theta;\lambda)$.
We will show that $\mathbb{E}[R_{j,a,t}^{\pm}(\theta;\lambda)\mid\mathscr{F}_{t-1}]
\leq 1$.
Since $E_{j,a,t-1}^{\pm}(\theta;\lambda)$ is $\mathscr{F}_{t-1}$-measurable, this will imply that $\mathbb{E}[E_{j,a,t}^{\pm}(\theta;\lambda)\mid\mathscr{F}_{t-1}]
\leq E_{j,a,t-1}^{\pm}(\theta;\lambda)$.
Moreover, $E_{j,a,t}^{\pm}(\theta;\lambda)$ is non-negative by
definition. Conditional on $A_t=a$, we have $I_{t,a}=1$, and hence
\begin{align*}
    \mathbb{E}\left[
        R_{j,a,t}^{\pm}(\theta;\lambda)
        \mid \mathscr{F}_{t-1},A_t=a
    \right]
    &=
    \mathbb{E}\left[
        \exp\left\{
            \lambda\bigl[\pm(Y_t-\theta)-d_J\bigr]
            -
            \psi(\lambda)(Y_t-\widehat Y_{t,a})^2
        \right\}
        \,\middle|\,
        \mathscr{F}_{t-1},A_t=a
    \right]
    \\
    &=
    \exp\left\{
        \lambda\bigl[\pm(\mu_{t,a}-\theta)-d_J\bigr]
    \right\}
    \mathbb{E}\left[
        \exp\left\{
            \pm\lambda(Y_t-\mu_{t,a})
            -
            \psi(\lambda)(Y_t-\widehat Y_{t,a})^2
        \right\}
        \,\middle|\,
        \mathscr{F}_{t-1},A_t=a
    \right]
    \\
    &\leq
    \exp\left\{
        \lambda\bigl[\pm(\mu_{t,a}-\theta)-d_J\bigr]
    \right\}
    \\
    &\leq 1,
\end{align*}
where the first inequality follows from the conditional empirical-Bernstein
inequality of \citet[Theorem~4]{HRMS21}. For the final inequality, under the
null hypothesis $\theta=\mu_{b_j,a}$, H\"older continuity and the fact that
$t\in\mathcal{B}_j$ imply
$\lvert\mu_{t,a}-\theta\rvert
=\lvert\mu_{t,a}-\mu_{b_j,a}\rvert
\leq d_J$. Hence
$\pm(\mu_{t,a}-\theta)\leq
\lvert\mu_{t,a}-\theta\rvert\leq d_J$.

If $A_t\neq a$, then $I_{t,a}=0$, so the increments of both the linear
term and $V_{j,a}(t)$ are zero, and therefore
$R_{j,a,t}^{\pm}(\theta;\lambda)=1$. Writing
$p_t(a):=\mathbb{P}(A_t=a\mid\mathscr{F}_{t-1})$, the tower property
therefore gives
\begin{align*}
    \mathbb{E}\left[
        R_{j,a,t}^{\pm}(\theta;\lambda)
        \mid\mathscr{F}_{t-1}
    \right]
    &=
    p_t(a)
    \mathbb{E}\left[
        R_{j,a,t}^{\pm}(\theta;\lambda)
        \mid\mathscr{F}_{t-1},A_t=a
    \right]
    +(1-p_t(a))
    \\
    &\leq
    p_t(a)+(1-p_t(a))
    =1.
\end{align*}
Consequently,
\[
    \mathbb{E}\left[
        E_{j,a,t}^{\pm}(\theta;\lambda)
        \mid\mathscr{F}_{t-1}
    \right]
    =
    E_{j,a,t-1}^{\pm}(\theta;\lambda)
    \mathbb{E}\left[
        R_{j,a,t}^{\pm}(\theta;\lambda)
        \mid\mathscr{F}_{t-1}
    \right]
    \leq
    E_{j,a,t-1}^{\pm}(\theta;\lambda).
\]
Since $E_{j,a,b_j-1}^{\pm}(\theta;\lambda)=1$, non-negativity and the
preceding inequality also imply
$\mathbb{E}[E_{j,a,t}^{\pm}(\theta;\lambda)]\leq 1$, so the process is
integrable. Thus
$\{E_{j,a,t}^{\pm}(\theta;\lambda)\}_{t\geq b_j-1}$ is a non-negative
$\{\mathscr{F}_t\}$-supermartingale starting at one.

An identical argument using the conditional Hoeffding inequality with
$\lambda=w/4$, for which $\lambda^2/8=w^2/128$, shows that
$\{H_{j,a,t}^{\pm}(\theta)\}_{t\geq b_j-1}$ is also a non-negative
$\{\mathscr{F}_t\}$-supermartingale starting at one. This proves the claim.
\end{proof}

We now show that the stochastic processes constructed above yield the base confidence intervals $I_{j,a,t}$ in~(\ref{eq:adaptive-reference-ci}).

\begin{lem}
\label{appendix-lem:base-confidene-interval}
    Let $\lambda_k=2^{-k-1}$ for $k\geq 0$, and define the level-$\alpha$ sub-exponential uniform boundary
    \begin{equation}
        B_\alpha(v)
        :=
        \inf\left\{
            x\geq 0:
            \sum_{k=0}^{\infty}
            \frac{
                \exp\left\{
                    \lambda_k x-\psi(\lambda_k)v
                \right\}
            }{
                (k+1)(k+2)
            }
            \geq
            \frac{1}{\alpha}
        \right\}.
    \end{equation}
    Under the null hypothesis $\theta=\mu_{b_j,a}$, with probability at least $1-4\alpha$, simultaneously for every $t\in\mathcal B_j$ such that $N_{j,a}(t)>0$,
    \begin{equation}
        \left|
            \overline{Y}_{j,a}(t)-\theta
        \right|
        \leq
        d_J
        +
        \left[
            \frac{
                B_\alpha(V_{j,a}(t))
            }{
                N_{j,a}(t)
            }
            \wedge
            \left(
                \frac{w}{32}
                +
                \frac{
                    4\log(1/\alpha)
                }{
                    wN_{j,a}(t)
                }
            \right)
        \right].
    \end{equation}
\end{lem}

\begin{proof}[Proof of Lemma~\ref{appendix-lem:base-confidene-interval}]
First, define $q_k:=((k+1)(k+2))^{-1}$ and note that $\sum_{k\geq0}q_k=1$. Additionally, define $M_{j,a,t}^{\pm}(\theta):=\sum_{k\geq0}q_kE_{j,a,t}^{\pm}(\theta;\lambda_k)$. By Lemma~\ref{appendix-lem:supermartingale-result}, each $E_{j,a,t}^{\pm}(\theta;\lambda_k)$ is a non-negative supermartingale under the null hypothesis $\theta=\mu_{b_j,a}$. Since each summand is non-negative, conditional Tonelli's theorem gives
\begin{align*}
    \mathbb{E}\left[M_{j,a,t}^{\pm}(\theta)\mid\mathscr{F}_{t-1}\right]
    &=
    \sum_{k\geq0}q_k
    \mathbb{E}\left[E_{j,a,t}^{\pm}(\theta;\lambda_k)\mid\mathscr{F}_{t-1}\right] \\
    &\leq
    \sum_{k\geq0}q_kE_{j,a,t-1}^{\pm}(\theta;\lambda_k)
    =
    M_{j,a,t-1}^{\pm}(\theta).
\end{align*}
Moreover, $M_{j,a,b_j-1}^{\pm}(\theta)=\sum_{k\geq0}q_k=1$, so $M_{j,a,t}^{\pm}(\theta)$ is also a non-negative supermartingale starting at one.

Now consider $B_\alpha(v)$. For any fixed $k\geq0$, let $x:=\lambda_k^{-1}\left(\log(1/(\alpha q_k))+\psi(\lambda_k)v\right)$. Then $q_k\exp\{\lambda_kx-\psi(\lambda_k)v\}=1/\alpha$, and hence, by the definition of $B_\alpha(v)$,
\begin{equation}
    B_\alpha(v)
    \leq
    \inf_{k\geq0}
    \frac{\log((k+1)(k+2)/\alpha)+\psi(2^{-k-1})v}{2^{-k-1}}.
\end{equation}
We now use the mixture process to obtain a time-uniform deviation bound. Since $M_{j,a,t}^{\pm}(\theta)$ is a non-negative supermartingale starting at one, Ville's inequality \citep[Exercise~4.8.2]{durrett2019probability} gives $\mathbb{P}(\sup_{t\in\mathcal B_j}M_{j,a,t}^{\pm}(\theta)\geq1/\alpha)\leq\alpha$. Union bounding over the two signs shows that, with probability at least $1-2\alpha$, neither $M_{j,a,t}^{+}(\theta)$ nor $M_{j,a,t}^{-}(\theta)$ crosses $1/\alpha$ at any $t\in\mathcal B_j$.

For any $t$ such that $N_{j,a}(t)>0$, if $\overline{Y}_{j,a}(t)-\theta>d_J+B_\alpha(V_{j,a}(t))/N_{j,a}(t)$, then $N_{j,a}(t)\left(\overline{Y}_{j,a}(t)-\theta-d_J\right)
    >
    B_\alpha(V_{j,a}(t))$.
By the definition of $B_\alpha$, this implies $M_{j,a,t}^{+}(\theta)\geq1/\alpha$. Similarly, if $\theta-\overline{Y}_{j,a}(t)>d_J+B_\alpha(V_{j,a}(t))/N_{j,a}(t)$, then $M_{j,a,t}^{-}(\theta)\geq1/\alpha$. Therefore, under the null hypothesis $\theta=\mu_{b_j,a}$,
\[
    \mathbb{P} \left( \forall t \in \mathcal{B}_{j}, N_{j, a}(t) > 0, \left|\overline{Y}_{j,a}(t)-\theta\right| 
    \leq
    d_J+\frac{B_\alpha(V_{j,a}(t))}{N_{j,a}(t)} \right) \geq 1 - 2\alpha.
\]

Likewise, by Lemma~\ref{appendix-lem:supermartingale-result}, $H_{j,a,t}^{\pm}(\theta)$ is a non-negative supermartingale starting at one. Applying Ville's inequality and union bounding over the two signs shows that, with probability at least $1-2\alpha$, neither $H_{j,a,t}^{+}(\theta)$ nor $H_{j,a,t}^{-}(\theta)$ crosses $1/\alpha$ at any $t\in\mathcal B_j$. If $\overline{Y}_{j,a}(t)-\theta>d_J+w/32+4\log(1/\alpha)/(wN_{j,a}(t))$, then substituting this inequality into the definition of $H_{j,a,t}^{+}(\theta)$ gives $H_{j,a,t}^{+}(\theta)\geq1/\alpha$. The analogous lower-tail deviation implies $H_{j,a,t}^{-}(\theta)\geq1/\alpha$. Hence, we find:
\[
    \mathbb{P} \left( \forall t \in \mathcal{B}_{j}, N_{j, a}(t) > 0, \left|\overline{Y}_{j,a}(t)-\theta\right|
    \leq
    d_J+\frac{w}{32}+\frac{4\log(1/\alpha)}{wN_{j,a}(t)} \right) \geq 1-2\alpha.
\]
Let 
\begin{align*}
    &\mathscr{B} = \left\{ t \in \mathcal{B}_{j}: N_{j, a}(t) > 0, \left|\overline{Y}_{j,a}(t)-\theta\right|  \leq d_J+\frac{B_\alpha(V_{j,a}(t))}{N_{j,a}(t)} \right\}, \\
    &\mathscr{H} = \left\{ t \in \mathcal{B}_{j} : N_{j, a}(t) > 0, \left|\overline{Y}_{j,a}(t)-\theta\right| \leq 
    d_J+\frac{w}{32}+\frac{4\log(1/\alpha)}{wN_{j,a}(t)}\right\}.
\end{align*}
Then on the event $\mathscr{B} \cap \mathscr{H}$, we can finally conclude that:
\[
    \mathbb{P} \left( \forall t \in \mathcal{B}_{j}, N_{j, a}(t) > 0, \left|\overline{Y}_{j,a}(t)-\theta\right|
    \leq
    d_J
    +
    \left[
        \frac{B_\alpha(V_{j,a}(t))}{N_{j,a}(t)}
        \wedge
        \left(
            \frac{w}{32}
            +
            \frac{4\log(1/\alpha)}{wN_{j,a}(t)}
        \right)
    \right] \right) \geq 1-4\alpha
\]
simultaneously over all such $t$, which proves the claim.
\end{proof}

We can now prove Lemma~\ref{lem:eprocess-validity}.

\begin{proof}[Proof of Lemma~\ref{lem:eprocess-validity}]
Set $\alpha:=\delta/(4KJ)$. Fix any block $j\in[J]$ and arm $a\in\cA$. By Lemma~\ref{appendix-lem:base-confidene-interval}, under the null hypothesis $\theta=\mu_{b_j,a}$,
\[
    \mathbb{P} \left( \forall t \in \mathcal{B}_{j}, N_{j,a}(t) > 0, \left|
        \overline{Y}_{j,a}(t)-\mu_{b_j,a}
    \right|
    \leq
    d_J
    +
    \left[
        \frac{B_\alpha(V_{j,a}(t))}{N_{j,a}(t)}
        \wedge
        \left(
            \frac{w}{32}
            +
            \frac{4\log(1/\alpha)}{wN_{j,a}(t)}
        \right)
    \right] \right) \geq 1 - 4\alpha.
\]
By the definition of the base interval $I_{j,a,t}$ in~(\ref{eq:adaptive-reference-ci}), the preceding inequality is equivalent to $\mu_{b_j,a}\in I_{j,a,t}$ for every $t\in\mathcal B_j$ such that $N_{j,a}(t)>0$. Thus, if we define
$\mathcal E_{j,a}
    :=
    \left\{
        \mu_{b_j,a}\in I_{j,a,t}
        \text{ for every }t\in\mathcal B_j
        \text{ with }N_{j,a}(t)>0
    \right\}$,
then $\mathbb P(\mathcal E_{j,a}^{c})\leq4\alpha$. There are $J$ blocks and two arms $a\in\cA=\{0,1\}$. Therefore, a union bound over $(j,a)$ gives
\[
    \mathbb P\left(
        \bigcap_{j\in[J]}
        \bigcap_{a\in\cA}
        \mathcal E_{j,a}
    \right)
    \geq
    1-4KJ\alpha
    =
    1-\delta.
\]
Fix this event. Recall that $D_{j,a,t}$ is the running intersection of the base intervals $I_{j,a,s}$ obtained up to time $t$ within block $j$, with $D_{j,a,t}=[0,1]$ before the first observation of arm $a$. If $N_{j,a}(t)=0$, then $\mu_{b_j,a}\in[0,1]=D_{j,a,t}$. If $N_{j,a}(t)>0$, then every base interval entering the running intersection contains $\mu_{b_j,a}$, and hence
\[
    \mu_{b_j,a}
    \in
    \bigcap_{\substack{s\in\mathcal B_j:\,s\leq t\\N_{j,a}(s)>0}}
    I_{j,a,s}
    =
    D_{j,a,t}.
\]
In particular, the running intersection cannot be empty on this event, so the convention used to define $D_{j,a,t}$ after an empty intersection is never invoked.

We have therefore shown that, with probability at least $1-\delta$, simultaneously for every $j\in[J]$, $a\in\cA$, and $t\in\mathcal B_j$, $\mu_{b_j,a}\in D_{j,a,t}$. The argument requires only that $\widehat Y_{t,a}\in[0,1]$ be $\mathscr{F}_{t-1}$-measurable, and Lemma~\ref{appendix-lem:supermartingale-result} already establishes the required supermartingale property under adaptive action selection. Hence the conclusion holds for arbitrary predictable predictions and adaptive action selection.
\end{proof}

\section{Proof of Theorem~\ref{thm:adaptive-trace}}
\label{appendix-sec:proof-of-adaptive-trace}
\begin{proof}[Proof of Theorem~\ref{thm:adaptive-trace}]
By Lemma~\ref{lem:eprocess-validity}, with probability at least $1-\delta$, simultaneously for every $j\in[J]$, $a\in\cA$, and $t\in\mathcal B_j$, we have $\mu_{b_j,a}\in D_{j,a,t}$. Hence $\mu_{b_j,a}\in D_{j,a}^{\star}$ for every saved interval. Iterating $D_{j-1,a}^{\star}$ from $b_{j-1}$ to $t\in\mathcal B_j$ using H\"older continuity therefore gives $\tau_t\in \mathscr{C}_{t}$ simultaneously for all $t$. Moreover, since $\lvert D_{j,a}^{\star}\rvert\leq w/4$, $t-b_{j-1}\leq4T/J$, and $L(4/J)^\beta\leq w/32$, $\lvert \mathscr{C}_{t}\rvert \leq w$. Thus the procedure is $(w,\delta)$-valid.

For the regret bound, let $\mathcal T_{\mathrm{exp}}$ denote the random set of forced exploration rounds. The decomposition and the bounds on terms (ii) and (iii) in the proof of Theorem~\ref{thm:upper-bound} apply unchanged, since membership in $\mathcal T_{\mathrm{exp}}$ is determined before observing $Y_t$ and the \textsc{EXP3} weights are frozen on these rounds. Since there are $K$ arms and rewards are bounded in $[0,1]$, the instantaneous regret on any forced exploration round is at most one. Therefore,
\[
    \mathscr{R}_T(\pi,\mu)
    \leq
    \sqrt{2K\log K}\left(\frac{T}{\sqrt h}+\sqrt h\right)
    +
    2LT\left(\frac{h}{T}\right)^\beta
    +
    \E_{\mu}^{\pi}\left[\sum_{t\in\mathcal T_{\mathrm{exp}}}
    \left(\max_{a\in\mathcal A}\mu_{t,a}-\mu_{t,A_t}\right)\right].
\]

For the worst-case bound, define $m_{\star}:=\left\lceil128w^{-2}\log(4KJ/\delta)\right\rceil$. If $N_{j,a}(t)=m_{\star}$, then the base confidence radius is at most $d_J+w/32+4\log(4KJ/\delta)/(wm_{\star})\leq3w/32$, so its width is strictly smaller than $w/4$. Thus each arm is explored at most $m_{\star}$ times per block, and hence $\lvert\mathcal T_{\mathrm{exp}}\rvert\leq KJm_{\star}$. Since the instantaneous regret is at most one, we have $\E_{\mu}^{\pi}[\sum_{t\in\mathcal T_{\mathrm{exp}}}(\max_{a\in\mathcal A}\mu_{t,a}-\mu_{t,A_t})]\leq KJm_{\star}$. Using $J\asymp_{\beta}1+(L/w)^{1/\beta}$ and the same choice of $h$ as in Theorem~\ref{thm:upper-bound} gives
\[
    \mathscr{R}_T(\pi,\mu)
    \lesssim_{\beta}
    \sqrt{KT\log K}
    +
    (K\log K)^{\frac{\beta}{2\beta+1}}
    L^{\frac{1}{2\beta+1}}
    T^{\frac{\beta+1}{2\beta+1}}
    +
    \frac{K[1+(L/w)^{1/\beta}]}{w^2}
    \log\left(\frac{K[1+(L/w)^{1/\beta}]}{\delta}\right).
\]
This proves the result.
\end{proof}

\section{Proof of Corollary~\ref{cor:prediction-error}}
\label{appendix-sec:proof-of-trace-prediction-error}

\begin{proof}[Proof of Corollary~\ref{cor:prediction-error}]
Fix a block $j$ and arm $a$, and write $n:=N_{j,a}(t)$. Since $d_J\leq w/32$, if $B_\alpha(V_{j,a}(t))\leq wn/16$, then the empirical-Bernstein e-process of the confidence radius is at most $3w/32$. Hence the corresponding base interval has width at most $3w/16<w/4$, and therefore the running intersection $D_{j,a,t}$ also has width less than $w/4$. Thus exploration stops, proving the final claim.

Now suppose $V_{j,a}(t)\leq vn+1$ at every attained sample count before stopping. Since $Y_t,\widehat Y_{t,a}\in[0,1]$, we also have $V_{j,a}(t)\leq n\leq m$. Let
$A:=\log(4KJ/\delta)+\log\log(e+m)$. By the geometric-mixture empirical-Bernstein boundary of \citet[Section~3.3]{HRMS21},
\[
    \frac{B_\alpha(V_{j,a}(t))}{n}
    \lesssim
    \sqrt{\frac{vA}{n}}
    +
    \frac{A}{n}.
\]

Therefore, for a sufficiently large universal constant $C$, any attained sample count satisfying $n\geq C(v/w^2+1/w)A$ also satisfies $B_\alpha(V_{j,a}(t))/n\leq w/16$, and hence exploration must stop by that count. On the other hand, the deterministic Hoeffding concentration guarantees stopping after at most $m$ observations. Combining the two bounds gives
\[
    n_{j,a}
    \lesssim
    m
    \wedge
    \left(
        \frac{v}{w^2}+\frac{1}{w}
    \right)
    \left[
        \log\left(\frac{4KJ}{\delta}\right)
        +
        \log\log(e+m)
    \right].
\]
This proves the result.
\end{proof}

\section{Proof of Theorem~\ref{thm:local-normality}}
\label{appendix-sec:proof-of-local-normality}
Before proving Theorem~\ref{thm:local-normality}, we will establish some notation we will subsequently use throughout this section. First define $\mathcal{W}_{T} = \left\{ b_{j_{T} - 1}, b_{j_{T} - 1}, \dots, b_{j_{T}} \right\} := \left\{ b_{j_{T} - 1}, b_{j_{T} - 1}, \dots, t_{T} - 1 \right\} $. Let $s_{r} = b_{j_{T}-1} + r - 1$ denote rounds in $\mathcal{W}_{T}$ where $r \in [H_{T}]$ where $H_{T} = t_{T} - b_{j_{T}-1}$. We also define $\xi_{s} := Y_{s} - \mu_{s, A_{s}}$ for $a \in \mathcal{A}$ and define $S_{T, a} := \sum_{s \in \mathcal{W}_{T}} \mathbf{1}\left\{ A_s = a \right\}\xi_{s}$.

Additionally, we will define the following filtrations which we will find useful in our analysis. Let $\mathscr{E}_T$ denote the sigma-field generated by an exogenous process applied to the mean trajectory if it exists. Then we define the following filtrations:
\begin{align*}
    \mathscr{G}_{T, 0} &= \mathscr{F}_{b_{j_{T}-1} -1} \vee \mathscr{E}_{T} \\
    \mathscr{G}_{T, 2r-1} &= \mathscr{F}_{s_r - 1} \vee \sigma(A_{s_r}) \vee \mathscr{E}_{T} \\
    \mathscr{G}_{T. 2r} &= \mathscr{F}_{s_{r}} \vee \mathscr{E}_{T}.
\end{align*}
The filtration $\mathscr{G}_{T, 2r-1}$ contains the action $A_{s_{r}}$ being revealed but not the reward while $\mathscr{G}_{T, 2r}$ additionally contains $Y_{s_r}$. Under the Bernoulli observational model, we have:
\begin{equation*}
    \mathbb{E}[\xi_{s_{r}} \mid \mathscr{G}_{T, 2r-1}] = 0, \; \mathbb{E}[\xi_{s_{r}}^2 \mid \mathscr{G}_{T, 2r-1}] = \mu_{s_{r}, A_{s_r}} (1 - \mu_{s_{r}, A_{s_r}}), \; \lvert \xi_{s_{r}} \rvert \leq 1.
\end{equation*}
We first establish that $S_{T,a}$ is of stochastic order $\sqrt{n_{T,a}}$.

\begin{lem}[Stochastic order of $S_{T,a}$]
\label{lem:stochastic-order-S-Ta}
For each arm $a$, $S_{T,a} = \mathcal{O}_{P}\!\left(\sqrt{n_{T,a}}\right)$.
\end{lem}

\begin{proof}[Proof of Lemma~\ref{lem:stochastic-order-S-Ta}]
    Fix $a \in \mathcal{A}$. We will define the following truncated version of $S_{T, a}$:
    \begin{equation}
        \overline{S}_{T, a} = \sum_{s \in \mathcal{W}_T} \mathbf{1}\left\{ A_{s} = a \right\} \mathbf{1} \left\{ N_{T, a}^{(s-1)} < \lceil 2n_{T, a} \rceil \right\} \xi_{s},
    \end{equation}
    where $N_{T, a}^{(s-1)} = \sum_{s^{\prime} \in \mathcal{W}_{T} : s^{\prime} < s} \mathbf{1}\left\{ A_{s} = a \right\}$. Now notice the following:
    \begin{align}
        \label{eq:truncated-sum-squared}
        \overline{S}_{T, a}^2 = \sum_{s \in \mathcal{W}_T} \mathbf{1}\left\{ A_{s} = a \right\} \mathbf{1} \left\{ N_{T, a}^{(s-1)} < \lceil 2n_{T, a} \rceil \right\} \xi_{s}^{2} + 2\sum_{\substack{s,s'\in\mathcal{W}_T\\ s<s'}} \mathbf{1}\left\{ A_{s} = a \right\} \mathbf{1} \left\{ N_{T, a}^{(s-1)} < \lceil 2n_{T, a} \rceil \right\} \xi_{s}\xi_{s^{\prime}}.
    \end{align}
    Taking expectations of \cref{eq:truncated-sum-squared} yields:
    \begin{align*}
\mathbb{E}\!\left[\overline{S}_{T,a}^{\,2}\right]
&=
\sum_{s\in\mathcal{W}_T}
\mathbb{E}\!\left[
    \mathbf{1}\{A_s=a\}
    \mathbf{1}\!\left\{
        N_{T,a}^{(s-1)}<\lceil 2n_{T,a}\rceil
    \right\}
    \xi_s^2
\right] \\
&\;
+2\sum_{\substack{s,s'\in\mathcal{W}_T\\ s<s'}}
\mathbb{E}\!\left[
    \mathbf{1}\{A_s=a\}
    \mathbf{1}\!\left\{
        N_{T,a}^{(s-1)}<\lceil 2n_{T,a}\rceil
    \right\}
    \mathbf{1}\{A_{s'}=a\}
    \mathbf{1}\!\left\{
        N_{T,a}^{(s'-1)}<\lceil 2n_{T,a}\rceil
    \right\}
    \xi_s\xi_{s'}
\right] \\
&=
\sum_{s\in\mathcal{W}_T}
\mathbb{E}\!\left[
    \mathbf{1}\{A_s=a\}
    \mathbf{1}\!\left\{
        N_{T,a}^{(s-1)}<\lceil 2n_{T,a}\rceil
    \right\}
    \mathbb{E}\!\left[\xi_s^2\mid\mathscr{G}_{s-1}\right]
\right] \\
&\;
+2\sum_{\substack{s,s'\in\mathcal{W}_T\\ s<s'}}
\mathbb{E}\!\left[
    \mathbf{1}\{A_s=a\}
    \mathbf{1}\!\left\{
        N_{T,a}^{(s-1)}<\lceil 2n_{T,a}\rceil
    \right\}
    \mathbf{1}\{A_{s'}=a\}
    \mathbf{1}\!\left\{
        N_{T,a}^{(s'-1)}<\lceil 2n_{T,a}\rceil
    \right\}
    \xi_s
    \mathbb{E}\!\left[\xi_{s'}\mid\mathscr{G}_{s'-1}\right]
\right] \\
&=
\sum_{s\in\mathcal{W}_T}
\mathbb{E}\!\left[
    \mathbf{1}\{A_s=a\}
    \mathbf{1}\!\left\{
        N_{T,a}^{(s-1)}<\lceil 2n_{T,a}\rceil
    \right\}
\right]
\mu_{s,a}(1-\mu_{s,a}) \\
&\le
\frac14
\sum_{s\in\mathcal{W}_T}
\mathbb{E}\!\left[
    \mathbf{1}\{A_s=a\}
    \mathbf{1}\!\left\{
        N_{T,a}^{(s-1)}<\lceil 2n_{T,a}\rceil
    \right\}
\right] \\
&\le
\frac{\lceil 2n_{T,a}\rceil}{4},
\end{align*}
where the first inequality holds from $x(1-x) \leq 1/4 $ for $x \in [0, 1]$. Then for $M > 0$, Chebyshev's inequality yields $\mathbb{P} \left( \lvert \overline{S}_{T, a} \rvert > M \sqrt{n_{T, a}} \right) \leq c_0 / M^2$
for some universal constant $c_0 > 0$. Thus, for any $\epsilon > 0$, choosing $M = c_0^{\prime}\epsilon^{-1/2}$ sufficiently large, we find $\overline{S}_{T, a} = \mathcal{O}_{P}(\sqrt{n_{T, a}})$.

Now note that we can write $S_{T, a}$ can be written as follows:
\begin{equation}
    S_{T, a} = \underbrace{\sum_{s \in \mathcal{W}_T} \mathbf{1}\left\{ A_{s} = a \right\} \mathbf{1} \left\{ N_{T, a}^{(s-1)} \geq \lceil 2n_{T, a} \rceil \right\} \xi_{s}}_{\overline{S}^{\prime}_{T, a}} + \underbrace{\sum_{s \in \mathcal{W}_T} \mathbf{1}\left\{ A_{s} = a \right\} \mathbf{1} \left\{ N_{T, a}^{(s-1)} < \lceil 2n_{T, a} \rceil \right\} \xi_{s}}_{\overline{S}_{T, a}}.
\end{equation}
Since $\left\{ \lvert S_{T, a} \rvert > M \sqrt{n_{T, a}} \right\} \subseteq \left\{ \lvert \overline{S}^{\prime}_{T, a} \rvert > M \sqrt{n_{T, a}} / 2 \right\} \cup \left\{ \lvert \overline{S}_{T, a} \rvert > M \sqrt{n_{T, a}} / 2 \right\}$ and we have shown $\overline{S}_{T, a} = \mathcal{O}_{P}(\sqrt{n_{T, a}})$, it suffices to prove that $\mathbb{P}\left( \lvert \overline{S}^{\prime}_{T, a} \rvert > M \sqrt{n_{T, a}} / 2 \right) \rightarrow 0$ for every fixed $M > 0$. 

Clearly, $\left\{ \lvert \overline{S}^{\prime}_{T, a} \rvert > M \sqrt{n_{T, a}} / 2 \right\} \subseteq \left\{ \overline{S}^{\prime}_{T, a} \neq 0 \right\}$. Moreover, $\overline{S}^{\prime}_{T, a} \neq 0$ can occur only if there exists some $s \in \mathcal{W}_{T}$ such that $N_{T, a}^{(s-1)} \geq \lceil 2n_{T, a} \rceil$. Since $N_{T, a}^{(s-1)} \leq N_{T, a}$ for every $s$, we have $\left\{ \overline{S}^{\prime}_{T, a} \neq 0 \right\} \subseteq \left\{ N_{T, a} \geq \lceil 2n_{T, a} \rceil \right\}$. By Assumption~\ref{ass:local-inference}(i), $\mathbb{P}\left( N_{T, a} \geq \lceil 2n_{T, a} \rceil \right) \rightarrow 0$. Therefore, $\mathbb{P}\left( \lvert \overline{S}^{\prime}_{T, a} \rvert > M \sqrt{n_{T, a}} / 2 \right) \rightarrow 0$, which proves the claim.
\end{proof}

We further note that Assumption~\ref{ass:local-inference}(i) implies $\mathbb{P}(N_{T, a} = 0) \rightarrow 0$. Indeed if $N_{T,a}=0$, then $\left\{ N_{T, a} = 0 \right\} \subseteq \left\{ \lvert N_{T, a} / n_{T, a} - 1 \rvert > 1/2 \right\}$. Since Assumption~\ref{ass:local-inference}(i) implies $\mathbb{P}(\lvert N_{T, a} / n_{T, a} - 1 \rvert > 1/2) \rightarrow 0$, we can assume $N_{T, a} > 0$. We now will prove the first part of Theorem~\ref{thm:local-normality}.
\begin{lem}
    \label{appendix-lem:first-part}
    Suppose Assumption~\ref{ass:local-inference}(i) and Assumption~\ref{ass:local-inference}(ii)  hold. Then, for every $a\in\cA_+$,
\begin{equation}
\begin{aligned}
    \widetilde\tau_T(a)-\tau_{t_T}(a)
    ={}&
    \sum_{s\in\mathcal W_T}
    \left(
        \frac{\mathbf 1\{A_s=a\}}{n_{T,a}}
        -
        \frac{\mathbf 1\{A_s=0\}}{n_{T,0}}
    \right)
    (Y_s-\mu_{s,A_s}) +
    o_{P}\left(
        \sqrt{
            \frac{1}{n_{T,a}}
            +
            \frac{1}{n_{T,0}}
        }
    \right),
    \label{eq:local-linearization}
\end{aligned}
\end{equation}
where $o_{P}$ is taken under baseline law $\mathsf P_{T,0}$.
\end{lem}

\begin{proof}[Proof of Lemma~\ref{appendix-lem:first-part}]
    By definition, we have 
\begin{align*}
    \widetilde{\mu}_{T, a} - \mu_{t_{T}, a}
    &= \frac{1}{N_{T, a}} \sum_{s \in \mathcal{W}_{T}} \mathbf{1} \left\{ A_{s} = a \right\} (Y_{s} - \mu_{t_{T}, a}) \\
    &= \frac{1}{N_{T, a}} \sum_{s \in \mathcal{W}_{T}} \mathbf{1} \left\{ A_{s} = a \right\} (Y_{s} - \mu_{s, a}) 
    + \frac{1}{N_{T, a}} \sum_{s \in \mathcal{W}_{T}} \mathbf{1} \left\{ A_{s} = a \right\} (\mu_{s, a} - \mu_{t_{T}, a}) \\
    &= \frac{S_{T, a}}{N_{T, a}} 
    + \frac{1}{N_{T, a}} \sum_{s \in \mathcal{W}_{T}} \mathbf{1} \left\{ A_{s} = a \right\} (\mu_{s, a} - \mu_{t_{T}, a}) \\
    &= \frac{S_{T, a}}{n_{T, a}} 
    + \left( \frac{1}{N_{T, a}} - \frac{1}{n_{T, a}}\right) S_{T, a} 
    + \frac{1}{N_{T, a}} \sum_{s \in \mathcal{W}_{T}} \mathbf{1} \left\{ A_{s} = a \right\} (\mu_{s, a} - \mu_{t_{T}, a}).
\end{align*}
Note that $n_{T, a}/N_{T, a} - 1 = o_{P}(1)$ by Assumption~\ref{ass:local-inference}(i) and $S_{T, a} / \sqrt{n_{T, a}} = \mathcal{O}_{P}(1)$ by Lemma~\ref{lem:stochastic-order-S-Ta}. Therefore,
\[
\left( \frac{1}{N_{T, a}} - \frac{1}{n_{T, a}} \right) S_{T,a}
=
\left( \frac{n_{T,a}}{N_{T,a}} - 1 \right)
\frac{S_{T,a}}{\sqrt{n_{T,a}}}
\frac{1}{\sqrt{n_{T,a}}}
=
o_{P}(n_{T,a}^{-1/2}).
\]

We will now show that the last term is also of order $o_{P}(n_{T,a}^{-1/2})$. Let $\mathcal{D}_{T} := \left\{ \max_{s \in \mathcal{W}_{T}} \lvert \mu_{s, a} - \mu_{t_{T}, a} \rvert \leq b_{T, a} \; \forall a \in \mathcal{A} \right\}$. On $\mathcal{D}_{T}$ and whenever $N_{T,a}>0$, we have
\begin{align*}
    \left\lvert 
    \frac{1}{N_{T, a}} 
    \sum_{s \in \mathcal{W}_{T}}
    \mathbf{1} \left\{ A_{s} = a \right\} 
    (\mu_{s, a} - \mu_{t_{T}, a})
    \right\rvert
    \leq 
    \frac{1}{N_{T,a}}
    \sum_{s \in \mathcal{W}_{T}}
    \mathbf{1}\left\{A_s=a\right\}
    \left\lvert \mu_{s,a}-\mu_{t_T,a}\right\rvert 
    \leq b_{T,a}.
\end{align*}
It follows that
\begin{align*}
    &\left\{ 
    \sqrt{n_{T, a}} 
    \left\lvert 
    \frac{1}{N_{T, a}} 
    \sum_{s \in \mathcal{W}_{T}}
    \mathbf{1} \left\{ A_{s} = a \right\} 
    (\mu_{s, a} - \mu_{t_{T}, a})
    \right\rvert  
    > \epsilon, 
    N_{T, a} > 0 
    \right\} \subseteq
    \left\{ \sqrt{n_{T,a}} b_{T,a} > \epsilon \right\}
    \cup \mathcal{D}_{T}^{c}.
\end{align*}
Therefore,
\begin{align*}
    &\mathbb{P} \left( 
    \sqrt{n_{T, a}} 
    \left\lvert 
    \frac{1}{N_{T, a}} 
    \sum_{s \in \mathcal{W}_{T}}
    \mathbf{1} \left\{ A_{s} = a \right\} 
    (\mu_{s, a} - \mu_{t_{T}, a})
    \right\rvert  
    > \epsilon, 
    N_{T, a} > 0 
    \right) \leq
    \mathbb{P}(\mathcal{D}_{T}^{c})
    +
    \mathbb{P}\left(
    \sqrt{n_{T,a}} b_{T,a} > \epsilon
    \right)
    \rightarrow 0,
\end{align*}
where the convergence follows from Assumption~\ref{ass:local-inference}(ii). Moreover, Assumption~\ref{ass:local-inference}(i) implies $\mathbb{P}(N_{T,a}=0)\rightarrow 0$. Hence,
\[
\frac{1}{N_{T, a}} 
\sum_{s \in \mathcal{W}_{T}}
\mathbf{1} \left\{ A_{s} = a \right\} 
(\mu_{s, a} - \mu_{t_{T}, a})
=
o_{P}(n_{T,a}^{-1/2}).
\]
We have therefore established
$\widetilde{\mu}_{T, a} - \mu_{t_{T}, a}
=
S_{T, a} / n_{T, a}
+
o_{P}(n_{T,a}^{-1/2})$. 
The desired expression for $\widetilde{\tau}_{T}$ now follows by substituting this expression into the definitions of $\widetilde{\tau}_{T}$ and $\tau_{t_{T}}$.
\end{proof}
To prove the second claim, we first show $\sum_{s \in \mathcal{W}_{T}} \mathbf{1} \left\{ A_{s} = a \right\} \mu_{s, a}(1-\mu_{s,a}) / n_{T, a} \xrightarrow{\mathsf P_{T,0}} \sigma_{a}^{2} := p_{a}(1-p_{a})$.
\begin{lem}[Convergence of the conditional variance]
\label{lem:conditional-variance-convergence}
For each arm $a \in \mathcal{A}$,
\[
\frac{1}{n_{T,a}}
\sum_{s \in \mathcal{W}_{T}}
\mathbf{1}\left\{A_s=a\right\}
\mu_{s,a}(1-\mu_{s,a})
\xrightarrow{\mathsf P_{T,0}}
\sigma_a^2
:=
p_a(1-p_a).
\]
\end{lem}

\begin{proof}[Proof of Lemma~\ref{lem:conditional-variance-convergence}]
Consider $f(x) := x(1-x)$. Since $|f'(x)| = |1-2x| \leq 1$ on $[0,1]$, $f$ is $1$-Lipschitz. Recall $\mathcal{D}_T := \left\{ \max_{s \in \mathcal{W}_T} |\mu_{s,a}-\mu_{t_T,a}| \leq b_{T,a} \;\; \forall a \in \mathcal{A} \right\}$. On $\mathcal{D}_T$, for every $s \in \mathcal{W}_T$,
$|\mu_{s,a}-p_a| \leq b_{T,a}+|\mu_{t_T,a}-p_a|$. Therefore, using $f(p_a)=\sigma_a^2$ and $\sum_{s \in \mathcal{W}_T}\mathbf{1}\{A_s=a\}=N_{T,a}$,
\begin{align*}
    \left|
    \frac{1}{n_{T,a}}
    \sum_{s \in \mathcal{W}_T}
    \mathbf{1}\{A_s=a\}f(\mu_{s,a})
    -f(p_a)
    \right| 
    &\leq
    \frac{1}{n_{T,a}}
    \sum_{s \in \mathcal{W}_T}
    \mathbf{1}\{A_s=a\}
    |f(\mu_{s,a})-f(p_a)|
    +
    \sigma_a^2
    \left|
    \frac{N_{T,a}}{n_{T,a}}-1
    \right| \\
    &\leq
    \frac{N_{T,a}}{n_{T,a}}
    \left(
    b_{T,a}+|\mu_{t_T,a}-p_a|
    \right)
    +
    \sigma_a^2
    \left|
    \frac{N_{T,a}}{n_{T,a}}-1
    \right|.
\end{align*}
By Assumption~\ref{ass:local-inference}(i), $N_{T,a}/n_{T,a} \xrightarrow{\mathsf P_{T,0}} 1$ and $\mu_{t_T,a}\xrightarrow{\mathsf P_{T,0}}p_a$. Moreover, $\sqrt{n_{T,a}}b_{T,a}\to0$ and $n_{T,a}\to\infty$ imply $b_{T,a}\to0$. Hence, the right-hand side converges to zero in probability.

Finally, for every $\epsilon>0$,
\[
\mathbb{P}\left(
\left|
\frac{1}{n_{T,a}}
\sum_{s \in \mathcal{W}_T}
\mathbf{1}\{A_s=a\}\mu_{s,a}(1-\mu_{s,a})
-\sigma_a^2
\right|>\epsilon
\right)
\leq
\mathbb{P}(\mathcal{D}_T^c)
+
\mathbb{P}(R_T>\epsilon),
\]
where $R_T$ denotes the preceding upper bound. Since $\mathbb{P}(\mathcal{D}_T^c)\leq\delta_T\to0$ and $R_T\xrightarrow{\mathsf P_{T,0}}0$, the desired convergence follows.
\end{proof}

We will now prove the asymptotic normality result of Theorem~\ref{thm:local-normality}. Towards this end, fix a nonzero $c \in \mathbb{R}^{K-1}$ and let $\ell_a := c_a$ for $a \in \mathcal{A}_{+}$ and $\ell_0 := -\sum_{a \in \mathcal{A}_{+}} c_a$. By Lemma~\ref{appendix-lem:first-part}, we may write
\[
    \widetilde{\mu}_{T,a} - \mu_{t_T,a}
    =
    \frac{S_{T,a}}{n_{T,a}} + r_{T,a},
    \;\;
    \sqrt{n_{T,a}}\,r_{T,a} \xrightarrow{\mathsf P_{T,0}} 0.
\]
It follows that
\begin{align}
    c^\top\left(
        \widetilde{\boldsymbol\tau}_T
        -
        \boldsymbol\tau_{t_T}
    \right)
    =
    \sum_{a \in \mathcal{A}}
    \frac{\ell_a S_{T,a}}{n_{T,a}}
    +
    \sum_{a \in \mathcal{A}} \ell_a r_{T,a}, \;
    c^\top\Sigma_Tc
    =
    \sum_{a \in \mathcal{A}}
    \frac{\ell_a^2\sigma_a^2}{n_{T,a}}.
\end{align}
Since $c \neq 0$, at least one $\ell_a \neq 0$. Moreover, $\sigma_a^2>0$ for every $a \in \mathcal A$, so $c^\top\Sigma_Tc>0$.

\begin{lem}[Martingale CLT for the leading term]
\label{lem:contrast-martingale-clt}
Suppose Assumption~\ref{ass:local-inference}(i) and Assumption~\ref{ass:local-inference}(ii) hold. Then, for every fixed nonzero $c \in \mathbb{R}^{K-1}$,
\[
    \frac{
        \sum_{a \in \mathcal{A}}
        \ell_a S_{T,a}/n_{T,a}
    }{
        \sqrt{c^\top\Sigma_Tc}
    }
    \Rightarrow
    \mathcal{N}(0,1).
\]
\end{lem}

\begin{proof}
Write $\mathcal{W}_T=\{s_1<\cdots<s_{H_T}\}$. Define the martingale triangular array
\[
    X_{T,2r-1}:=0,
    \;\;
    X_{T,2r}
    :=
    \frac{\xi_{s_r}}{\sqrt{c^\top\Sigma_Tc}}
    \sum_{a \in \mathcal A}
    \frac{\ell_a\mathbf{1}\{A_{s_r}=a\}}{n_{T,a}},
    \;\;
    r=1,\ldots,H_T,
\]
where $\mathscr{G}_{T,2r-1}$ contains the history through the choice of $A_{s_r}$ and $\mathscr{G}_{T,2r}$ additionally contains the reward at round $s_r$. By construction, $X_{T,k}$ is $\mathscr{G}_{T,k}$-measurable and
\[
    \sum_{k=1}^{2H_T}X_{T,k}
    =
    \frac{
        \sum_{a \in \mathcal A}\ell_aS_{T,a}/n_{T,a}
    }{
        \sqrt{c^\top\Sigma_Tc}
    }.
\]

We proceed by verifying the conditions of \citet[Theorem~8.2.4]{durrett2019probability}. We first verify conditional centering. Since the coefficient multiplying $\xi_{s_r}$ is $\mathscr{G}_{T,2r-1}$-measurable and $\mathbb{E}[\xi_{s_r}\mid\mathscr{G}_{T,2r-1}]=0$, we have $\mathbb{E}[X_{T,2r}\mid\mathscr{G}_{T,2r-1}]=0$. The odd increments are identically zero, so $\{X_{T,k},\mathscr{G}_{T,k}\}$ is a martingale difference array.

We next verify square integrability and the conditional Lindeberg condition. Since exactly one arm is selected at each round, if $\ell_{A_{s_r}}\neq0$, then
\[
    c^\top\Sigma_Tc
    =
    \sum_{a \in \mathcal A}
    \frac{\ell_a^2\sigma_a^2}{n_{T,a}}
    \geq
    \frac{
        \ell_{A_{s_r}}^2\sigma_{A_{s_r}}^2
    }{
        n_{T,A_{s_r}}
    }.
\]
Using $|\xi_{s_r}|\leq1$,
\begin{align*}
    |X_{T,2r}|
    &\leq
    \frac{
        |\ell_{A_{s_r}}|
    }{
        n_{T,A_{s_r}}\sqrt{c^\top\Sigma_Tc}
    } \leq
    \frac{1}{
        \sigma_{A_{s_r}}
        \sqrt{n_{T,A_{s_r}}}
    }.
\end{align*}
If $\ell_{A_{s_r}}=0$, then $X_{T,2r}=0$. Hence
\[
    \max_{k\leq 2H_T}|X_{T,k}|
    \leq
    \max_{a\in\mathcal A}
    \frac{1}{\sigma_a\sqrt{n_{T,a}}}
    \longrightarrow 0,
\]
where the convergence is deterministic since $K$ is fixed, $\min_{a\in\mathcal A}\sigma_a>0$, and $\min_{a\in\mathcal A}n_{T,a}\to\infty$. In particular, every increment is square-integrable. Moreover, for every $\epsilon>0$, $|X_{T,k}|\leq\epsilon$ almost surely for all $k$ and all sufficiently large $T$. Therefore,
$\sum_{k=1}^{2H_T}
    \mathbb{E}\!\left[
        X_{T,k}^2
        \mathbf{1}\{|X_{T,k}|>\epsilon\}
        \mid
        \mathscr{G}_{T,k-1}
    \right]
    =
    0
$
for all sufficiently large $T$, so the conditional Lindeberg condition holds.

It remains to verify the conditional variance condition. Since exactly one arm is selected,
\begin{align*}
    \sum_{k=1}^{2H_T}
    \mathbb{E}\!\left[
        X_{T,k}^2
        \mid
        \mathscr{G}_{T,k-1}
    \right] 
    &=
    \sum_{a \in \mathcal A}
    \frac{\ell_a^2}{
        n_{T,a}^2 c^\top\Sigma_Tc
    }
    \sum_{s \in \mathcal W_T}
    \mathbf{1}\{A_s=a\}
    \mu_{s,a}(1-\mu_{s,a}) \\
    &=
    \sum_{a \in \mathcal A}
    w_{T,a}
    \left[
        \frac{
            \sum_{s \in \mathcal W_T}
            \mathbf{1}\{A_s=a\}
            \mu_{s,a}(1-\mu_{s,a})
        }{
            n_{T,a}\sigma_a^2
        }
    \right] \\
    &\xrightarrow{\mathsf P_{T,0}}1,
\end{align*}
where $w_{T,a} = (\ell_a^2\sigma_a^2/n_{T,a}) / (c^\top\Sigma_Tc)$.  The weights $w_{T,a}$ are deterministic, nonnegative, and satisfy $\sum_{a\in\mathcal A} w_{T,a}=1$. The final convergence step follows from Lemma~\ref{lem:conditional-variance-convergence}. Hence, the conditional variance condition is satisfied, and \citet[Theorem~8.2.4]{durrett2019probability} yields the claim.
\end{proof}

\begin{lem}[Consistency of the plug-in covariance]
\label{lem:plugin-covariance-consistency}
Suppose Assumption~\ref{ass:local-inference}(i) and Assumption~\ref{ass:local-inference}(ii) hold. Then, for every fixed nonzero $c \in \mathbb{R}^{K-1}$,
\[
    \frac{c^\top\widehat{\Sigma}_Tc}{c^\top\Sigma_Tc}
    \xrightarrow{\mathsf P_{T,0}}1.
\]
\end{lem}

\begin{proof}
By Lemma~\ref{appendix-lem:first-part},
$\widetilde{\mu}_{T,a}-\mu_{t_T,a}
= \mathcal{O}_P(n_{T,a}^{-1/2}) + o_P(n_{T,a}^{-1/2})
= o_P(1)$, since $n_{T,a}\to\infty$. Together with
$\mu_{t_T,a}\xrightarrow{\mathsf P_{T,0}}p_a$, this gives
$\widetilde{\mu}_{T,a}\xrightarrow{\mathsf P_{T,0}}p_a$. Let $f(x):=x(1-x)$. Since $f$ is continuous, the continuous mapping theorem gives
$f(\widetilde{\mu}_{T,a})\xrightarrow{\mathsf P_{T,0}}f(p_a)=\sigma_a^2$. Since $\sigma_a^2>0$,
\[
    \frac{\widetilde{\mu}_{T,a}(1-\widetilde{\mu}_{T,a})}{\sigma_a^2}
    \xrightarrow{\mathsf P_{T,0}}1.
\]
Moreover, Assumption~\ref{ass:local-inference}(i) gives
$N_{T,a}/n_{T,a}\xrightarrow{\mathsf P_{T,0}}1$. Since $x\mapsto x^{-1}$ is continuous at $1$, another application of the continuous mapping theorem yields
$n_{T,a}/N_{T,a}\xrightarrow{\mathsf P_{T,0}}1$ on the event $N_{T,a}>0$, whose probability tends to one. Therefore, by Slutsky's theorem,
\[
    \frac{
        \widetilde{\mu}_{T,a}(1-\widetilde{\mu}_{T,a})/N_{T,a}
    }{
        \sigma_a^2/n_{T,a}
    }
    =
    \frac{\widetilde{\mu}_{T,a}(1-\widetilde{\mu}_{T,a})}{\sigma_a^2}
    \frac{n_{T,a}}{N_{T,a}}
    \xrightarrow{\mathsf P_{T,0}}1.
\]

By definition,
\[
    c^\top\widehat{\Sigma}_Tc
    =
    \sum_{a \in \mathcal A}
    \ell_a^2
    \frac{\widetilde{\mu}_{T,a}(1-\widetilde{\mu}_{T,a})}{N_{T,a}},
\]
while
$c^\top\Sigma_Tc
=
\sum_{a \in \mathcal A}\ell_a^2\sigma_a^2/n_{T,a}$.
Hence, with
$w_{T,a}:=(\ell_a^2\sigma_a^2/n_{T,a})/(c^\top\Sigma_Tc)$,
\[
    \frac{c^\top\widehat{\Sigma}_Tc}{c^\top\Sigma_Tc}
    =
    \sum_{a \in \mathcal A}
    w_{T,a}
    \frac{
        \widetilde{\mu}_{T,a}(1-\widetilde{\mu}_{T,a})/N_{T,a}
    }{
        \sigma_a^2/n_{T,a}
    }.
\]
The weights $w_{T,a}$ are deterministic, nonnegative, and satisfy
$\sum_{a\in\mathcal A}w_{T,a}=1$. Therefore,
\[
    \left|
        \frac{c^\top\widehat{\Sigma}_Tc}{c^\top\Sigma_Tc}
        -1
    \right|
    \leq
    \max_{a\in\mathcal A}
    \left|
        \frac{
            \widetilde{\mu}_{T,a}(1-\widetilde{\mu}_{T,a})/N_{T,a}
        }{
            \sigma_a^2/n_{T,a}
        }
        -1
    \right| \xrightarrow{\mathsf P_{T,0}}0.
\]
This concludes the proof. 
\end{proof}
We now establish the following result. This combined with Lemma~\ref{appendix-lem:first-part} proves Theorem~\ref{thm:local-normality}. 

\begin{lem}
\label{appendix-lem:second-part}
Suppose Assumption~\ref{ass:local-inference}(i) and Assumption~\ref{ass:local-inference}(ii) hold. Let $\widetilde{\boldsymbol\tau}_T :=(\widetilde\tau_T(a))_{a\in\mathcal A_+}$ and
$\boldsymbol\tau_{t_T} :=(\tau_{t_T}(a))_{a\in\mathcal A_+}$. Then, for every fixed nonzero $c\in\mathbb R^{K-1}$,
\[
    \frac{c^\top\widehat\Sigma_Tc}{c^\top\Sigma_Tc}
    \xrightarrow{\mathsf P_{T,0}}1,
    \;\;
    \frac{
        c^\top(
            \widetilde{\boldsymbol\tau}_T
            -
            \boldsymbol\tau_{t_T}
        )
    }{
        \sqrt{c^\top\widehat\Sigma_Tc}
    }
    \Rightarrow
    \mathcal{N}(0,1).
\]
\end{lem}

\begin{proof}
The first claim follows directly from Lemma~\ref{lem:plugin-covariance-consistency}. For the second claim, recall that
\[
    c^\top\left(
        \widetilde{\boldsymbol\tau}_T
        -
        \boldsymbol\tau_{t_T}
    \right)
    =
    \sum_{a \in \mathcal A}
    \frac{\ell_aS_{T,a}}{n_{T,a}}
    +
    \sum_{a \in \mathcal A}\ell_ar_{T,a}.
\]
We first show that the remainder term is negligible relative to $\sqrt{c^\top\Sigma_Tc}$. For every $a$ such that $\ell_a\neq0$, we have
$c^\top\Sigma_Tc \geq \ell_a^2\sigma_a^2/n_{T,a}$, and therefore
$|\ell_a|n_{T,a}^{-1/2}/\sqrt{c^\top\Sigma_Tc}\leq1/\sigma_a$. Hence,
\begin{align*}
    \left|
        \frac{
            \sum_{a\in\mathcal A}\ell_ar_{T,a}
        }{
            \sqrt{c^\top\Sigma_Tc}
        }
    \right|
    &\leq
    \sum_{a:\ell_a\neq0}
    \frac{
        |\ell_a|n_{T,a}^{-1/2}
    }{
        \sqrt{c^\top\Sigma_Tc}
    }
    \left|
        \sqrt{n_{T,a}}\,r_{T,a}
    \right| \\
    &\leq
    \sum_{a:\ell_a\neq0}
    \frac{1}{\sigma_a}
    \left|
        \sqrt{n_{T,a}}\,r_{T,a}
    \right|
    =
    o_P(1).
\end{align*}
Indeed, $\sqrt{n_{T,a}}\,r_{T,a}\xrightarrow{\mathsf P_{T,0}}0$ for every $a$, and the sum contains only finitely many terms since $K$ is fixed. Combining this with Lemma~\ref{lem:contrast-martingale-clt} and Slutsky's theorem gives
\[
    \frac{
        c^\top(
            \widetilde{\boldsymbol\tau}_T
            -
            \boldsymbol\tau_{t_T}
        )
    }{
        \sqrt{c^\top\Sigma_Tc}
    }
    \Rightarrow
    \mathcal{N}(0,1).
\]

By Lemma~\ref{lem:plugin-covariance-consistency},
$c^\top\widehat\Sigma_Tc/(c^\top\Sigma_Tc)\xrightarrow{\mathsf P_{T,0}}1$. Since $x\mapsto x^{-1/2}$ is continuous at $1$, the continuous mapping theorem gives
\[
    \left(
        \frac{c^\top\Sigma_Tc}
             {c^\top\widehat\Sigma_Tc}
    \right)^{1/2}
    \xrightarrow{\mathsf P_{T,0}}1,
\]
on an event whose probability tends to one. Therefore,
\[
    \frac{
        c^\top(
            \widetilde{\boldsymbol\tau}_T
            -
            \boldsymbol\tau_{t_T}
        )
    }{
        \sqrt{c^\top\widehat\Sigma_Tc}
    }
    =
    \frac{
        c^\top(
            \widetilde{\boldsymbol\tau}_T
            -
            \boldsymbol\tau_{t_T}
        )
    }{
        \sqrt{c^\top\Sigma_Tc}
    }
    \left(
        \frac{c^\top\Sigma_Tc}
             {c^\top\widehat\Sigma_Tc}
    \right)^{1/2}.
\]
The first factor converges in distribution to $\mathcal{N}(0,1)$ and the second converges in probability to $1$. A final application of Slutsky's theorem proves the claim.
\end{proof}

\section{Proof of Theorem~\ref{thm:local-efficiency}}
\label{appendix-sec:proof-of-local-efficiency}
For fixed $u\in\mathbb R^K$, define
$\overline p_{T,a}(u):=p_a+u_a/\sqrt{n_{T,a}}$.
Under the locally varying experiment $\mathsf P_{T,u}$, conditional on $A_s=a$, the reward $Y_s$ is Bernoulli with mean $\overline p_{T,a}(u)+g_{T,s,a}$ for every $s\in\mathcal W_T$. Under the locally constant experiment $\mathsf Q_{T,u}$, the corresponding conditional mean is $\overline p_{T,a}(u)$. Since $g_{T,t_T,a}=0$, the target at $t_T$ is identical under the two experiments and is given by
\[
    \theta_{T,a}(u)
    =
    p_a-p_0
    +
    \frac{u_a}{\sqrt{n_{T,a}}}
    -
    \frac{u_0}{\sqrt{n_{T,0}}}.
\]

We first establish a truncated comparison between the locally varying and locally constant experiments that we will use throughout the proof.

\begin{lem}[Truncated local comparison]
\label{lem:truncated-local-comparison}
Fix $u\in\mathbb R^K$ and let $m_{T,a}:=\lceil2n_{T,a}\rceil$ and
$N_{T,a}^{(s-1)}:=\sum_{r\in\mathcal W_T:r<s}\mathbf 1\{A_r=a\}$.
For a fixed drift path $g$, let $\mathsf P_{T,u}^{g,\mathrm{tr}}$ denote the experiment whose conditional reward mean at round $s$ is $\overline p_{T,A_s}(u) +
\mathbf 1\left\{ N_{T,A_s}^{(s-1)}<m_{T,A_s}
    \right\} g_{T,s,A_s}$.
Let $\mathcal D_T
    :=
    \left\{
        g:
        \max_{s\in\mathcal W_T}
        |g_{T,s,a}|
        \leq
        b_{T,a}
        \text{ for every }a\in\cA
    \right\}$. 
Then
\begin{equation}
    \sup_{g\in\mathcal D_T}
    \mathsf{TV}
    \left(
        \mathsf P_{T,u}^{g,\mathrm{tr}},
        \mathsf Q_{T,u}
    \right)
    \leq
    C
    \left(
        \sum_{a\in\cA}
        n_{T,a}b_{T,a}^2
    \right)^{1/2}
    =:
    \epsilon_T
    \longrightarrow0.
    \label{eq:local-truncated-TV}
\end{equation}
\end{lem}

\begin{proof}
Since $\overline p_{T,a}(u)\to p_a$, $\sum_a n_{T,a}b_{T,a}^2\to0$, and $p_a\in(0,1)$ for every $a\in\cA$, there exists $\eta>0$ such that, for all sufficiently large $T$, both $\overline p_{T,a}(u)$ and $\overline p_{T,a}(u)+g_{T,s,a}$ belong to $[\eta,1-\eta]$ for every $a$, $s$, and $g\in\mathcal D_T$. Applying the sequential divergence decomposition
\citep[Lemma~15.1]{LS20},
\begin{align*}
    &D_{\mathrm{KL}}
    \left(
        \mathsf Q_{T,u}^{g,\mathrm{tr}}
        \,\middle\|\,
        \mathsf P_{T,u}
    \right)
    =
    \E_{\mathsf Q_{T,u}^{g,\mathrm{tr}}}
    \left[
        \sum_{s\in\mathcal W_T}
        D_{\mathrm{KL}}
        \left(
            \mathsf Q_{T,u}^{g,\mathrm{tr}}
            \bigl((A_s,Y_s)\mid\mathscr F_{s-1}\bigr)
            \,\middle\|\,
            \mathsf P_{T,u}
            \bigl((A_s,Y_s)\mid\mathscr F_{s-1}\bigr)
        \right)
    \right]\\
    &=
    \E_{\mathsf Q_{T,u}^{g,\mathrm{tr}}}
    \left[
        \sum_{s\in\mathcal W_T}
        \left\{
        \underbrace{
        D_{\mathrm{KL}}
        \left(
            \pi_s(\cdot\mid\mathscr F_{s-1})
            \,\middle\|\,
            \pi_s(\cdot\mid\mathscr F_{s-1})
        \right)
        }_{=\,0}
        +
        D_{\mathrm{KL}}
        \left(
            \mathsf Q_{T,u}^{g,\mathrm{tr}}
            (Y_s\mid\mathscr F_{s-1},A_s)
            \,\middle\|\,
            \mathsf P_{T,u}
            (Y_s\mid\mathscr F_{s-1},A_s)
        \right)
        \right\}
    \right]\\
    &=
    \E_{\mathsf Q_{T,u}^{g,\mathrm{tr}}}
    \left[
        \sum_{s\in\mathcal W_T}
        D_{\mathrm{KL}}
        \left(
            \operatorname{Ber}\!\left(
                \overline p_{T,A_s}(u)
                +
                \mathbf 1\left\{
                    N_{T,A_s}^{(s-1)}
                    <
                    m_{T,A_s}
                \right\}
                g_{T,s,A_s}
            \right)
            \,\middle\|\,
            \operatorname{Ber}\!\left(
                \overline p_{T,A_s}(u)
            \right)
        \right)
    \right]\\
    &\leq
    C_\eta
    \E_{\mathsf Q_{T,u}^{g,\mathrm{tr}}}
    \left[
        \sum_{s\in\mathcal W_T}
        \mathbf 1\left\{
            N_{T,A_s}^{(s-1)}
            <
            m_{T,A_s}
        \right\}
        g_{T,s,A_s}^2
    \right]\\
    &\leq
    C_\eta
    \sum_{a\in\cA}
    m_{T,a}b_{T,a}^2.
\end{align*}
The second inequality holds pathwise because the indicator can equal one on at most $m_{T,a}$ observations from arm $a$. Since $n_{T,a}\to\infty$, $m_{T,a}\leq3n_{T,a}$ for all sufficiently large $T$. Therefore, uniformly over $g\in\mathcal D_T$,
\[
    D_{\mathrm{KL}}
    \left(
        \mathsf P_{T,u}^{g,\mathrm{tr}}
        \,\middle\|\,
        \mathsf Q_{T,u}
    \right)
    \leq
    C
    \sum_{a\in\cA}
    n_{T,a}b_{T,a}^2
    \longrightarrow0.
\]
Pinsker's inequality gives \eqref{eq:local-truncated-TV}.
\end{proof}

We next use the preceding truncated comparison to transfer count stability from the locally varying baseline experiment to the locally constant baseline experiment. 

\begin{cor}[Baseline equivalence and count stability]
\label{cor:baseline-local-equivalence}
Suppose Assumption~\ref{ass:local-inference}(i) holds under $\mathsf P_{T,0}$ and \cref{eq:local-equivalence-condition} holds. Then
\begin{equation}
    \mathsf{TV}
    \left(
        \mathsf P_{T,0},
        \mathsf Q_{T,0}
    \right)
    \longrightarrow0.
    \label{eq:baseline-local-equivalence}
\end{equation}
Consequently,
\begin{equation}
    \max_{a\in\cA}
    \left|
        \frac{N_{T,a}}{n_{T,a}}-1
    \right|
    \xrightarrow{\mathsf Q_{T,0}}0.
    \label{eq:constant-count-stability}
\end{equation}
\end{cor}

\begin{proof}
Let $G_T=(g_{T,s,a})_{s,a}$ denote the possibly random exogenous drift path, and let $\nu_T$ denote its law. Conditional on $G_T=g$, let $\mathsf P_{T,0}^{g}$ denote the law of the observed history, so
$\mathsf P_{T,0}=\int\mathsf P_{T,0}^{g}\nu_T(dg)$. By \cref{eq:local-equivalence-condition}, $\nu_T(\mathcal D_T)\geq1-\delta_T$. We also define
\[
    \mathcal E_T^{\mathrm{tr}}
    :=
    \left\{
        \exists s\in\mathcal W_T:
        N_{T,A_s}^{(s-1)}
        \geq
        m_{T,A_s}
    \right\},
    \;\;
    \mathscr{C}_{T}
    :=
    \left\{
        N_{T,a}>m_{T,a}
        \text{ for some }a\in\cA
    \right\}.
\]
By construction, $\mathcal E_T^{\mathrm{tr}}\subseteq\mathscr C_T$. Moreover, since $m_{T,a}\geq2n_{T,a}$,
$\mathscr C_T\subseteq\{\max_{a\in\cA}|N_{T,a}/n_{T,a}-1|>1\}$.
Assumption~\ref{ass:local-inference}(i) and fixed $K$ therefore imply
$\mathsf P_{T,0}(\mathscr C_T)\to0$.

For fixed $g\in\mathcal D_T$, couple $\mathsf P_{T,0}^{g}$ and
$\mathsf P_{T,0}^{g,\mathrm{tr}}$ using the same action randomization and reward-generating uniforms. The two experiments coincide until $\mathcal E_T^{\mathrm{tr}}$ occurs. Hence, by the coupling inequality \citep[Proposition~4.7]{LP17} and Lemma~\ref{lem:truncated-local-comparison} with $u=0$,
$\mathsf{TV}(\mathsf P_{T,0}^{g},\mathsf Q_{T,0})
\leq\mathsf P_{T,0}^{g}(\mathcal E_T^{\mathrm{tr}})+\epsilon_T$. Therefore,
\begin{align*}
    \mathsf{TV}
    \left(
        \mathsf P_{T,0},
        \mathsf Q_{T,0}
    \right)
    &\leq
    \int_{\mathcal D_T}
    \mathsf{TV}
    \left(
        \mathsf P_{T,0}^{g},
        \mathsf Q_{T,0}
    \right)
    \nu_T(dg)
    +
    \nu_T(\mathcal D_T^c)\\
    &\leq
    \mathsf P_{T,0}(\mathscr C_T)
    +
    \epsilon_T
    +
    \delta_T
    \longrightarrow0,
\end{align*}
which proves \eqref{eq:baseline-local-equivalence}. Finally, for every $\epsilon>0$,
\[
    \mathsf Q_{T,0}
    \left(
        \max_{a\in\cA}
        \left|
            \frac{N_{T,a}}{n_{T,a}}-1
        \right|>\epsilon
    \right)
    \leq
    \mathsf P_{T,0}
    \left(
        \max_{a\in\cA}
        \left|
            \frac{N_{T,a}}{n_{T,a}}-1
        \right|>\epsilon
    \right)
    +
    \mathsf{TV}
    \left(
        \mathsf P_{T,0},
        \mathsf Q_{T,0}
    \right)
    \longrightarrow0.
\]
This proves \eqref{eq:constant-count-stability}.
\end{proof}

We now establish the joint asymptotic normality of the score vector in the locally constant experiment.

\begin{lem}[Joint score convergence]
\label{lem:local-score-clt}
Suppose Assumption~\ref{ass:local-inference}(i) holds under $\mathsf P_{T,0}$ and \cref{eq:local-equivalence-condition} holds. Let $\sigma_a^2:=p_a(1-p_a)$ and define
\[
    \Delta_{T,a}
    :=
    \frac{
        \sum_{s\in\mathcal W_T}
        \mathbf 1\{A_s=a\}(Y_s-p_a)
    }{
        \sigma_a^2\sqrt{n_{T,a}}
    },
    \;\;
    I:=
    \operatorname{diag}_{a\in\mathcal A}
    (\sigma_a^{-2}).
\]
Then
\[
    \Delta_T
    \Rightarrow
    \mathcal N_K(0,I)
    \;\;
    \text{under }\mathsf Q_{T,0}.
\]
\end{lem}

\begin{proof}
Write $\mathcal W_T=\{s_1<\cdots<s_{H_T}\}$ and use the filtration
$\{\mathscr G_{T,k}\}$ defined in Appendix~\ref{appendix-sec:proof-of-local-normality}. We prove the result by applying the martingale-array central limit theorem \citep[Corollary~3.1]{HH80} to every fixed linear combination of the score vector, and then applying the Cram\'er--Wold device \citep[Theorem~2.13]{vdV98}. We therefore verify conditional centering, square integrability, convergence of the conditional variance, and the conditional Lindeberg condition.

Recall that $\mathscr G_{T,2r-1}$ contains the observed history through the choice of $A_{s_r}$ but not the reward $Y_{s_r}$, while $\mathscr G_{T,2r}$ additionally contains $Y_{s_r}$. Define $D_{T,2r-1}:=0\in\mathbb R^K$ and, for each $a\in\mathcal A$,
\[
    D_{T,2r,a}
    :=
    \frac{
        \mathbf 1\{A_{s_r}=a\}(Y_{s_r}-p_a)
    }{
        \sigma_a^2\sqrt{n_{T,a}}
    }.
\]
Then $\Delta_T=\sum_{k=1}^{2H_T}D_{T,k}$. Each $D_{T,2r,a}$ is $\mathscr G_{T,2r}$-measurable. Moreover,
\begin{align*}
    \mathbb E[
        D_{T,2r,a}
        \mid
        \mathscr G_{T,2r-1}
    ]
    &=
    \frac{
        \mathbf 1\{A_{s_r}=a\}
    }{
        \sigma_a^2\sqrt{n_{T,a}}
    }
    \left(
        \mathbb E[
            Y_{s_r}
            \mid
            \mathscr G_{T,2r-1}
        ]
        -
        p_a
    \right)\\
    &=
    \frac{
        \mathbf 1\{A_{s_r}=a\}
    }{
        \sigma_a^2\sqrt{n_{T,a}}
    }
    (p_{A_{s_r}}-p_a)
    =
    0.
\end{align*}
Hence $\{D_{T,k},\mathscr G_{T,k}\}$ is a vector martingale difference array. Since exactly one arm is selected at each round and $|Y_{s_r}-p_a|\leq1$,
$\|D_{T,2r}\|_2^2
\leq
\max_{a\in\mathcal A}(\sigma_a^4n_{T,a})^{-1}$.
Thus every increment is square-integrable.

We next compute the conditional covariance matrix. For $a\neq b$,
$D_{T,2r,a}D_{T,2r,b}=0$ identically. For a diagonal entry,
\[
    \mathbb E[
        D_{T,2r,a}^2
        \mid
        \mathscr G_{T,2r-1}
    ]
    =
    \frac{
        \mathbf 1\{A_{s_r}=a\}
    }{
        \sigma_a^2n_{T,a}
    }.
\]
Therefore,
\begin{equation}
    \sum_{k=1}^{2H_T}
    \mathbb E[
        D_{T,k}D_{T,k}^{\top}
        \mid
        \mathscr G_{T,k-1}
    ]
    =
    \operatorname{diag}_{a\in\mathcal A}
    \left(
        \frac{N_{T,a}}{\sigma_a^2n_{T,a}}
    \right).
    \label{eq:local-score-covariance}
\end{equation}
By \cref{eq:constant-count-stability},
\[
    \left\|
        \sum_{k=1}^{2H_T}
        \mathbb E[
            D_{T,k}D_{T,k}^{\top}
            \mid
            \mathscr G_{T,k-1}
        ]
        -
        I
    \right\|_{\mathrm{op}}
    =
    \max_{a\in\mathcal A}
    \frac1{\sigma_a^2}
    \left|
        \frac{N_{T,a}}{n_{T,a}}-1
    \right|
    \xrightarrow{\mathsf Q_{T,0}}0.
\]

We also verify the vector Lindeberg condition. Since every $p_a\in(0,1)$, $K$ is fixed, and $\min_a n_{T,a}\to\infty$,
\[
    \max_{k\leq2H_T}
    \|D_{T,k}\|_2
    \leq
    \max_{a\in\mathcal A}
    \frac1{\sigma_a^2\sqrt{n_{T,a}}}
    \longrightarrow0
\]
deterministically. Hence, for every $\epsilon>0$, there exists $T_\epsilon$ such that $\|D_{T,k}\|_2\leq\epsilon$ almost surely for every $k$ whenever $T\geq T_\epsilon$. Consequently,
\[
    \sum_{k=1}^{2H_T}
    \mathbb E\left[
        \|D_{T,k}\|_2^2
        \mathbf 1\{\|D_{T,k}\|_2>\epsilon\}
        \mid
        \mathscr G_{T,k-1}
    \right]
    =
    0
\]
for all sufficiently large $T$. It remains to establish joint convergence. Fix $v\in\mathbb R^K$. The result is trivial when $v=0$. Suppose $v\neq0$ and define
$Z_{T,k}(v):=
v^\top D_{T,k}/\sqrt{v^\top Iv}$.
The increments $Z_{T,k}(v)$ are adapted, square-integrable, and conditionally centered. By \cref{eq:local-score-covariance},
\[
    \sum_{k=1}^{2H_T}
    \mathbb E[
        Z_{T,k}(v)^2
        \mid
        \mathscr G_{T,k-1}
    ]
    \xrightarrow{\mathsf Q_{T,0}}
    1.
\]
Moreover, Cauchy--Schwarz gives
$|Z_{T,k}(v)|
\leq
\|v\|_2\|D_{T,k}\|_2/\sqrt{v^\top Iv}$, so the scalar conditional Lindeberg sum is eventually zero. The martingale central limit theorem therefore yields
$v^\top\Delta_T\Rightarrow\mathcal N(0,v^\top Iv)$.
Since this holds for every $v\in\mathbb R^K$, the Cramér--Wold device gives
$\Delta_T\Rightarrow\mathcal N_K(0,I)$.
\end{proof}

We next use Lemma~\ref{lem:local-score-clt} to establish local asymptotic normality of the locally constant experiment.

\begin{lem}[Local asymptotic normality]
\label{lem:stationary-lan}
Suppose Assumption~\ref{ass:local-inference}(i) holds under $\mathsf P_{T,0}$ and \cref{eq:local-equivalence-condition} holds. Then, for every fixed $u\in\mathbb R^K$,
\begin{equation}
    \log
    \frac{
        d\mathsf Q_{T,u}
    }{
        d\mathsf Q_{T,0}
    }
    =
    u^\top\Delta_T
    -
    \frac12u^\top Iu
    +
    o_P(1).
    \label{eq:stationary-lan-expansion}
\end{equation}
The expansion holds uniformly over $u$ in every fixed bounded subset of $\mathbb R^K$. Hence $\{\mathsf Q_{T,u}\}$ is locally asymptotically normal with information matrix $I$ in the sense of \citet[Chapter~7]{vdV98}. Moreover, for every fixed $u$, $\mathsf Q_{T,u}\triangleleft\mathsf Q_{T,0}$.
\end{lem}

\begin{proof}
Write $L(y,p):=y\log p+(1-y)\log(1-p)$. Let $H_{s-1}$ denote the observed history immediately before choosing $A_s$, and let $\pi_s(\cdot\mid H_{s-1})$ denote the action-selection kernel. By the chain rule for likelihoods,
\[
    \frac{d\mathsf Q_{T,u}}{d\mathsf Q_{T,0}}
    =
    \prod_{s\in\mathcal W_T}
    \frac{
        \pi_s(A_s\mid H_{s-1})
        f_{T,u}(Y_s\mid H_{s-1},A_s)
    }{
        \pi_s(A_s\mid H_{s-1})
        f_{T,0}(Y_s\mid H_{s-1},A_s)
    }
    =
    \prod_{s\in\mathcal W_T}
    \frac{
        f_{T,u}(Y_s\mid H_{s-1},A_s)
    }{
        f_{T,0}(Y_s\mid H_{s-1},A_s)
    },
\]
since the same policy is used under both experiments. Taking logarithms therefore gives
\begin{equation}
    \log
    \frac{
        d\mathsf Q_{T,u}
    }{
        d\mathsf Q_{T,0}
    }
    =
    \sum_{a\in\mathcal A}
    \sum_{s\in\mathcal W_T}
    \mathbf 1\{A_s=a\}
    \left[
        L\left(
            Y_s,
            p_a+\frac{u_a}{\sqrt{n_{T,a}}}
        \right)
        -
        L(Y_s,p_a)
    \right].
    \label{eq:stationary-likelihood-ratio}
\end{equation}

Since each $p_a\in(0,1)$ and $\min_a n_{T,a}\to\infty$, for every fixed bounded set $\mathcal U\subset\mathbb R^K$ there exists $\eta>0$ such that $p_a+u_a/\sqrt{n_{T,a}}\in[\eta,1-\eta]$ for every $a\in\mathcal A$, $u\in\mathcal U$, and all sufficiently large $T$. We have
$\nabla_pL(y,p)=(y-p)/(p(1-p))$ and
$\nabla_{p}^2L(y,p)=-y/p^2-(1-y)/(1-p)^2$.
Moreover, $\nabla_{p}^3L(y,p)$ is uniformly bounded over $y\in\{0,1\}$ and $p\in[\eta,1-\eta]$. Taylor's theorem therefore gives
\begin{align*}
    L\left(
        y,
        p_a+\frac{u_a}{\sqrt{n_{T,a}}}
    \right)
    -
    L(y,p_a)
    &=
    \frac{
        u_a(y-p_a)
    }{
        \sigma_a^2\sqrt{n_{T,a}}
    }-
    \frac{u_a^2}{2n_{T,a}}
    \left(
        \frac{y}{p_a^2}
        +
        \frac{1-y}{(1-p_a)^2}
    \right)
    +
    \widetilde{\mathscr R}_{T,a}(y;u),
\end{align*}
where $\left| \widetilde{\mathscr R}_{T,a}(y;u) \right| \leq C |u_a|^3 / n_{T,a}^{3/2}$
uniformly over $y\in\{0,1\}$ and $u\in\mathcal U$. Substituting this expansion into \cref{eq:stationary-likelihood-ratio} yields
\begin{align}
    \log
    \frac{
        d\mathsf Q_{T,u}
    }{
        d\mathsf Q_{T,0}
    }
    ={}&
    \sum_{a\in\mathcal A}
    \frac{u_a}{
        \sigma_a^2\sqrt{n_{T,a}}
    }
    \sum_{s\in\mathcal W_T}
    \mathbf 1\{A_s=a\}(Y_s-p_a)
    \notag\\
    &-
    \frac12
    \sum_{a\in\mathcal A}
    \frac{u_a^2}{n_{T,a}}
    \sum_{s\in\mathcal W_T}
    \mathbf 1\{A_s=a\}
    \left(
        \frac{Y_s}{p_a^2}
        +
        \frac{1-Y_s}{(1-p_a)^2}
    \right)
    +
    \mathscr R_T(u),
    \label{eq:stationary-taylor-expansion}
\end{align}
where
$\widetilde{\mathscr R}_T(u)
:=
\sum_{a\in\mathcal A}
\sum_{s\in\mathcal W_T}
\mathbf 1\{A_s=a\}
\widetilde{\mathscr R}_{T,a}(Y_s;u)$.
Consequently,
$|\widetilde{\mathscr R}_T(u)|
\leq
C\sum_{a\in\mathcal A}|u_a|^3N_{T,a}/n_{T,a}^{3/2}$.
For fixed $u$, \cref{eq:constant-count-stability} gives
$|u_a|^3N_{T,a}/n_{T,a}^{3/2}
=
|u_a|^3(N_{T,a}/n_{T,a})n_{T,a}^{-1/2}
=
o_P(1)$
under $\mathsf Q_{T,0}$ for every arm. Since $K$ is fixed, $\widetilde{\mathscr R}_T(u)=o_P(1)$.

The same argument is uniform over bounded $u$. Indeed, if
$\sup_{u\in\mathcal U}\|u\|_\infty\leq M$, then
\[
    \sup_{u\in\mathcal U}
    |\widetilde{\mathscr R}_T(u)|
    \leq
    CM^3
    \sum_{a\in\mathcal A}
    \frac{N_{T,a}}{n_{T,a}}
    \frac1{\sqrt{n_{T,a}}}
    =
    o_P(1),
\]
since $N_{T,a}/n_{T,a}=O_P(1)$ under $\mathsf Q_{T,0}$ for every arm, $\min_a n_{T,a}\to\infty$, and $K$ is fixed. We next simplify the quadratic term. 
\begin{align*}
    &\frac1{n_{T,a}}
    \sum_{s\in\mathcal W_T}
    \mathbf 1\{A_s=a\}
    \left(
        \frac{Y_s}{p_a^2}
        +
        \frac{1-Y_s}{(1-p_a)^2}
    \right) =
    \frac{N_{T,a}}{n_{T,a}\sigma_a^2}
    +
    \left(
        \frac1{p_a^2}
        -
        \frac1{(1-p_a)^2}
    \right)
    \frac1{n_{T,a}}
    \sum_{s\in\mathcal W_T}
    \mathbf 1\{A_s=a\}(Y_s-p_a).
\end{align*}
By Lemma~\ref{lem:local-score-clt},
$\sum_{s\in\mathcal W_T}
\mathbf 1\{A_s=a\}(Y_s-p_a)
=
O_P(\sqrt{n_{T,a}})$
under $\mathsf Q_{T,0}$. Hence the second term is
$O_P(n_{T,a}^{-1/2})=o_P(1)$, while the first converges in probability to
$\sigma_a^{-2}$ by \cref{eq:constant-count-stability}. Thus
\begin{equation}
    \frac1{n_{T,a}}
    \sum_{s\in\mathcal W_T}
    \mathbf 1\{A_s=a\}
    \left(
        \frac{Y_s}{p_a^2}
        +
        \frac{1-Y_s}{(1-p_a)^2}
    \right)
    =
    \frac1{\sigma_a^2}
    +
    o_P(1).
    \label{eq:stationary-quadratic-term}
\end{equation}

Since $K$ is fixed, multiplying \cref{eq:stationary-quadratic-term} by the bounded coefficients $u_a^2$ and summing over $a$ gives, uniformly over $u$ in every fixed bounded set,
\[
    \sum_{a\in\mathcal A}
    \frac{u_a^2}{n_{T,a}}
    \sum_{s\in\mathcal W_T}
    \mathbf 1\{A_s=a\}
    \left(
        \frac{Y_s}{p_a^2}
        +
        \frac{1-Y_s}{(1-p_a)^2}
    \right)
    =
    u^\top Iu
    +
    o_P(1).
\]
The linear term in \cref{eq:stationary-taylor-expansion} is exactly $u^\top\Delta_T$. Combining this with the preceding quadratic expansion and $\widetilde{\mathscr R}_T(u)=o_P(1)$ gives
\[
    \log
    \frac{
        d\mathsf Q_{T,u}
    }{
        d\mathsf Q_{T,0}
    }
    =
    u^\top\Delta_T
    -
    \frac12u^\top Iu
    +
    o_P(1).
\]
Together with Lemma~\ref{lem:local-score-clt}, this is the quadratic likelihood expansion defining local asymptotic normality \citep[Chapter~7]{vdV98}. The preceding uniform bounds also show that the expansion holds along every convergent sequence $u_T\to u$.

It remains to establish contiguity. By Lemma~\ref{lem:local-score-clt} and the continuous mapping theorem, under $\mathsf Q_{T,0}$,
\[
    \log
    \frac{
        d\mathsf Q_{T,u}
    }{
        d\mathsf Q_{T,0}
    }
    \Rightarrow
    u^\top Z
    -
    \frac12u^\top Iu,
    \;\;
    Z\sim\mathcal N_K(0,I).
\]
Exponentiating gives
$d\mathsf Q_{T,u}/d\mathsf Q_{T,0}
\Rightarrow
V_u
:=
\exp\{u^\top Z-\frac12u^\top Iu\}$.
Since $u^\top Z\sim\mathcal N(0,u^\top Iu)$, it holds that $\mathbb E[V_u] = 1$. 
Moreover, $V_u>0$ almost surely. Le Cam's first lemma \citep[Lemma~6.4]{vdV98} therefore implies $\mathsf Q_{T,u}\triangleleft\mathsf Q_{T,0}$.
\end{proof}

We now show that the locally varying and locally constant experiments are asymptotically equivalent.

\begin{lem}[Local equivalence]
\label{lem:local-experiment-equivalence}
Suppose Assumption~\ref{ass:local-inference}(i) holds under $\mathsf P_{T,0}$ and \cref{eq:local-equivalence-condition} holds. Then, for every fixed $u\in\mathbb R^K$,
\[
    \mathsf{TV}
    \left(
        \mathsf P_{T,u},
        \mathsf Q_{T,u}
    \right)
    \longrightarrow0.
\]
\end{lem}

\begin{proof}
The case $u=0$ follows from Corollary~\ref{cor:baseline-local-equivalence}. Fix $u\in\mathbb R^K$ and recall
$\mathscr C_T:=\{N_{T,a}>m_{T,a}\text{ for some }a\in\cA\}$.
By \eqref{eq:constant-count-stability} and fixed $K$,
$\mathsf Q_{T,0}(\mathscr C_T)\to0$. Lemma~\ref{lem:stationary-lan} gives
$\mathsf Q_{T,u}\triangleleft\mathsf Q_{T,0}$, so
$\mathsf Q_{T,u}(\mathscr C_T)\to0$.

Let $G_T=(g_{T,s,a})_{s,a}$ denote the possibly random exogenous drift path, let $\nu_T$ denote its law, and let $\mathsf P_{T,u}^{g}$ denote the conditional law of the observed history given $G_T=g$. Thus
$\mathsf P_{T,u}=\int\mathsf P_{T,u}^{g}\nu_T(dg)$ and
$\nu_T(\mathcal D_T)\geq1-\delta_T$. For $g\in\mathcal D_T$, Lemma~\ref{lem:truncated-local-comparison} gives $\mathsf{TV}(\mathsf P_{T,u}^{g,\mathrm{tr}},\mathsf Q_{T,u})\leq\epsilon_T$.
Since $\mathcal E_T^{\mathrm{tr}}\subseteq\mathscr C_T$, the basic total-variation inequality gives
\[
    \sup_{g\in\mathcal D_T}
    \mathsf P_{T,u}^{g,\mathrm{tr}}
    (\mathcal E_T^{\mathrm{tr}})
    \leq
    \mathsf Q_{T,u}(\mathscr C_T)
    +
    \epsilon_T
    \longrightarrow0.
\]
As in the proof of Corollary~\ref{cor:baseline-local-equivalence}, couple
$\mathsf P_{T,u}^{g}$ and $\mathsf P_{T,u}^{g,\mathrm{tr}}$ using the same action randomization and reward-generating uniforms. The two experiments coincide until $\mathcal E_T^{\mathrm{tr}}$ occurs, so the coupling inequality \citep[Proposition~4.7]{LP17} gives
$\sup_{g\in\mathcal D_T}
\mathsf{TV}(\mathsf P_{T,u}^{g},\mathsf P_{T,u}^{g,\mathrm{tr}})
\leq
\sup_{g\in\mathcal D_T}
\mathsf P_{T,u}^{g,\mathrm{tr}}(\mathcal E_T^{\mathrm{tr}})
\to0$.
Combining this with Lemma~\ref{lem:truncated-local-comparison} and the triangle inequality yields
$\sup_{g\in\mathcal D_T}
\mathsf{TV}(\mathsf P_{T,u}^{g},\mathsf Q_{T,u})\to0$.

Finally,
\begin{align*}
    \mathsf{TV}
    \left(
        \mathsf P_{T,u},
        \mathsf Q_{T,u}
    \right)
    &\leq
    \int_{\mathcal D_T}
    \mathsf{TV}
    \left(
        \mathsf P_{T,u}^{g},
        \mathsf Q_{T,u}
    \right)
    \nu_T(dg)
    +
    \nu_T(\mathcal D_T^c)\\
    &\leq
    \sup_{g\in\mathcal D_T}
    \mathsf{TV}
    \left(
        \mathsf P_{T,u}^{g},
        \mathsf Q_{T,u}
    \right)
    +
    \delta_T
    \longrightarrow0.
\end{align*}
This proves the claim.
\end{proof}

We next establish the convolution lower bound and efficiency of the all-observation estimator in the locally constant experiment.

\begin{lem}[Stationary convolution bound and efficiency]
\label{lem:stationary-local-efficiency}
Fix $a\in\cA_+$ and suppose $n_{T,a}/n_{T,0}$ converges in $[0,\infty]$. Then every regular estimator of $\theta_{T,a}(u)$ in the locally constant experiment has limiting distribution $\mathcal N(0,1)\ast M$ after normalization by $\sqrt{\Sigma_T(a,a)}$, for some probability law $M$. Moreover, for every fixed $u\in\mathbb R^K$,
\[
    \frac{
        \widetilde\tau_T(a)-\theta_{T,a}(u)
    }{
        \sqrt{\Sigma_T(a,a)}
    }
    \Rightarrow
    \mathcal N(0,1)
    \;\;
    \text{under }\mathsf Q_{T,u}.
\]
\end{lem}

\begin{proof}
Write $\Sigma_T(a,a):=\sigma_0^2/n_{T,0}+\sigma_a^2/n_{T,a}$ and define $c_T\in\mathbb R^K$ by $c_{T,0}:=-1/\sqrt{n_{T,0}\Sigma_T(a,a)}$, $c_{T,a}:=1/\sqrt{n_{T,a}\Sigma_T(a,a)}$, and $c_{T,b}:=0$ for $b\notin\{0,a\}$. Let $r_T:=n_{T,a}/n_{T,0}$. Since $r_T$ converges in $[0,\infty]$, $c_T\to c$ for some deterministic vector $c$. Moreover, since $I^{-1}=\operatorname{diag}_{b\in\cA}(\sigma_b^2)$,
\[
    c_T^\top I^{-1}c_T
    =
    \frac{
        \sigma_0^2/n_{T,0}
        +
        \sigma_a^2/n_{T,a}
    }{
        \Sigma_T(a,a)
    }
    =
    1,
    \;\;
    \frac{
        \theta_{T,a}(u)-\theta_{T,a}(0)
    }{
        \sqrt{\Sigma_T(a,a)}
    }
    =
    c_T^\top u
    \longrightarrow
    c^\top u.
\]
Thus the limiting derivative of the normalized target is $u\mapsto c^\top u$, with $c^\top I^{-1}c=1$. Lemma~\ref{lem:stationary-lan} and the H\'ajek--Le Cam convolution theorem \citep[Theorem~8.8]{vdV98} therefore imply that every regular estimator has limiting law $\mathcal N(0,1)\ast M$ for some probability law $M$.

We now show that $\widetilde\tau_T(a)$ attains this bound. Under $\mathsf Q_{T,0}$, the zero-drift version of Lemma~\ref{appendix-lem:first-part} gives
\begin{equation}
    \frac{
        \widetilde\tau_T(a)-(p_a-p_0)
    }{
        \sqrt{\Sigma_T(a,a)}
    }
    =
    c_T^\top I^{-1}\Delta_T
    +
    o_P(1).
    \label{eq:stationary-efficient-linearization}
\end{equation}
By Lemma~\ref{lem:stationary-lan} and Lemma~\ref{lem:local-score-clt}, under $\mathsf Q_{T,0}$,
\[
    \left(
        c_T^\top I^{-1}\Delta_T,
        \log
        \frac{
            d\mathsf Q_{T,u}
        }{
            d\mathsf Q_{T,0}
        }
    \right)
    \Rightarrow
    \left(
        c^\top I^{-1}Z,
        u^\top Z-\frac12u^\top Iu
    \right),
    \;\;
    Z\sim\mathcal N_K(0,I).
\]
Since $\operatorname{Var}(c^\top I^{-1}Z)=1$ and
$\operatorname{Cov}(c^\top I^{-1}Z,u^\top Z)=c^\top u$, Le Cam's third lemma \citep[Example~6.7]{vdV98} gives
$c_T^\top I^{-1}\Delta_T\Rightarrow\mathcal N(c^\top u,1)$ under $\mathsf Q_{T,u}$. Moreover, Lemma~\ref{lem:stationary-lan} gives $\mathsf Q_{T,u}\triangleleft\mathsf Q_{T,0}$, so the remainder in \eqref{eq:stationary-efficient-linearization} is $o_{\mathsf Q_{T,u}}(1)$. Therefore,
\[
    \frac{
        \widetilde\tau_T(a)-\theta_{T,a}(u)
    }{
        \sqrt{\Sigma_T(a,a)}
    }
    =
    c_T^\top I^{-1}\Delta_T
    -
    c_T^\top u
    +
    o_{\mathsf Q_{T,u}}(1)
    \Rightarrow
    \mathcal N(0,1),
\]
where the last step follows from $c_T^\top u\to c^\top u$ and Slutsky's theorem. Hence $\widetilde\tau_T(a)$ attains the convolution lower bound and is locally asymptotically efficient.
\end{proof}

We can now prove Theorem~\ref{thm:local-efficiency}.

\begin{proof}[Proof of Theorem~\ref{thm:local-efficiency}]
Part~(i) follows from Lemma~\ref{lem:local-experiment-equivalence}. Part~(ii) follows from the convolution bound in Lemma~\ref{lem:stationary-local-efficiency}. Part~(iii) follows from the efficiency result in Lemma~\ref{lem:stationary-local-efficiency}. This proves the theorem.
\end{proof}

\section{Proof of Theorem~\ref{thm:smoothness-adaptation}}
\label{appendix-sec:proof-of-smoothness-adaptation}

\begin{proof}[Proof of Theorem~\ref{thm:smoothness-adaptation}]
Let $N_0(T):=\sum_{t=1}^T\mathbf 1\{A_t=0\}$. Under the constant environment $\mu^\circ$, arm $1$ is optimal at every round and each pull of arm $0$ incurs regret $1/2$. Hence
$\mathscr{R}_T(\pi,\mu^\circ)=\frac12\E_{\mu^\circ}^\pi[N_0(T)]\leq r$, which implies
\begin{equation}
    \E_{\mu^\circ}^\pi[N_0(T)]
    \leq
    2r.
    \label{eq:adaptation-baseline-count}
\end{equation}

We next construct a collection of disjoint local perturbations in the second half of the horizon. Let
$x:=(2\Delta/L)^{1/\beta}$ and $\ell:=\lceil Tx\rceil$. By \cref{eq:adaptation-range}, $Tx>1$ and $x\leq1/32$. Therefore $Tx\leq\ell\leq2Tx$ and $\ell\leq T/16$. Define
$q:=\lfloor T/(4\ell)\rfloor$ and
$s_i:=\lfloor T/2\rfloor+(2i-1)\ell$ for $i\in[q]$. For each $i$, let
$\mathcal S_i:=\{t\in[T]:|t-s_i|<\ell\}$ and define
\[
    \psi_i(u)
    :=
    \left(
        1-\frac{|Tu-s_i|}{\ell}
    \right)_+^\beta.
\]
The sets $\{\mathcal S_i\}_{i\in[q]}$ are pairwise disjoint, and every peak $s_i$ belongs to $[\lceil T/2\rceil,T]$. Moreover, since $T/(4\ell)\geq4$, we have
\[
    q
    \geq
    \frac{T}{8\ell}
    \geq
    \frac{1}{16x}
    =
    2^{-4-1/\beta}
    \left(
        \frac{L}{\Delta}
    \right)^{1/\beta}.
\]
Thus $q\gtrsim_\beta(L/\Delta)^{1/\beta}$.

For each $i\in[q]$, define the number of arm-$0$ observations in $\mathcal S_i$ by
$N_i:=\sum_{t\in\mathcal S_i}\mathbf 1\{A_t=0\}$. Since the supports are pairwise disjoint, $\sum_{i=1}^qN_i\leq N_0(T)$ on every sample path. Taking expectations under $\mu^\circ$ and using \cref{eq:adaptation-baseline-count},
\[
    \sum_{i=1}^q
    \E_{\mu^\circ}^\pi[N_i]
    \leq
    2r.
\]
Hence there exists an index $i^\star\in[q]$ such that
$\E_{\mu^\circ}^\pi[N_{i^\star}]\leq2r/q$. Fix this index and define an alternative environment $\mu^{i^\star}$ by
$\mu_1^{i^\star}\equiv3/4$ and
$\mu_0^{i^\star}(u):=1/4+\Delta\psi_{i^\star}(u)$.
Since $|x^\beta-y^\beta|\leq|x-y|^\beta$ for $\beta\in(0,1]$, the function $\psi_{i^\star}$ is $\beta$-H\"older with constant $(T/\ell)^\beta$. Hence $\Delta\psi_{i^\star}$ has H\"older constant
$ \Delta(T / \ell)^\beta
    \leq
    \Delta x^{-\beta}
    \leq
    L$. 
Also, $\Delta\leq1/16$ implies
$\mu_0^{i^\star}(u)\in[1/4,5/16]$ for every $u$. Thus
$\mu^{i^\star}\in\mathcal M_{\beta,L}$.

Let $\mathsf P_\circ^\pi$ and $\mathsf P_{i^\star}^\pi$ denote the observed historical laws under $\mu^\circ$ and $\mu^{i^\star}$, respectively. The two environments differ only when $t\in\mathcal S_{i^\star}$ and arm $0$ is selected. Therefore, by the adaptive divergence decomposition \citep[Lemma~15.1]{LS20},
\begin{align*}
    D_{\mathrm{KL}}
    \left(
        \mathsf P_\circ^\pi
        \,\middle\|\,
        \mathsf P_{i^\star}^\pi
    \right)
    &=
    \E_{\mu^\circ}^\pi
    \left[
        \sum_{t\in\mathcal S_{i^\star}}
        \mathbf 1\{A_t=0\}
        \operatorname{kl}
        \left(
            \frac14,
            \frac14+\Delta\psi_{i^\star}(t/T)
        \right)
    \right] \\
    &\leq
    32\Delta^2
    \E_{\mu^\circ}^\pi[N_{i^\star}]
    \leq
    \frac{64r\Delta^2}{q}.
\end{align*}
Indeed, for $0\leq z\leq\Delta$, the second Bernoulli parameter $1/4+z$ belongs to $[1/4,5/16]$, and
$\operatorname{kl}(p,q)\leq(p-q)^2/(q(1-q))\leq32\Delta^2$. Using the lower bound on $q$, there exists a constant $C_\beta>0$ depending only on $\beta$ such that
\[
    D_{\mathrm{KL}}
    \left(
        \mathsf P_\circ^\pi
        \,\middle\|\,
        \mathsf P_{i^\star}^\pi
    \right)
    \leq
    C_\beta
    \frac{
        r\Delta^{2+1/\beta}
    }{
        L^{1/\beta}
    }.
\]
By choosing the constant $c_\beta$ in \cref{eq:adaptation-range} sufficiently small, the right-hand side is at most $1/8$. Pinsker's inequality then gives
\begin{equation}
    \mathsf{TV}
    \left(
        \mathsf P_\circ^\pi,
        \mathsf P_{i^\star}^\pi
    \right)
    \leq
    \sqrt{
        \frac12
        D_{\mathrm{KL}}
        \left(
            \mathsf P_\circ^\pi
            \,\middle\|\,
            \mathsf P_{i^\star}^\pi
        \right)
    }
    \leq
    \frac14.
    \label{eq:adaptation-TV}
\end{equation}

At the peak $s_{i^\star}$, $\psi_{i^\star}(s_{i^\star}/T)=1$. Hence the current effect under $\mu^\circ$ is
$\tau_{s_{i^\star}}^\circ=3/4-1/4=1/2$, while under $\mu^{i^\star}$ it is
$\tau_{s_{i^\star}}^{i^\star}=3/4-(1/4+\Delta)=1/2-\Delta$. Simultaneous coverage implies
$\mathsf P_\circ^\pi(1/2\in \mathscr{C}_{s_{i^\star}})\geq1-\delta$ and
$\mathsf P_{i^\star}^\pi(1/2-\Delta\in \mathscr{C}_{s_{i^\star}})\geq1-\delta$.
By \cref{eq:adaptation-TV},
\[
    \mathsf P_\circ^\pi
    \left(
        \frac12-\Delta
        \in
        \mathscr{C}_{s_{i^\star}}
    \right)
    \geq
    \mathsf P_{i^\star}^\pi
    \left(
        \frac12-\Delta
        \in
        \mathscr{C}_{s_{i^\star}}
    \right)
    -
    \mathsf{TV}
    \left(
        \mathsf P_\circ^\pi,
        \mathsf P_{i^\star}^\pi
    \right)
    \geq
    1-\delta-\frac14.
\]
Therefore, using the standing assumption $\delta\leq1/8$,
\begin{align*}
    \mathsf P_\circ^\pi
    \left(
        \left\{
            \frac12,
            \frac12-\Delta
        \right\}
        \subseteq
        \mathscr{C}_{s_{i^\star}}
    \right)
    &\geq
    1
    -
    \mathsf P_\circ^\pi
    \left(
        \frac12\notin \mathscr{C}_{s_{i^\star}}
    \right)
    -
    \mathsf P_\circ^\pi
    \left(
        \frac12-\Delta\notin \mathscr{C}_{s_{i^\star}}
    \right)\\
    &\geq
    1-2\delta-\frac14
    \geq
    \frac12.
\end{align*}

On this event, the interval $\mathscr{C}_{s_{i^\star}}$ contains two points separated by $\Delta$, so
$|\mathscr{C}_{s_{i^\star}}|\geq\Delta$. Consequently,
\[
    \E_{\mu^\circ}^\pi
    \left[
        |\mathscr{C}_{s_{i^\star}}|
    \right]
    \geq
    \Delta\,
    \mathsf P_\circ^\pi
    \left(
        \left\{
            \frac12,
            \frac12-\Delta
        \right\}
        \subseteq
        \mathscr{C}_{s_{i^\star}}
    \right)
    \geq
    \frac{\Delta}{2}.
\]
Since $s_{i^\star}\in[\lceil T/2\rceil,T]$, it follows that
\[
    \max_{\lceil T/2\rceil\leq t\leq T}
    \E_{\mu^\circ}^\pi|\mathscr{C}_t|
    \geq
    \E_{\mu^\circ}^\pi
    |\mathscr{C}_{s_{i^\star}}|
    \geq
    \frac{\Delta}{2}.
\]
This proves \cref{eq:adaptation-lower-bound}.
\end{proof}

\section{Additional experimental results}
\label{app:additional-experimental-results}

For completeness, we report the full numerical results corresponding to the real-data experiments in Section~\ref{sec:numerical-experiments}, including Yahoo! results across varying values of $m$, item-level ZOZOTOWN results, and additional HeartSteps estimates.

\begin{table}[H]
    \centering
    \small
    \setlength{\tabcolsep}{5pt}
    \renewcommand{\arraystretch}{1.1}
    \caption{\textbf{Yahoo! R6B results.} Dynamic regret, current-effect RMSE,
    and fraction of rounds used for inference across inferential budgets.
    Adaptive \textsc{TRACE} uses the Beta$(1,1)$-smoothed historical
    click-through rate as its predictable reward forecast. Results are
    averaged over $200$ Monte Carlo replications.}
    \label{tab:yahoo-full-results}
    \begin{tabular}{ccccccc}
        \toprule
        & \multicolumn{3}{c}{\textsc{TRACE}}
        & \multicolumn{3}{c}{Adaptive \textsc{TRACE}} \\
        \cmidrule(lr){2-4}
        \cmidrule(lr){5-7}
        $m$
        & Inference fraction
        & Regret
        & Effect RMSE
        & Inference fraction
        & Regret
        & Effect RMSE \\
        \midrule
        25  & $2.5\%$  & $1353.8$ & $0.0578$
            & $2.5\%$  & $1354.3$ & $0.0574$ \\
        50  & $5.0\%$  & $1363.5$ & $0.0407$
            & $5.0\%$  & $1349.0$ & $0.0410$ \\
        100 & $10.0\%$ & $1387.3$ & $0.0290$
            & $9.9\%$  & $1385.8$ & $0.0295$ \\
        200 & $20.0\%$ & $1430.6$ & $0.0207$
            & $15.0\%$ & $1400.8$ & $0.0231$ \\
        400 & $40.0\%$ & $1506.8$ & $0.0147$
            & $22.2\%$ & $1432.0$ & $0.0191$ \\
        600 & $60.0\%$ & $1566.1$ & $0.0121$
            & $28.2\%$ & $1441.4$ & $0.0166$ \\
        800 & $80.0\%$ & $1603.5$ & $0.0108$
            & $33.8\%$ & $1451.9$ & $0.0154$ \\
        \bottomrule
    \end{tabular}
\end{table}

\begin{table}[H]
    \centering
    \small
    \setlength{\tabcolsep}{6pt}
    \renewcommand{\arraystretch}{1.1}
    \caption{\textbf{ZOZOTOWN results.} Prediction RMSE and number of
    randomized observations required by Adaptive \textsc{TRACE} for each
    selected item. The predictable reward forecast is the Beta$(1,1)$-smoothed
    historical click-through rate computed using only observations preceding
    the current impression.}
    \label{tab:zozo-full-results}
    \begin{tabular}{ccccc}
        \toprule
        Item
        & Available
        & Observations used
        & Prediction RMSE
        & Reduction \\
        \midrule
        18 & $5{,}312$ & $3{,}441$ & $0.0628$ & $35.2\%$ \\
        30 & $5{,}156$ & $3{,}012$ & $0.0462$ & $41.6\%$ \\
        29 & $5{,}584$ & $2{,}705$ & $0.0354$ & $51.6\%$ \\
        24 & $5{,}486$ & $3{,}685$ & $0.0687$ & $32.8\%$ \\
        43 & $5{,}146$ & $2{,}892$ & $0.0442$ & $43.8\%$ \\
        \midrule
        Mean
        & $5{,}336.8$
        & $3{,}147.0$
        & $0.0514$
        & $41.0\%$ \\
        \bottomrule
    \end{tabular}
\end{table}

\begin{table}[H]
    \centering
    \small
    \setlength{\tabcolsep}{5pt}
    \renewcommand{\arraystretch}{1.1}
    \caption{\textbf{Weekly randomized contrasts in HeartSteps.} Confidence
    intervals use participant-clustered standard errors.}
    \label{tab:heartsteps-weekly}
    \begin{tabular}{cccccc}
        \toprule
        Week & Decisions & Treated & Control & Estimate & $95\%$ CI \\
        \midrule
        1 & $960$     & $576$ & $384$ & $0.621$  & $[0.298,\ 0.945]$ \\
        2 & $1{,}061$ & $635$ & $426$ & $0.015$  & $[-0.391,\ 0.421]$ \\
        3 & $1{,}037$ & $603$ & $434$ & $0.104$  & $[-0.257,\ 0.466]$ \\
        4 & $1{,}049$ & $608$ & $441$ & $0.055$  & $[-0.319,\ 0.428]$ \\
        5 & $1{,}034$ & $624$ & $410$ & $-0.163$ & $[-0.567,\ 0.241]$ \\
        6 & $920$     & $548$ & $372$ & $0.098$  & $[-0.265,\ 0.461]$ \\
        \bottomrule
    \end{tabular}
\end{table}

\begin{figure}[H]
    \centering
    \includegraphics[width=0.60\textwidth]
    {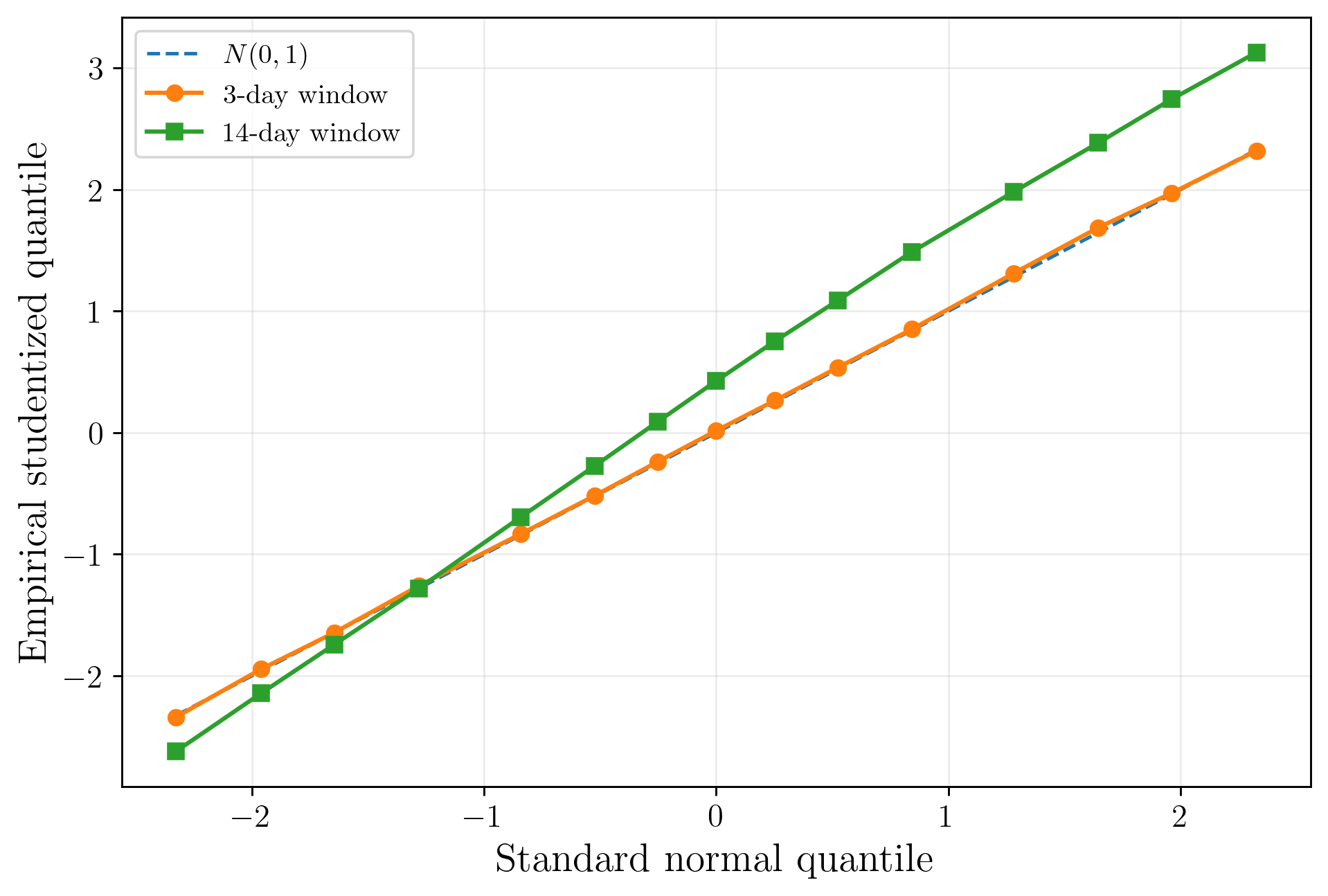}
    \caption{\textbf{Studentized local inference in HeartSteps.}
    Quantiles of the studentized all-observation estimator against standard
    normal quantiles for a locally stable three-day window and a fourteen-day
    window with larger temporal drift. The three-day estimator closely follows
    the standard normal reference, while the fourteen-day distribution is
    shifted by the accumulated temporal bias.}
    \label{fig:heartsteps-qq}
\end{figure}

\end{document}